\documentclass[10pt,twocolumn,letterpaper]{article}

\usepackage[pagenumbers]{cvpr} 

\usepackage{graphicx}
\usepackage{amsmath,bm}
\usepackage{amssymb}
\usepackage{booktabs}
\usepackage{siunitx}
\usepackage{algorithm}
\usepackage{algpseudocode}
\usepackage{makecell}
\usepackage{multirow}
\usepackage{arydshln}
\usepackage{stackengine}
\usepackage{subcaption}
\usepackage[export]{adjustbox}

\DeclareMathOperator*{\argmin}{argmin}
\providecommand{\tightlist}{%
	\setlength{\itemsep}{0pt}\setlength{\parskip}{0pt}}

\usepackage[pagebackref,breaklinks,colorlinks]{hyperref}

\usepackage[capitalize]{cleveref}
\crefname{section}{Sec.}{Secs.}
\crefname{section}{Section}{Sections}
\crefname{table}{Table}{Tables}
\crefname{table}{Tab.}{Tabs.}
\crefname{algorithm}{Alg.}{Algs.}

\begin{document}

\title{Learning Privacy-Enhancing Optical Embeddings with a Programmable Mask}
\title{Privacy-Enhancing Optical Embeddings for Lensless Classification}

\author{%
	Eric Bezzam$^\dagger $, Martin Vetterli$^\dagger $, Matthieu Simeoni$^\ddagger $ \\
	Audiovisual Communications Laboratory$^\dagger $, Center for Imaging$^\ddagger $\\
	\'{E}cole Polytechnique F\'{e}d\'{e}rale de Lausanne (EPFL)\\
	{\tt\small first.last@epfl.ch}
}

\maketitle

\begin{abstract}
Lensless imaging can provide visual privacy due to the highly multiplexed characteristic of its measurements. However,
this alone is
a weak form of security, as various adversarial attacks can be designed to invert the one-to-many scene mapping of such cameras. In this work, we enhance the privacy provided by lensless imaging by (1) downsampling at the sensor and (2) using a programmable mask with variable patterns as our optical encoder. We build a prototype from a low-cost LCD and Raspberry Pi components, for a total cost of around 100 USD. This very low price point allows our system to be deployed and leveraged in a broad range of applications. In our experiments, we first demonstrate the viability and reconfigurability of our system by applying it to various classification tasks: MNIST, CelebA (face attributes), and CIFAR10. By jointly optimizing the mask pattern and a digital classifier in an end-to-end fashion, low-dimensional, privacy-enhancing embeddings are learned directly at the sensor. Secondly, we show how the proposed system, through variable mask patterns, can thwart adversaries that attempt to invert the system (1) via plaintext attacks or (2) in the event of camera parameters leaks. We demonstrate the defense of our system to both risks, with \SI{55}{\percent} and \SI{26}{\percent} drops in image quality metrics for attacks based on model-based convex optimization and generative neural networks respectively.  We open-source a wave propagation and camera simulator needed for end-to-end optimization, the training software, and a library for interfacing with the camera.
\end{abstract}

\section{Introduction}
\label{sec:intro}

\begin{figure}[t!]
	\centering
	\includegraphics[width=0.95\linewidth]{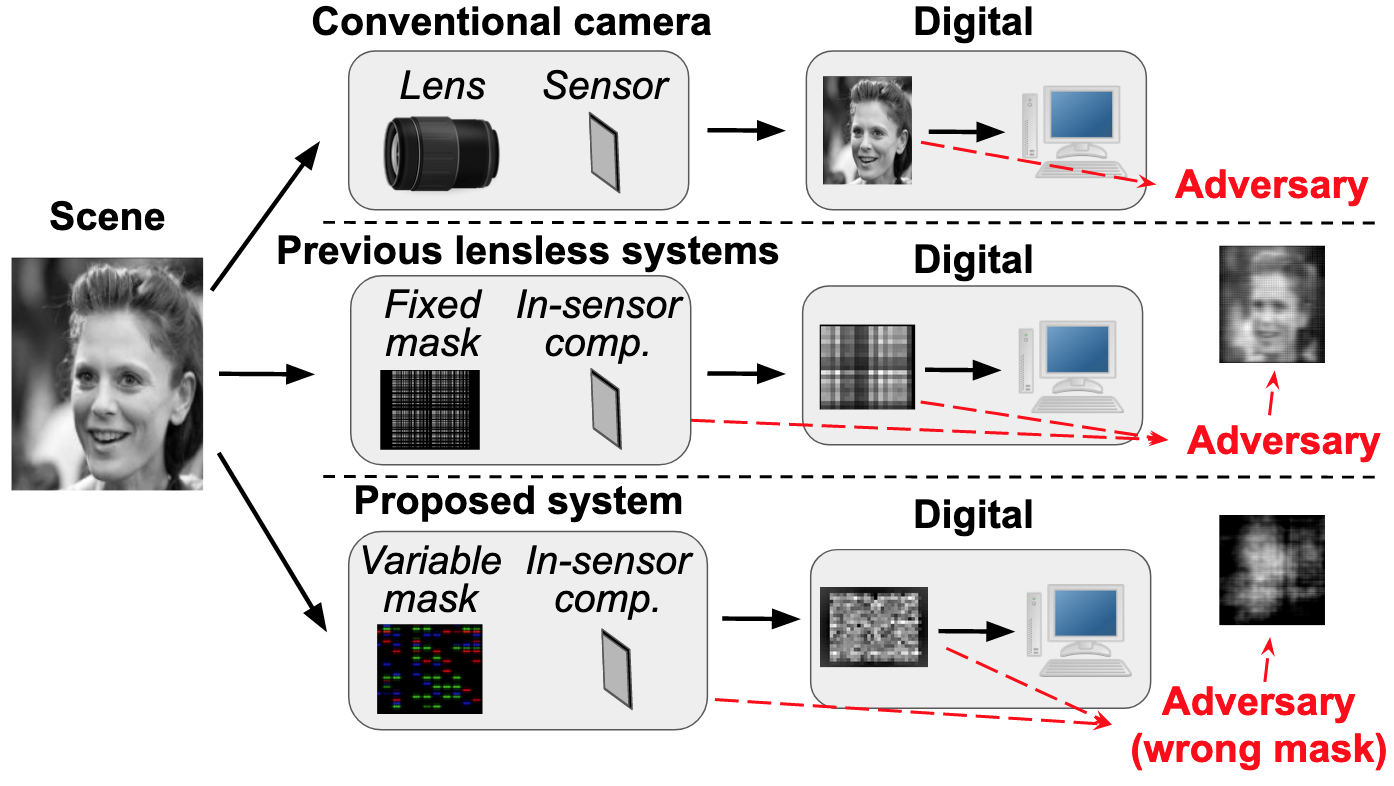} 
	\caption{In a conventional camera (top), the optics and hardware try to measure an image as close as possible to the original scene. Leaks in captured data can reveal sensitive information such as identity. To achieve visual privacy, previous work has introduced distortions at the optics and sensor (middle). However, in the event of camera parameters leaks, an adversary can still recover an image of the scene to obtain sensitive information. To protect against both data and system leaks, we propose using a programmable mask with a varying pattern, which makes it difficult for an adversary to invert the degradation performed by the system.
}
	\label{fig:overview_proposed}
\end{figure}

Cameras are ubiquitous, and they are increasingly connected to cloud services for streaming, post-processing, or inference purposes. The connectivity of such cameras leaves them prone to hacks or data leaks. While strong encryption is important in the digital sphere, we investigate the use of optical and analog hardware as a first layer of privacy.
A simple trick to ``encrypt'' an image at this initial layer is to modify the conventional measurement process, which normally attempts to capture an image as close as possible to the physical scene. 
To achieve visual privacy, previous work has corrupted the image formation process through: low-resolution measurements~\cite{7351605,10.5555/3298023.3298185}, replacing the lens with a multiplexing mask~\cite{9102956,9021989,wang2019privacy}, and modifying the quantization function~\cite{9102956}. By accounting for this degradation in the optimization of the downstream task (end-to-end training), robust performance can be obtained for a variety of tasks, \eg human activity, pose estimation, face classification~\cite{10.5555/3298023.3298185,9710270,shi2022loen,9102956}. While this approach can provide protection in the case of leaks in measurement data, protection in the event of camera parameter leaks has (to the best of our knowledge) not been investigated. 
For example, a leak in the camera's point spread function (PSF) makes it possible to recover an image of the underlying object using standard convex optimization techniques from computational imaging~\cite{boominathan2022recent}. 

In this work, we propose using a programmable mask with \textit{varying patterns} to provide protection to both data and system parameter leaks.
Using a programmable mask allows the system to adapt its encoding function, making it harder for an adversary to decode the normally visually-private measurement in the case of camera parameter leaks. Furthermore, a programmable mask allows for the system to be reconfigured to another inference task.
\cref{fig:overview_proposed} shows how our proposed system differs from a conventional lensed camera and previous lensless systems that use a fixed modulating mask in place of a lens.
Additionally, we propose a physical embodiment of our system that can be put together from cheap and accessible components, totaling at around $100$~USD. As a programmable mask, we use a low-cost liquid crystal display (LCD), which costs around $20$~USD.
We are the first to employ such a device in end-to-end optimization for computational optics. 
A differentiable digital twin with the selected LCD component is used to optimize the proposed system in simulation.
To show the robustness and versatility of our proposed system, we evaluate its performance on several classification tasks:  handwritten digit classification (MNIST~\cite{lecun1998mnist}), gender and smiling binary classification  (CelebA~\cite{liu2015faceattributes}), and object classification (CIFAR10~\cite{krizhevsky2009learning}). For all tasks, we find that learning the mask function (as opposed to using a heuristically designed or random encoding) is most resilient to decreases in the embedding dimension, which helps to enhance privacy.
To encourage reproducibility, we open source a wave propagation and camera simulator needed for end-to-end optimization,\footnote{\href{https://github.com/ebezzam/waveprop}{\texttt{github.com/ebezzam/waveprop}}; pip install waveprop
} the training software,\footnote{\href{https://github.com/ebezzam/LenslessClassification}{\texttt{github.com/ebezzam/LenslessClassification}}
} 
and library for interfacing with the baseline and proposed cameras.\footnote{\href{https://github.com/LCAV/LenslessPiCam}{\texttt{github.com/LCAV/LenslessPiCam}}; pip install lensless
}

\section{Related work}
\label{sec:related}

The work in this paper is at the intersection of multiple disciplines: privacy-preserving imaging, optical computing, and compressive imaging. In this section, we describe how our work differs from the relevant state-of-the-art in the respective fields.  \cref{sec:deepoptics} combines these elements to express our objective as a supervised learning task. 

\paragraph{Privacy though optical and analog operations.} Privacy-enhancement through degradations in the optical/analog domain has been implemented in previous works by:
downsampling at the sensor~\cite{7351605,10.5555/3298023.3298185,9102956}, using a coded-aperture mask~\cite{9021989,wang2019privacy}, and learning a multi-kernel mask~\cite{9102956,shi2022loen}. 
In the latter case, end-to-end optimization is used to jointly learn the mask function and the digital processing through supervised learning, and
the learned mask is either printed via lithography techniques or displayed on a spatial light modulator (SLM).
To make the trained system robust to adversarial attacks, a privacy term can be added to the loss, namely the performance of adversarial network is minimized while maximizing the performance of the desired task~\cite{9893382,9102956}. 
However, the above works assume the privacy-enhancing operations \textit{and adversarial attacks} are fixed. In~\cite{8805097}, the authors show that through known-plaintext attacks (obtaining encrypted versions of known inputs), the encoding mask(s) can be obtained to decrypt the visually-private image, raising in question the privacy of such systems. To achieve privacy-enhancing optical embeddings, we combine (1) optical and analog operations (learning a mask pattern and downsampling at the sensor as~\cite{9102956}) with (2) a programmable mask to protect against plaintext attacks and camera parameters leaks. 
\begin{figure*}[t!]
	\centering
	\includegraphics[width=0.7\linewidth]{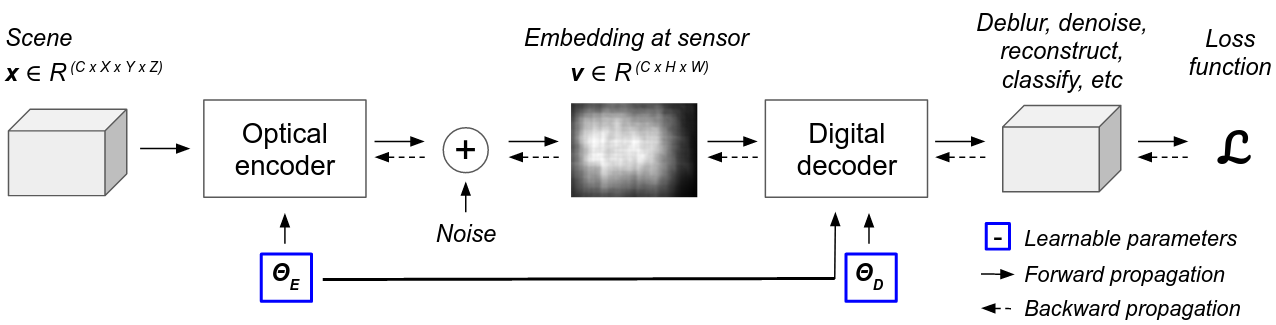} 
	\caption{Encoder-decoder perspective of cameras for end-to-end optimization. The scene could be four-dimensional (channel, height, width, depth) whereas the embedding measured at the sensor is at most three-dimensional (channel, height, width).}
	\label{fig:enc_dec}
\end{figure*}

\paragraph{Optical computing.} The use of optical components for data processing -- \emph{optical computing} -- is a rapidly evolving field, with most works attempting to replace digital operations with optical equivalents, \eg dot products~\cite{wang2022optical}, detection/correlation~\cite{1053650}, random projections~\cite{10.1109/ICASSP.2016.7472872}, and convolutional neural networks~\cite{chang2018hybrid}. While having the equivalent digital operation can be convenient, these approaches typically involve bulky, specialized setups that are not compatible or do not scale for mobile and edge device applications, \eg multiple lenses or expensive components such as SLMs (that cost a few thousand USD). Previous work has successfully shown the use of a low cost, LCD to perform optical operations: as a controllable aperture~\mbox{\cite{zomet2006}} and for single-pixel imaging~\mbox{\cite{huang2013}}. However, no work has attempted to use such low-cost components in an end-to-end fashion, nor in a compact design in the spirit of lensless cameras~\cite{boominathan2022recent}. In this work, the imaging system consists solely of an LCD placed at a short distance from the sensor, and end-to-end optimization is used to determine its pattern. 

\paragraph{Compressive imaging and classification.}

If the sensor resolution is reduced to the extreme, we arrive at a setup akin to single-pixel imaging~\cite{duarte2008}.
While single-pixel imaging requires multiple measurements illuminated with different patterns, mask-based imaging typically needs a single capture.
Nonetheless, both share a notion of \emph{sufficient statistics}~\cite{10.1117/12.714460,duarte2008} with regards to tasks such as detection/classification that do not require reconstruction. When reducing the sensor dimension of our proposed system to enhance privacy, we are also interested in the sufficient statistics that are necessary to provide robust classification by learning the best multiplexing from scene to sensor. 

\section{Deep optics for learning encoding}
\label{sec:deepoptics}

End-to-end approaches for optimizing optical components, also known as \emph{deep optics}~\cite{wetzstein2020inference}, is a recent trend enabled by improved fabrication techniques and the continual development of more powerful and efficient hardware and libraries for machine learning. It is motivated by faster and cheaper inference for edge computing 
and a desire to co-design the optics and the computational algorithm to obtain optimal performance for a particular application.

An encoder-decoder perspective is often used to frame such end-to-end approaches, casting the optics as the encoder and the subsequent computational algorithm as the decoder, as shown in \cref{fig:enc_dec}, and can be formulated as the following optimization problem minimized for a labeled dataset $ \{\bm{x}_i, \bm{y}_i\}_{i = 1}^N $:
\begin{equation}
	\hat{\bm{\theta}}_E, \hat{\bm{\theta}}_D = \argmin_{\bm{\theta}_E, \bm{\theta}_D} \sum_{i=1}^{N} \mathcal{L} \Big(\bm{y}_i, \underbrace{ D_{\bm{\theta}_E,\bm{\theta}_D} \big( \overbrace{O_{\bm{\theta}_E}  ( \bm{x}_i)}^{\text{embedding } \bm{v}_i} \big)}_{\text{decoder output }\bm{\hat{y}}_i}  \Big). \label{eq:optimization}
\end{equation}
$ \bm{\theta}_E, \bm{\theta}_D $ are the optical encoder and digital decoder parameters that we seek to optimize for a given task.
The optical encoder $ O_{\bm{\theta}_E}(\cdot)  $ encapsulates propagation in free space and through all optical components prior to the sensor, and downsampling and additive noise at the sensor. Given a scene $ \bm{x}_i $, it outputs the sensor embedding  $ \bm{v}_i $.
$ D_{\bm{\theta}_E,\bm{\theta}_D}(\cdot) $ is the digital decoder (\eg a neural network), which can perform a whole slew of tasks: deblurring, denoising, image reconstruction, classification, \etc. It can optionally make use of the optical encoder parameters, \eg the PSF can be useful for image deconvolution/reconstruction. The decoder's output is fed to an application-dependent loss function $ \mathcal{L}(\cdot) $ along with the ground-truth output $  \bm{y}_i $, and the error is backpropagated to update $ \bm{\theta}_E, \bm{\theta}_D $ .

\newcommand{\figsize}{0.19}
\newcommand{\multiimage}{0.18}

Within the context of privacy-preserving tasks, a common approach is to apply adversarial training~\cite{9207852,pittaluga2019learning,9102956,hinojosa2021learning,9893382}, namely simultaneously minimize a task-related loss such as \cref{eq:optimization} and maximize a privacy-related term, \eg the classification or reconstruction error of an adversarial network. Similar to~\cite{wang2019privacy,7351605,10.5555/3298023.3298185}, we choose not to incorporate an adversarial component in our training, since the latter would  only enhance privacy for the \emph{fixed adversarial attacks} it assumes, resulting in potential blind spots -- in particular for large datasets with complicated structures~\cite{zhang2018the}. Moreover, as we shall soon demonstrate, the design of our camera already produces such degenerate measurements that an attacker with knowledge about the camera system would struggle to reconstruct an image of the original scene. This is achieved by means of a (re)programmable mask component that randomly changes its configuration to mitigate potential privacy-infringing attacks on the system. 
Such a design is motivated by the findings of Jiao \etal \cite{8805097}, who show that decrypted images with the wrong mask appear noise-like and that no information about the actual image can be visually perceived or recovered. 
This is confirmed by \cref{fig:example_cvx_recon}, which shows reconstructions when a \textit{correct} and an \textit{incorrect}  mask are used for decoding simulated images with our system (top and bottom row respectively).

\section{Proposed system}
\label{sec:proposed}


\subsection{Hardware design}

\begin{figure}[t!]
	\centering
		\includegraphics[width=0.9\linewidth]{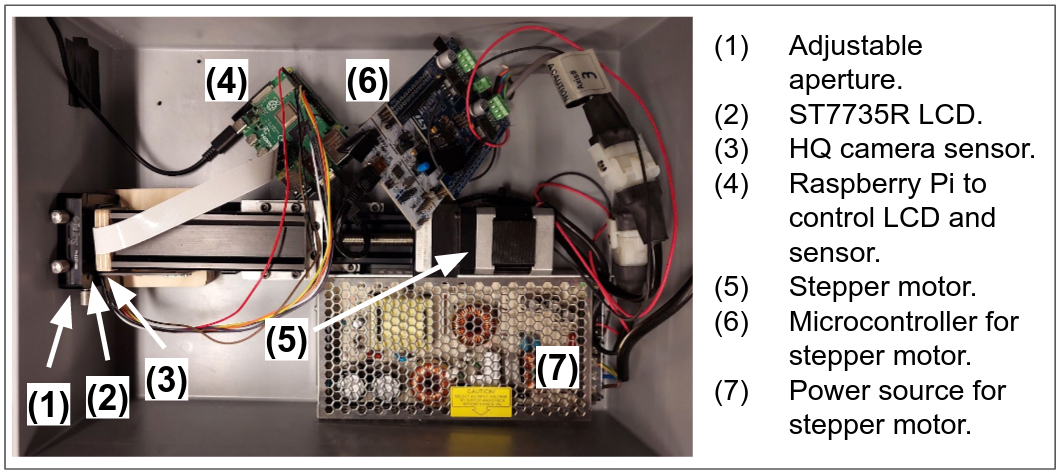}
		\caption{Experimental prototype of programmable, mask-based lensless camera.}
		\label{fig:prototype_labeled}
\end{figure}

Our camera design is motivated by the benefits of lensless cameras (compact, low-cost, visual privacy) and programmability.
To this end, a programmable mask serves as the only optical component in our encoder, specifically an off-the-shelf LCD driven by the ST7735R device which can be purchased for $\$20$.\footnote{\url{https://www.adafruit.com/product/358}} 
An experimental prototype of the proposed design can be seen in \cref{fig:prototype_labeled}. The LCD is wired to a Raspberry Pi ($\$35$) with the Raspberry Pi High Quality $12.3$ MP Camera ($\$50$) as a sensor, totaling our design to just $\$105$. The prototype includes an adjustable aperture, and a stepper motor for programmatically setting the distance between the LCD and the sensor, both of which can be removed to produce a more compact design.

\subsection{Digital twin for training}

End-to-end optimization requires a sufficiently accurate and differentiable simulation of the physical setup. 
Our digital twin of the imaging system
accounts for wave-based image formation for spatially incoherent, polychromatic illumination, as is typical of natural scenes. 
A simulation based on wave optics (as opposed to ray optics) is necessary to account for diffraction due to the small apertures of the mask and for wavelength-dependent propagation. 
To determine whether a wave-optics simulation is necessary, the Fresnel number $ N_F $ can be used, with ray optics generally requiring $N_F\gg 1$~\cite{boominathan2022recent} .
For our setup, $ N_F \in [1.2, 2] $ suggesting that diffraction effects need to be explicitly accounted for.\footnote{The Fresnel number is given by $ N_F = a^2 / d\lambda $, where $ a $ is the size of the mask's open apertures, $ d $ the propagation distance, and $ \lambda $ the wavelength. For our setup, $ a = \SI{0.06}{\milli\meter} $, $ d = \SI{4}{\milli\meter}$, and $ \lambda \in [\SI{450}{\nano\meter}, \SI{750}{\nano\meter}] $.}

In our setup, the scene of interest is at a fixed distance $d_1$ from the optical encoder, which itself is at a distance $d_2$ from the image plane. We adopt a common assumption from \textit{scalar diffraction theory}, namely that image formation is a linear shift-invariant (LSI) system between two parallel planes for a given wavelength $ \lambda $~\mbox{\cite{Goodman2005}}. Consequently, there exists an impulse response that can be convolved with the \emph{scaled} scene to obtain its image at a given distance $ z $:
\begin{align}
\label{eq:incoherent_main}
I_2(\bm{x}; z, \lambda) = \int_{\mathbb{R}^2}d\bm{r} \, p(\bm{x} -\bm{r} ; z, \lambda) \Bigg[ \frac{1}{|M|^2} I_0 \Big( \frac{\bm{r}}{M} ; \lambda\Big)\Bigg],
\end{align}
where $I_0$ and $I_2$ are the intensities at the scene and image planes respectively, $ \bm{x}\in \mathbb{R}^2 $ and $ \bm{r} \in \mathbb{R}^2$ are coordinates on the image and scene planes respectively, $M = - d_2 / d_1$ is a magnification/inversion factor, and $p$ is the intensity PSF~\cite{Goodman2005}.
As we have an incoherent illumination, the imaging is linear in intensity, lending to the above convolutional relationship.
Our digital twin modeling amounts to obtaining a PSF that encapsulates propagation from a given plane in the scene to the sensor plane. There are two ways to obtain this PSF: measuring it with a physical setup or simulating it.
Below we describe how it can be simulated for a mask-based encoder as in our proposed system.


\subsection{Simulating the PSF for a mask-based encoder}
\label{sec:sim_psf}

We model a programmable mask as a superposition of apertures for each adjustable pixel:
\begin{align}
\label{eq:mask_gen_main}
M(\bm{x}) = \sum_{k}^K w_{k} A(\bm{x} - \bm{x}_k),
\end{align}
where the complex-valued weights $\{w_{k}\}_{k=1}^K$ satisfy $|w_{k}| \leq 1$, the coordinates $ \{(\bm{x}_k)\}_{k=1}^{K} $ are the centers of the mask pixels, and the aperture function $A(\bm{x})$ is assumed to be identical for each pixel. The weights $\{w_{k}\}_{k=1}^K$ correspond to the mask pattern we would like to determine for a given task. 
However, the intensity PSF of the mask function, and not the mask function itself, is needed to obtain the embedding of a scene as given by \cref{eq:incoherent_main}. 
The intensity PSF is the squared magnitude of the amplitude PSF~\cite{Goodman2005}, namely $ p(\bm{x}; z, \lambda) =|h(\bm{x}; z, \lambda)|^2 $. The amplitude PSF for our setup can be modeled as convolution between (1) spherical waves impinging on the mask and (2) the free-space propagation kernel. Computationally, it is best performed in the spatial frequency domain (via FFTs) as
\begin{align}
\label{eq:output_main}
h(\bm{x}; z=d_1 + d_2, \lambda) = \mathcal{F}^{-1}\Big(& \mathcal{F} \Big( M(\bm{x}) e^{j \frac{2\pi}{\lambda} \sqrt{\|\bm{x}\|_2^2 +  d_1^2}}
\Big) \nonumber \\ &\times H(\bm{u}; z=d_2, \lambda) \Big),
\end{align}
where $ \mathcal{F}$ and $\mathcal{F}^{-1} $ denote the spatial Fourier transform and its inverse,  $ H(\bm{u}; z, \lambda)$ is the free-space propagation frequency response~\mbox{\cite{Matsushima:09}} (defined in \mbox{\cref{eq:freespace})}, and $\bm{u} \in \mathbb{R}^2$ are spatial frequencies of $\bm{x}$.
In \cref{sec:model_prop}, we describe in more detail the modeling of the PSF, programmable masks, and specifics for the ST7735R component.


\section{Experiments}
\label{sec:experiments}

\begin{figure*}[t!]
	\begin{subfigure}{0.16\linewidth}
		\centering
		\includegraphics[width=0.99\linewidth]{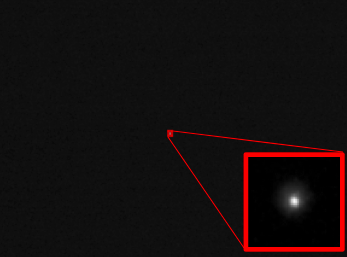} 
		\caption{Lens.}
		\label{fig:lens_zoom_section}
	\end{subfigure}
	\begin{subfigure}{0.16\linewidth}
		\centering
		\includegraphics[width=0.99\linewidth]{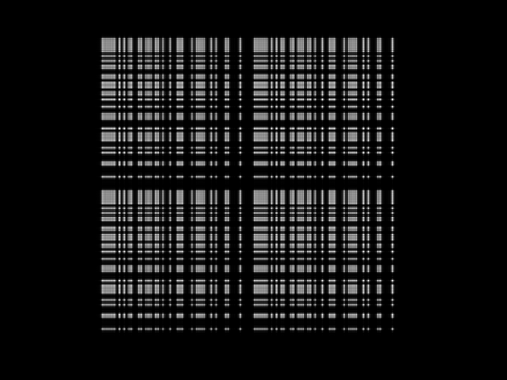} 
		\caption{Coded aperture.}
		\label{fig:ca_psf}
	\end{subfigure}
	\begin{subfigure}{0.16\linewidth}
		\centering
		\includegraphics[width=0.99\linewidth]{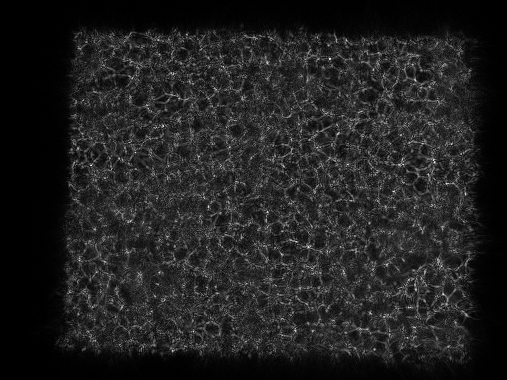}
		\caption{Diffuser.}
		\label{fig:diffuser_psf}
	\end{subfigure}
	\begin{subfigure}{0.16\linewidth}
		\centering
		\includegraphics[width=0.99\linewidth]{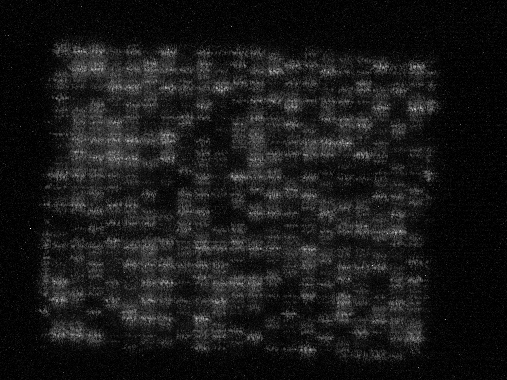}
		\caption{Fixed mask (m).}
		\label{fig:adafruit_psf}
	\end{subfigure}
	\begin{subfigure}{0.16\linewidth}
		\centering
		\includegraphics[width=0.99\linewidth]{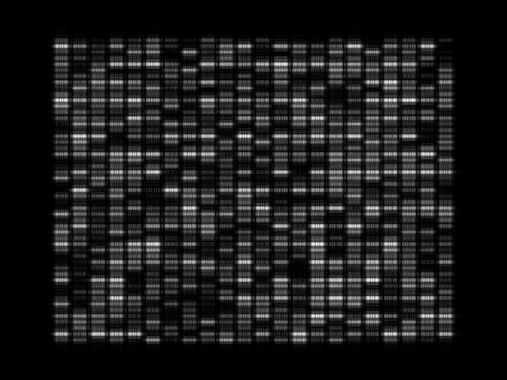} 
		\caption{Fixed mask (s).}
		\label{fig:adafruit_sim_psf}
	\end{subfigure}
	\begin{subfigure}{0.16\linewidth}
		\centering
		\includegraphics[width=0.99\linewidth]{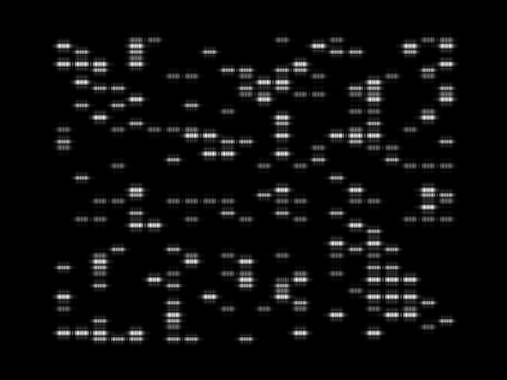} 
		\caption{Learned mask.}
		\label{fig:learned_mask_gender_768_psf}
	\end{subfigure}
	\caption{Point spread functions (PSFs) for baseline and proposed cameras. \textit{Learned mask} is unique for each embedding dimension, digital decoder, and task. The one shown in \cref{fig:learned_mask_gender_768_psf} was optimized for an embedding dimension of  $ (24\times 32) $, a two-layer fully connected neural network, and for gender classification. See \cref{sec:learned_psfs} for the learned PSFs for other sensor dimensions, digital decoders, and tasks.}
	\label{fig:psfs}
\end{figure*}

In this section, we apply our proposed camera and end-to-end optimization to various classification tasks. We optimize a classifier that directly operates on the camera measurement, rather than introducing an (unnecessary)  intermediate reconstruction step as in~\cite{8590781}. Consequently, the digital decoder $ D_{\bm{\theta}_D}(\cdot) $ in \cref{eq:optimization} does not require explicit information from the encoder.

For our proposed system with the ST7735R component, the learnable optical encoder parameters $\bm{\theta}_E$ for \cref{eq:optimization} are the mask weights $ \{w_k\}_{k=1}^{K} $ in \cref{eq:mask_gen_main}, which are restricted to be real and non-negative for amplitude modulation.

We conduct the following experiments to demonstrate:
\begin{enumerate}
	\tightlist
	\item \cref{sec:mnist,sec:celeba,sec:cifar10}: the viability of the proposed, low-cost system and end-to-end optimization of (1) the mask weights and (2) the digital classifier on multiple tasks: handwritten digits (MNIST~\cite{lecun1998mnist}), gender and smiling (CelebA~\cite{liu2015faceattributes}), and objects (CIFAR10~\cite{krizhevsky2009learning}) classification.
	\item \cref{sec:defense}: the proposed system's defense (low-dimensional measurements and variable mask patterns) against parameters leaks and plaintext attacks that try to recover the underlying image. 
\end{enumerate}
In our experiments, we consider various architectures for the digital decoder: logistic regression (LR), a fully-connected neural network with a single hidden layer of 800 units (FC), and VGG11~\cite{simonyan2014very}. Further details on model architectures, as well as training hardware and software, can be found in \cref{sec:arch}. Training hyperparameters, example images, and train-test accuracy curves for each experiment can be found in \cref{sec:mnist_vary_app,sec:mnist_robust_app,sec:celeba_app,sec:cifar10_app}.

\paragraph{Datasets.} For MNIST, we use the provided train-test split: $60'000$ training and $10'000$ test examples. 
Referring to \cref{eq:optimization}, the $ \{\bm{x}_i\}_{i = 1}^N $ coming from MNIST are $ (28\times 28) $ images of handwritten digits and $ \{\bm{y}_i\}_{i = 1}^N $ one-hot encoded vectors corresponding to the labels from $ 0 $ to $ 9 $. 

For CelebA, we use a subset of $100'000$ examples from the original dataset of $202'599$ images for training the optical encoder, allocating $85\%$ for training and $15\%$ for testing. This split results in a gender distribution of \SI{58}{\percent} female and \SI{42}{\percent} male, and \SI{52}{\percent} not smiling and \SI{48}{\percent} smiling in both the train and test set. Gender classification is chosen as it represents a global feature while smiling classification is localized. As these tasks are binary, the outputs are scalar, namely $ \{y_i\}_{i = 1}^N \in \{0, 1\}  $.
The original images are $ (178\times 218) $ and RGB, but we convert them to grayscale. For the privacy-preserving experiments in \cref{sec:defense}, we use a separate subset of $100'000$ examples.

For CIFAR10, we use the provided train-test split: $50'000$ training and $10'000$ test examples. 
The original images are $ (32\times 32) $ and RGB. As there are $ 10 $ classes, $ \{\bm{y}_i\}_{i = 1}^N $ are length-$ 10 $ one-hot encoded vectors. 

\paragraph{Baseline and proposed cameras.}

\begin{algorithm}[t!]
	\centering
	\caption{Simulating the optical encoder $ O_{\bm{\theta}_E}(\cdot) $ in \cref{eq:optimization}. See \cref{sec:simulation} for more detailed description.}
\label{alg:sim}
	\begin{algorithmic}[1]
		\Require Original image $\bm{x} \in \mathbb{R}^{C\times H\times W} $,
		PSF $ \bm{p} \in \mathbb{R}^{C\times H_{\text{PSF}}\times W_{\text{PSF}}}$, object height $ h_{\text{obj}} $, scene-to-encoder distance $ d_1 $, encoder-to-sensor distance $ d_2 $, downsampling factor $ D \geq 1 $, SNR $ \sigma $
		\Ensure At sensor $\bm{v} \in \mathbb{R}^{C\times DH_{\text{PSF}}\times DW_{\text{PSF}}} $ 
		\State$\bm{x}_{\text{scene}} \gets  \text{Prep}(\bm{x}, h_{\text{obj}}, d_1, d_2, \bm{p} ) $ \Comment{\textit{Pad image according to physical setup and sensor aspect ratio, and resize to PSF dimensions.}} 
		\State$\bm{v} \gets \bm{x}_{\text{scene}} \ast \bm{p} $ \Comment{\textit{Convolve each channel to obtain measurement at sensor.}}
		\If{$D > 1 $}
		\State$\bm{v} \gets \text{Downsample}(\bm{v}, D)$
		\EndIf
		\State$\bm{v} \gets \text{ShotNoise}(\bm{v}, \sigma)$
	\end{algorithmic}
\end{algorithm}

To produce the embeddings $ \{\bm{v}_i\}_{i = 1}^N $ in \cref{eq:optimization}, the original images $ \{\bm{x}_i\}_{i = 1}^N $ are simulated according to \cref{alg:sim}.
We compare six imaging systems: \textit{Lens}, \textit{Coded aperture}~\cite{flatcam}, \textit{Diffuser}~\cite{Antipa:18}, \textit{Fixed mask (m)}, \textit{Fixed mask (s)}, and \textit{Learned mask}.
\textit{Lens} shows how a non-visually-private system would perform, while \textit{Coded aperture} and \textit{Diffuser} represent typical mask design strategies: a binary amplitude modulation scheme that simplifies reconstruction and a phase modulation mask respectively. \textit{Fixed mask (m)} and \textit{Fixed mask (s)} correspond to the proposed imaging system discussed in \cref{sec:proposed}, but with a randomly-set mask pattern. \textit{Fixed mask (m)} is measured with hardware, while \textit{Fixed mask (s)} is a simulated to gauge how our digital twin compares with a measured PSF. \textit{Learned mask} corresponds to the proposed imaging system with learning the mask weights in \cref{eq:mask_gen_main}, \ie $\bm{\theta}_E = \{w_k\}_{k=1}^{K} $ for \cref{eq:optimization}. In contrast, all other investigated imaging systems are \emph{fixed optical encoders}, with \emph{frozen} parameters $\bm{\theta}_E$.   

For each camera, a PSF is needed to perform the simulation procedure described in \cref{alg:sim}.
The PSFs for \textit{Lens}, \textit{Diffuser}, and \textit{Fixed mask (m)} are measured with hardware, and the rest are simulated.
The learnable parameters of \textit{Learned mask} are updated at each batch to obtain a new simulated PSF, as described in \cref{sec:sim_psf}.
Further technical details, such as the components for the measured PSFs and parameters for the simulated PSFs, can be found in~\cref{sec:baseline}. The PSFs themselves are shown in \cref{fig:psfs}.


\subsection{MNIST proof-of-concept}
\label{sec:mnist}

\paragraph{Varying embedding dimension.}
\label{sec:vary_dimension}

\begin{table*}[t!]
	\caption{MNIST accuracy on test set, simulated accordingly for each camera and dimension. \textit{Relative drop} indicates relative drop in accuracy for each camera (for a given architecture) from $ (24\times32) $ to $ (3\times4) $.}
	\label{tab:mnist_vary_embedding}
	\centering
	\scalebox{0.93}{
		\begin{tabular}{lccccc  ccccc}
			\toprule
			\textit{Classifier} $\rightarrow $  & \multicolumn{5}{c}{Logistic regression} & \multicolumn{5}{c}{Two-layer fully connected, 800 hidden units}  \\
			\cmidrule(r){2-6} \cmidrule(r){7-11} 
			\textit{Embedding}   $\rightarrow $  &  24$\times$32  &12$\times$16 & 6$\times$8 & 3$\times$4 & \textit{Relative} & 24$\times$32   &12$\times$16 & 6$\times$8 &   3$\times$4 & \textit{Relative} \\
			\textit{Encoder}  $\downarrow $     &  =768 &=192   &=48   &  =12 &\textit{drop}& =768&=192   &=48   & =12 &\textit{drop}\\
			\midrule
			\textit{Lens}   &
			  $ \mathit{92.4}\% $  & $ \mathit{75.2}\% $& $ \mathit{41.7}\% $& \multicolumn{1}{c|}{$\mathit{21.9}\%$}  &  \multicolumn{1}{c|}{$\mathit{76.3\%}$}  &
			 $\mathit{98.4\%}$ &$ \mathit{83.8\%} $ & $ \mathit{40.7\%} $& \multicolumn{1}{c|}{$\mathit{22.5\%}$} & \multicolumn{1}{c|}{$\mathit{77.1\%}$}\\ \hdashline\noalign{\vskip 0.5ex}
			Coded aperture   &
			$ 82.4\% $  & $ 87.2\% $& $ 71.5\% $& \multicolumn{1}{c|}{$57.3\%$}  & \multicolumn{1}{c|}{$30.5\%$}&
			$97.7\%$ &$ 97.0\% $ & $ 92.2\% $& \multicolumn{1}{c|}{$71.2\%$} & \multicolumn{1}{c|}{$27.2\%$}\\
			Diffuser  &
			$ 91.7\% $  & $ 82.3\% $& $ 76.6\% $& \multicolumn{1}{c|}{$54.5\%$}  & \multicolumn{1}{c|}{$40.6\%$}&
			$98.6\%$ &$ 98.6\% $ & $ 97.0\% $& \multicolumn{1}{c|}{$78.9\%$} & \multicolumn{1}{c|}{$20.0\%$}\\
			Fixed mask (m)   &
			$ 92.8\% $  & $ 91.9\% $& $ 87.7\% $& \multicolumn{1}{c|}{$73.4\%$}  & \multicolumn{1}{c|}{$21.0\%$}&
			$\bm{98.7}\%$ &$ \bm{98.7}\% $ & $ 97.8\% $& \multicolumn{1}{c|}{$87.5\%$} & \multicolumn{1}{c|}{$11.4\%$}\\
			Fixed mask (s)   &
			$ 92.7\% $  & $ 91.8\% $& $ 86.5\% $& \multicolumn{1}{c|}{$72.7\%$}  & \multicolumn{1}{c|}{$21.6\%$}&
			$98.6\%$ &$ \bm{98.7}\% $ & $ 97.5\% $& \multicolumn{1}{c|}{$88.6\%$} & \multicolumn{1}{c|}{$10.2\%$}\\
			Learned mask  &
			$ \bm{95.5}\% $  & $ \bm{93.5}\% $& $ \bm{91.4}\% $& \multicolumn{1}{c|}{$\bm{79.3}\%$}  & \multicolumn{1}{c|}{$\bm{17.0}\%$}&
			$\bm{98.7}\%$ &$ 98.4\% $ & $ \bm{98.0}\% $& \multicolumn{1}{c|}{$\bm{91.7\%}$} & \multicolumn{1}{c|}{$\bm{7.10}\%$}\\
			\bottomrule
		\end{tabular}}
	\end{table*}

\cref{tab:mnist_vary_embedding} reports the best test accuracy for each optical encoder, for a varying sensor embedding dimension and for two classification architectures: LR and FC. While all approaches decrease in performance as the dimension reduces, \textit{Learned mask} is the most resilient as quantified by \textit{Relative drop} in \cref{tab:mnist_vary_embedding}. 
While the performance gap between \textit{Learned mask} and fixed lensless encoders decreases when a digital classifier with more parameters is used (FC), the benefits of learning this multiplexing are clear for a very low embedding dimension of $ (3 \times 4)$.  
\cref{tab:mnist_examples} shows example sensor embeddings for various camera and embedding dimensions pairs.

%
\paragraph{Robustness to common image transformations.}
\label{sec:robustness}

\begin{table}[t]
	\caption{Relative drop in MNIST accuracy for two-layer fully connected network with 800 hidden units and an embedding dimension of $ (24\times 32) $ for a randomly transformed data. See \cref{tab:mnist_robustness_abs} for absolute performance. Lower is better.}
	\label{tab:mnist_robustness}
	\centering
	\scalebox{0.93}{
		\begin{tabular}{l cccc}
			\toprule
			   &  Shift & Rescale   &Rotate   &  Perspective \\
			\midrule
			\textit{Lens} & 
			$ \mathit{12.1\%} $  & $ \mathit{13.5\%} $& $ \mathit{3.20\%} $  & 
			$\mathit{13.5\%}$ \\ \hdashline\noalign{\vskip 0.5ex}
			Coded aperture  &
			$ 77.2\% $  & $ 15.3\% $& $ 6.80\% $  & 
			$67.9\%$  \\
			Diffuser & 
			$ 67.8\% $  & $ 2.20\% $& $ \bm{3.00\%} $  & 
			$27.6\%$  \\
			Fixed mask (m)  &
			$ 56.7\% $  & $ 2.00\% $& $ 3.20\% $  & 
			$18.3\%$  \\
			Fixed mask (s)  &
			$ 61.6\% $  & $ \bm{1.90\%} $& $ 3.10\% $  & 
			$19.8\%$ \\
			Learned mask  &
			$ \bm{30.4\%} $  & $ 6.40\% $& $ \bm{3.00\%} $  & 
			$\bm{14.4\%}$ \\
			\bottomrule
		\end{tabular}}
	\end{table}


The MNIST dataset is size-normalized and centered, making it ideal for training and testing image classification systems but is not representative of how images may be taken in-the-wild. In this experiment, we evaluate the robustness of lensless encoders to common image transformation: shifting, rescaling, rotating, and perspective changes. 
The range of values for each transformation, and augmented examples can be found in \cref{sec:mnist_robust_app}.

\cref{tab:mnist_robustness} reports the relative drop in classification accuracy for each optical encoder and for each image transformation when using the FC architecture and an embedding dimension of $ (24 \times 32) $, namely when all models perform rather equivalently.
The main difficulty for lensless cameras is shifting as the whole sensor no longer captures multiplexed information on the entire sensor (see \cref{sec:mnist_shift} for example embeddings).
This leads to a significant reduction in classification accuracy for all lensless approaches. \textit{Learned mask} is able to cope with shifting much better than the fixed encoding strategies for lensing imaging, perhaps because it can adapt its multiplexing for such perturbations.
For rescaling and rotating, we observe no benefit in performance in learning the mask pattern. 
Moreover, the multiplexing characteristic of lensless cameras allows them to be more robust to rescaling than a classical lensed camera.

\begin{table*}[t!]
	\caption{Classification results for gender and smiling classification (CelebA~\cite{liu2015faceattributes}) and object classification (CIFAR10~\cite{krizhevsky2009learning}). Gender and smiling classification with CelebA use a two-layer fully-connected neural network with a hidden layer of 800 units as a digital classifier, while object classification with CIFAR10 uses VGG11. \textit{Relative drop} indicates relative drop in accuracy from $ (24\times32) $ to $ (3\times4) $. 
		}
	\label{tab:harder_tasks}
	\centering
	\scalebox{0.93}{
		\begin{tabular}{lccc  ccc cccc}
			\toprule
			\textit{Task} $\rightarrow $  & \multicolumn{3}{c}{Gender (FC)} & \multicolumn{3}{c}{Smiling (FC)} & \multicolumn{4}{c}{CIFAR10 (VGG11)} \\
			\cmidrule(r){2-4} \cmidrule(r){5-7} \cmidrule(r){8-11}
			\textit{Embedding}   $\rightarrow $  &  24$\times$32  & 3$\times$4 & \textit{Relative} &  24$\times$32  & 3$\times$4 & \textit{Relative} &  3$ \times $27$\times$36 & 3$ \times $13$\times$17& 3$ \times $6$\times$8   & 3$ \times $3$\times$4  \\
			\textit{Encoder}  $\downarrow $     &  =768 &=12   &  \textit{drop} & =768 &=12   &  \textit{drop} &  =2916 & =663 &=144 &=36  \\
			\midrule
			\textit{Lens}   & $\mathit{91.2\%}$ & \multicolumn{1}{c|}{$\mathit{71.1\%}$}  & \multicolumn{1}{c|}{$\mathit{22.1\%}$}  & $\mathit{89.3}\%$  & \multicolumn{1}{c|}{$\mathit{56.9\%}$} & \multicolumn{1}{c|}{$\mathit{36.3\%}$} & $\mathit{86.8\%}$  & $\mathit{79.8\%}$  & $\mathit{66.4\%}$ & $\mathit{46.5\%}$ \\ \hdashline\noalign{\vskip 0.5ex}
			Coded aperture   & $83.9\%$ & \multicolumn{1}{c|}{$61.8\%$}  & \multicolumn{1}{c|}{$26.3\%$}  & $84.2\%$  & \multicolumn{1}{c|}{$56.1\%$} & \multicolumn{1}{c|}{$33.4\%$} & $47.5\%$ & $51.5\%$ & $49.1\%$  & $38.6\%$ \\
			Diffuser   & $87.4\%$ & \multicolumn{1}{c|}{$67.8\%$}  & \multicolumn{1}{c|}{$22.4\%$}  & $86.1\%$  & \multicolumn{1}{c|}{$61.4\%$} & \multicolumn{1}{c|}{$28.7\%$} & $45.7\%$ & $45.3\%$ & $43.0\%$   & $39.1\%$ \\
			Fixed mask (m) & $89.9\%$ & \multicolumn{1}{c|}{$71.1\%$}  & \multicolumn{1}{c|}{$20.9\%$}  & $87.5\%$  & \multicolumn{1}{c|}{$63.4\%$} & \multicolumn{1}{c|}{$27.6\%$} & $51.3\%$  & $49.0\%$  & $44.7\%$  & $41.0\%$\\
			Fixed mask (s)& $90.9\%$ & \multicolumn{1}{c|}{$68.9\%$}  & \multicolumn{1}{c|}{$24.2\%$}  & $87.7\%$  & \multicolumn{1}{c|}{$63.1\%$} & \multicolumn{1}{c|}{$28.0\%$} & $50.8\%$ & $48.3\%$& $46.8\%$ & $38.9\%$  \\
			Learned mask & $\bm{91.0}\%$ & \multicolumn{1}{c|}{$\bm{80.7}\%$}  & \multicolumn{1}{c|}{$\bm{11.3}\%$}  & $\bm{88.4}\%$  & \multicolumn{1}{c|}{$\bm{76.0}\%$} & \multicolumn{1}{c|}{$\bm{14.0}\%$} & $\bm{63.0}\%$  & $\bm{61.6\%}$& $\bm{56.9\%}$ & $\bm{48.2}\%$  \\
			\bottomrule
	\end{tabular}}
\end{table*}

\subsection{Face attribute classification (CelebA)}
\label{sec:celeba}

For this experiment, we apply the baseline and proposed cameras to a dataset and task that may benefit from privacy-enhancing features: face classification. Precisely, we target two tasks -- gender and smiling binary classification -- a global and local feature respectively.
The left and middle sections of \cref{tab:harder_tasks} report the best test accuracy for both tasks and for two embedding dimensions -- (24$ \times $32) and (3$ \times $4). Out of the lensless approaches, \textit{Learned mask} is best performing.
Similar to MNIST, the relative drop in accuracy -- from (24$ \times $32) to (3$ \times $4) -- again shows how \textit{Learned mask} is the most robust in performance if one is interested in improving the privacy by decreasing the sensor resolution. \cref{tab:celeba_examples} shows example sensor embeddings as the dimension decreases. 

\subsection{RGB object classification (CIFAR10)}
\label{sec:cifar10}

For this experiment, we address a more complicated task: object classification from RGB data. To this end, we apply a more powerful model at the digital end: VGG11. The rightmost section of \cref{tab:harder_tasks} reports the best test accuracy for a varying sensor embedding. Among the lensless approaches, \textit{Learned mask} performs the best. However, the performance gap with the non-private \textit{Lens} is large, which may need to be addressed with a more potent optical encoder or digital classifier. The confusion matrices in \cref{fig:cifar_confusion} show which classes lensless encoders struggle to distinguish, most notably \textit{cats-dogs}. \cref{tab:cifar10_examples} shows example sensor embeddings as the dimension decreases.

\subsection{Defense against leaks and plaintext attacks}
\label{sec:defense}

While the multiplexing property of lensless cameras combined with lowering the embedding dimension appear to enhance the \textit{visual} privacy of the embeddings, it does not imply that a computational imaging technique cannot recover a discernible image.
In the following experiments, we assume that an adversary has obtained a set of visually private measurements from our proposed system. We demonstrate and quantify how the joint effect of (1) lowering the sensor resolution and (2) using a programmable mask enhances the privacy of the resulting embeddings.

\paragraph{Leak of optical encoder parameters.}

\begin{figure}[t!]
	\centering
	
	\begin{subfigure}{\figsize\linewidth}
		\centering
		\includegraphics[width=0.99\linewidth]{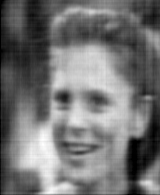} 
		\caption{$ 384\times 512 $}
		\label{fig:celeba_recovered_scene2mask0.55_height0.27_384x512_1}
	\end{subfigure}
	\begin{subfigure}{\figsize\linewidth}
		\centering
		\includegraphics[width=0.99\linewidth]{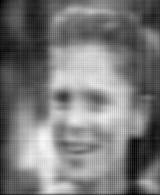} 
		\caption{$ 192\times 256 $}
		\label{fig:celeba_recovered_scene2mask0.55_height0.27_192x256_1}
	\end{subfigure}
	\begin{subfigure}{\figsize\linewidth}	
		\centering
		\includegraphics[width=0.99\linewidth]{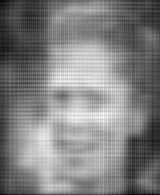}
		\caption{$ 96\times 128 $}
		\label{fig:celeba_recovered_scene2mask0.55_height0.27_96x128_1}
	\end{subfigure}
	\begin{subfigure}{\figsize\linewidth}
		\centering
		\includegraphics[width=0.99\linewidth]{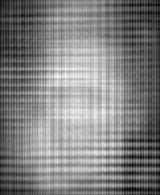} 
		\caption{$ 48\times 64 $}
		\label{fig:celeba_recovered_scene2mask0.55_height0.27_48x64_1}
	\end{subfigure}	
	\begin{subfigure}{\figsize\linewidth}
		\centering
		\includegraphics[width=0.99\linewidth]{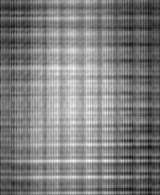}
		\caption{$ 24\times 32 $}
		\label{fig:celeba_recovered_scene2mask0.55_height0.27_24x32_1}
	\end{subfigure}\\
	
	\begin{subfigure}{\figsize\linewidth}
		\centering
		\includegraphics[width=0.99\linewidth]{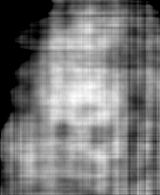} 
		\caption{$ 384\times 512 $}
		\label{fig:celeba_recovered_scene2mask0.55_height0.27_384x512_diff_slm_1}
	\end{subfigure}
	\begin{subfigure}{\figsize\linewidth}
		\centering
		\includegraphics[width=0.99\linewidth]{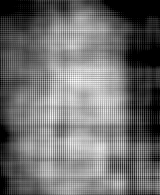} 
		\caption{$ 192\times 256 $}
		\label{fig:celeba_recovered_scene2mask0.55_height0.27_192x256_diff_slm_1}
	\end{subfigure}
	\begin{subfigure}{\figsize\linewidth}
		\centering
		\includegraphics[width=0.99\linewidth]{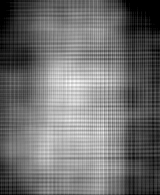}
		\caption{$ 96\times 128 $}
		\label{fig:celeba_recovered_scene2mask0.55_height0.27_96x128_diff_slm_1}
	\end{subfigure}
	\begin{subfigure}{\figsize\linewidth}
		\centering
		\includegraphics[width=0.99\linewidth]{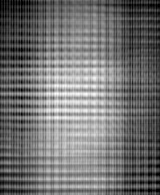} 
		\caption{$ 48\times 64 $}
		\label{fig:celeba_recovered_scene2mask0.55_height0.27_48x64_diff_slm_1}
	\end{subfigure}
	\begin{subfigure}{\figsize\linewidth}
		\centering
		\includegraphics[width=0.99\linewidth]{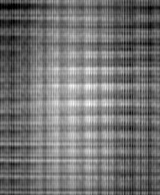}
		\caption{$ 24\times 32 $}
		\label{fig:celeba_recovered_scene2mask0.55_height0.27_24x32_diff_slm_1}
	\end{subfigure}

	\caption{Discerning content from lensless camera raw measurements is next to impossible, motivating privacy-preserving imaging with such cameras (see example measurements in \cref{tab:celeba_examples}).
		However, with sufficient knowledge about the camera, \eg a point spread function (PSF), and an appropriate computational algorithm, one is able to recover an estimate of the underlying object (top row, see \cref{sec:cvx} for inverse problem formulation).
		As the raw sensor measurement is increasingly under-sampled, it becomes more and more difficult to recover a meaningful estimate of the underlying object. The caption indicates the resolution of the raw measurement, while all recovered images are at the resolution of the PSF -- $ (384 \times 512) $.
		Bottom row demonstrates the recovery estimate when an incorrect PSF is used, motivating our use of a programmable mask to defend against leaks of camera system parameters.
	}
	\label{fig:example_cvx_recon}
\end{figure}

Given a lensless camera measurement, a malicious user could formulate an inverse problem in order to recover the underlying scene (see \cref{sec:cvx} for formulation). Such an attack would require an estimate of the PSF which could be obtained two ways: (1) there was a leak in mask values and an estimate PSF could be simulated using the approach described in \cref{sec:sim_psf} or (2) through a plaintext attack that reveals the mask function directly at the sensor. Note that the latter is less realistic since it requires that the adversary control the lighting in which the camera is placed (to emit a single point source) and that they can measure the response at a sufficiently high resolution.

Assuming that the malicious user is able to obtain an accurate estimate of the PSF, we can observe in \cref{fig:example_cvx_recon} reconstructed images as the sensor resolution decreases: from (384$ \times $ 512) in \cref{fig:celeba_recovered_scene2mask0.55_height0.27_384x512_1} down to (24$\times $32) in \cref{fig:celeba_recovered_scene2mask0.55_height0.27_24x32_1}. As the resolution decreases, the quality of the recovered scene deteriorates: at (24$\times $32) it is impossible to discern a face. In the last row of \cref{fig:example_cvx_recon}, we emulate the scenario that the adversary uses a wrong estimate of the PSF: either the mask values were reconfigured since the leak, or the adversary simply guessed the mask values when simulating the PSF. We can observe the recovered scenes resemble noise (similar to the findings of \cite{8805097} when trying to perform single-pixel imaging with the wrong illuminations).

\begin{table}[t!]
	\centering
	\caption{Image quality metrics (PSNR/SSIM) of reconstruction via convex optimization for varying sensor measurement resolution and when using a bad PSF estimate  (see \cref{sec:cvx} for inverse problem formulation). Results are averaged over $ 50 $ files. Higher is better. Yellow hightlight indicates a discernible image, see \cref{fig:example_cvx_recon} for examples. \SI{47}{\percent} / \SI{55}{\percent} drop in metrics for $(384\times 512)$.}
	\label{tab:cvx_attack}
	\scalebox{0.73}{
		\begin{tabular}{cccccc}
			\toprule
			\textit{PSF}& 384$\times $512 & 192$\times $256 & 96$ \times$128 & 48$ \times $64 & 24$ \times $ 32 \\  
			\midrule
			Good & \colorbox{yellow}{$ 22.6 $ / $ 0.80 $}  & \colorbox{yellow}{$ 19.7 $ / $ 0.67 $}   & \colorbox{yellow}{$16.0  $ / $0.52 $} & $12.9 $ / $ 0.26 $   & $ 11.7 $ / $ 0.23 $ \\ 
			Bad &  $ 11.9 $ / $ 0.36 $ & $ 11.7  $ / $ 0.25 $& $ 12.1  $ / $ 0.28 $ & $ 11.9 $ / $ 0.20 $  &  $ 11.4 $ / $ 0.22 $\\ 
			\bottomrule 
			
		\end{tabular}
	}
\end{table}

\cref{tab:cvx_attack} quantifies the quality of reconstruction for varying embedding dimensions and knowledge of the PSF. When a good estimate of the PSF can be obtained, the multiplexing property of lensless cameras may not be enough to preserve privacy, but can be when coupled with a decrease in sensor resolution. With a bad estimate of the PSF, it is not possible to recover a meaningful image of the underlying scene. Through the use of a programmable mask, the system can protect against leaks in mask values, and even allow imaging at higher resolutions if the task at hand requires it.

\paragraph{Plaintext attacks on compromised camera.}

\newcommand{\celebfig}{0.18}
\newcommand{\newlineceleb}{3pt}
\begin{figure}[t!]
	\centering
\stackunder[4pt]{Original}{\includegraphics[width=\celebfig\linewidth,valign=m]{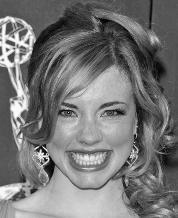}}
\stackunder[6pt]{Fixed}{\includegraphics[width=\celebfig\linewidth,valign=m]{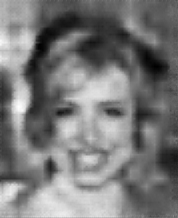}}
\stackunder[6pt]{10 masks}{\includegraphics[width=\celebfig\linewidth,valign=m]{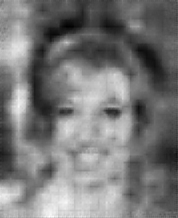}}
\stackunder[6pt]{100 masks}{\includegraphics[width=\celebfig\linewidth,valign=m]{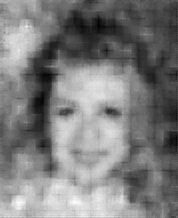}} \\[\newlineceleb]

\includegraphics[width=\celebfig\linewidth,valign=m]{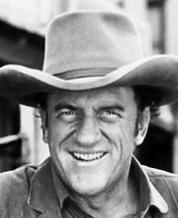} \includegraphics[width=\celebfig\linewidth,valign=m]{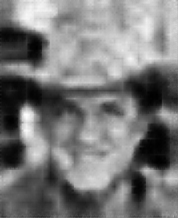}
\includegraphics[width=\celebfig\linewidth,valign=m]{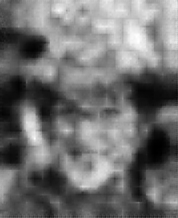}
\includegraphics[width=\celebfig\linewidth,valign=m]{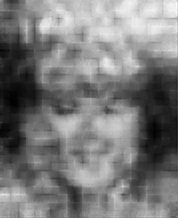}\\[\newlineceleb]

\includegraphics[width=\celebfig\linewidth,valign=m]{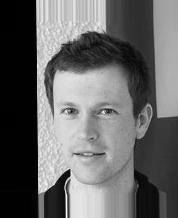} \includegraphics[width=\celebfig\linewidth,valign=m]{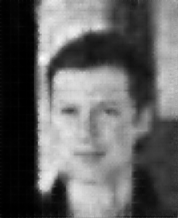}
\includegraphics[width=\celebfig\linewidth,valign=m]{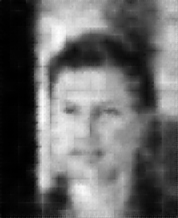}
\includegraphics[width=\celebfig\linewidth,valign=m]{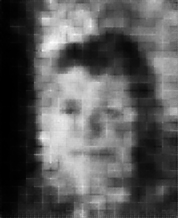}\\

	\caption{Example outputs of a decoder that was trained with $ 100'000 $ plaintext attacks of embeddings of resolution $ (24\times 32) $. See \cref{sec:example_generated} for more example outputs.}
	\label{fig:celeba_decoder}
\end{figure}

\begin{table}[t!]
	\caption{Image quality (PSNR/SSIM) of trained decoder for varying number of plaintext attacks and number of varying masks. Higher is better. \SI{17}{\percent} / \SI{26}{\percent} drop in metrics for $ 100'000 $.}
	\label{tab:celeba_decoder}
	\centering
	\scalebox{0.9}{
		\begin{tabular}{l c  c c}
			\toprule
			\textit{\# attacks}   $\downarrow $ & Fixed mask & 10 masks & 100 masks \\
			\midrule
			100    & $ 13.9 $  / $ 0.26 $ & $12.5 $  / $  0.21 $& $11.8$ / $0.20$  \\
			$ 1'000 $  & $ 16.3 $  / $ 0.40$ & $ 14.2$  / $ 0.32 $&  $13.4$ / $0.29$  \\
			$ 10'000 $   & \colorbox{yellow}{$ 18.1 $  / $ 0.53 $} & $16.2$  / $ 0.43 $& $ 14.6$ / $0.38 $  \\
			$ 100'000 $  & \colorbox{yellow}{$ 19.3 $  / $ 0.61 $} & \colorbox{yellow}{$ 18.0$  / $ 0.53 $}& $16.1$ / $0.45 $  \\
			\bottomrule
	\end{tabular}}
\end{table}

If obtaining an estimate of the PSF proves to be too difficult, an adversary could attempt to learn a decoder through a series of plaintext attacks, namely collect the corresponding embeddings for a set of known images. For this scenario, the adversary should (1) know the data that the system is configured to encode, and (2) be able to obtain the sensor embeddings of their set of known images.
A defense against a series of such plaintext attacks could be to vary the mask pattern, \eg with a programmable mask as in our proposed system, so that the dataset collected from the adversary is a combination of embeddings from different projections rather than just a single fixed projection.  
Note that at inference a different classifier would need to be used as the mask and classifier are jointly trained.  
The hypothesis is that learning a decoder for different projections is more difficult, as the decoder needs to learn the high entropic distribution of multiple masks.

In \cref{tab:celeba_decoder} we present results that confirm this hypothesis. For a sensor resolution of (24 $ \times $ 32) -- where the previous attack failed to produce an image where a face could be discerned (see \cref{fig:celeba_recovered_scene2mask0.55_height0.27_24x32_1}) -- we train a decoder for a varying number of plaintext attacks and number of masks used at capture. We train a convolutional network similar to that of generative adversarial networks~\cite{karras2017progressive},\footnote{We also tried fine-tuning a pre-trained StyleGAN2~\cite{Karras2020ada} but found the generated images were better when training a regressor from scratch.}
see \cref{sec:generator} for architecture and training details.

In the previous experiment, the identity could be discerned for a PSNR and SSIM of $ 16.0 $ and $ 0.52 $ respectively (resolution of $( 96\times 128) $ as seen in \cref{fig:celeba_recovered_scene2mask0.55_height0.27_96x128_1}). The scenarios with similar (or better) metrics are highlighted in yellow. As the number of varying masks increases ($ 1 $ to $ 100 $), we see that it becomes increasingly difficult for the decoder to produce a meaningful image. More plaintext attacks are needed which becomes impractical for an adversary to collect. \cref{fig:celeba_decoder} shows examples of reconstructed outputs. 
 

\section{Conclusion}
\label{sec:conclusion}

We have proposed a low-cost, privacy-enhancing system for performing lensless imaging with an off-the-shelf LCD component. Leveraging an end-to-end optimization procedure to jointly learn the mask pattern and a reconstruction-free classifier, we have demonstrated its ability to outperform previous approaches that use heuristic or random mask patterns. A notable feature of the proposed system is the use of a programmable mask to mitigate data and camera parameter leaks, and therefore enhance privacy. By varying the mask pattern, the system can thwart an adversary that attempts to invert the lensless encoding.
As future work, it would be interesting to investigate how the proposed programmable mask strategy thwarts adversarial attacks that seek to identify attributes, \eg ethnicity and age group, rather than reconstructing an image.

\paragraph{Limitations.} 
A single layer in the optical encoder limits its abilities to learn rich embeddings. Increasing the number of layers could improve performance~\cite{lin2018all}, but at the expense of a less compact system and a more involved learning procedure.
Moreover, relying on physical devices for computation can have drawbacks. They are more susceptible to degradation (due to usage and over time) than purely digital computations. Finally, device tolerances can lead to unwanted differences between two seemingly identical setups. Such differences may be more prominent for low-cost components such as the cheap LCD used in this paper, as opposed to commercial SLMs. 

\paragraph{Funding.}
	
	This work was in part funded by the Swiss National Science Foundation (SNSF) 
	under grants 200021\textunderscore181978/1 ``SESAM - Sensing and Sampling: Theory and Algorithms''
	(E.~Bezzam) and CRSII5 193826 ``AstroSignals - A New Window on the Universe, with 
	the New Generation of Large Radio-Astronomy Facilities'' (M.~Simeoni).


{\small
\bibliographystyle{ieee_fullname}
\bibliography{references}
}

\newpage
~
\newpage

\appendix

\renewcommand\thefigure{\thesection.\arabic{figure}}  
\setcounter{figure}{0}   
\renewcommand{\theequation}{\thesection.\arabic{equation}}
\setcounter{equation}{0}   
\renewcommand{\thetable}{\thesection.\arabic{table}}
\setcounter{table}{0}


\section{Digital twin modeling} 
\label{sec:model_prop}

\begin{figure*}[t!]
	\centering
	\begin{subfigure}[b]{.5\textwidth}
		\centering
		\includegraphics[width=0.99\linewidth]{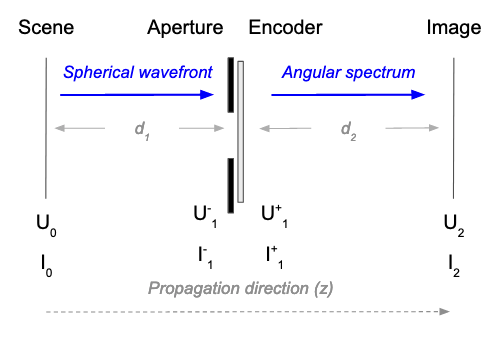} 
		\caption{}
		\label{fig:propagation}
	\end{subfigure}
	\hfill
	\begin{subfigure}[b]{.4\textwidth}
		\centering
		\includegraphics[width=0.99\linewidth]{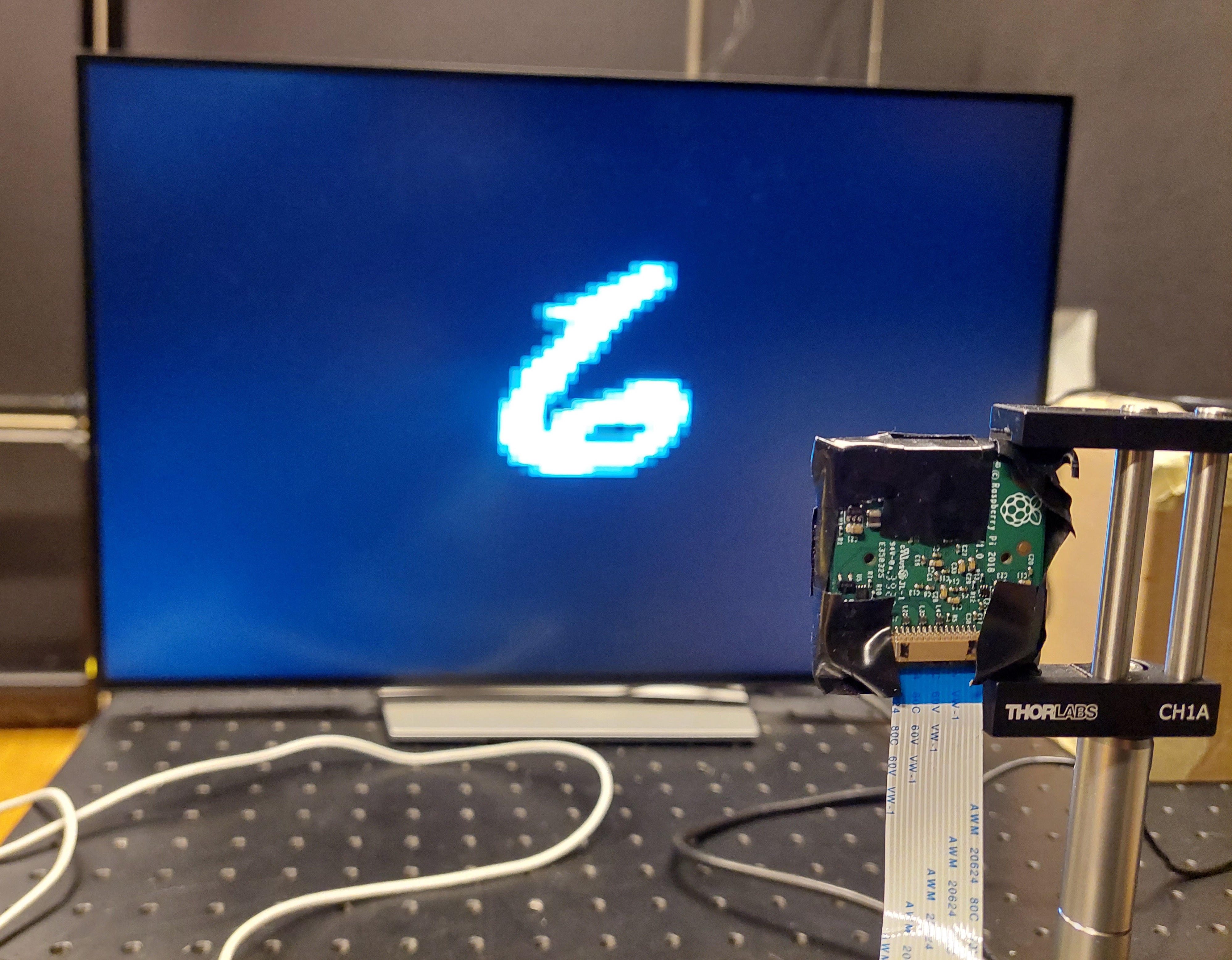}
		\caption{}
		\label{fig:measurement_setup}
	\end{subfigure}
	\caption{(a) Propagation setup. Not drawn to scale for visualization purposes. (b) Example physical measurement setup.}
\end{figure*}

\cref{fig:propagation} illustrates our physical setup: the scene of interest is at a fixed distance $d_1$ from the optical encoder, which itself is at a distance $d_2$ from the image plane.

\subsection{Convolutional relationship} 
\label{sec:conv}

We adopt a common assumption from 
\textit{scalar diffraction theory}, namely that image formation is a linear shift-invariant (LSI) system~\cite{Goodman2005}. This implies that there exists an impulse response, \ie a point spread function (PSF), that can be convolved with the input scene to obtain the output image. In Fourier optics, this convolutional relationship is between the  \emph{scaled} scene and its image at a distance $ z $, namely 
\begin{equation}
\label{eq:coherent}
U_2(\bm{x}; z, \lambda) = \int_{\mathbb{R}^2}d\bm{r}\, h(\bm{x} - \bm{r}; z, \lambda) \Bigg[ \frac{1}{|M|} U_0 \Big( \frac{\bm{r}}{M} ; \lambda\Big)\Bigg],
\end{equation}
where $U_0$ and $U_2$ are the wave fields, \ie complex amplitudes, at the scene and image planes respectively, $ \bm{x}\in \mathbb{R}^2 $ and $ \bm{r} \in \mathbb{R}^2$ are coordinates on the image and scene planes respectively, $h$ is the PSF, and $M = - d_2 / d_1$ is a magnification factor that also accounts for inversion~\cite{Goodman2005}. Note that this convolution is dependent on the wavelength $ \lambda $.

This LSI assumption significantly reduces the computational load for simulating optical wave propagation, as the convolution theorem and the fast Fourier transform (FFT) algorithm can be used to efficiently evaluate the wave field at the image plane via the spatial frequency domain. With the scaled Fourier transform, or equivalently chirp Z-transform, one can overcome the sampling limitations of the FFT and obtain the wave field at arbitrary resolutions~\cite{Muffoletto:07,doi:10.1137/21M1448641}. An aperture, or some form of cropping, helps to enforce the LSI assumption in order to avoid new patterns from emerging at the sensor for lateral shifts at the scene plane. This is particularly necessary for encoders with a large support, \eg those of lensless cameras.

For lenses, the PSF in \cref{eq:coherent} can be approximated by the Fraunhofer diffraction pattern (scaled Fourier transform) of the aperture function. 
For an arbitrary mask, the simplifications resulting from a lens are not possible and a more exact diffraction model is needed to predict the image pattern, \eg Fresnel propagation or the angular spectrum method~\cite{Goodman2005}. We employ the bandlimited angular spectrum method (BLAS) which produces accurate simulations for both near- and far-field~\cite{Matsushima:09}.

\paragraph{Incoherent, polychromatic illumination.} The convolutional relationship in \cref{eq:coherent} is for coherent illumination, \eg coming from a laser. However, illumination from natural scenes typically consists of diffuse or extended sources which are considered to be \textit{incoherent}. In such cases, impulses at the image plane vary in a statistically independent fashion, thus requiring them to be added on an intensity basis~\cite{Goodman2005}. In order words, for incoherent illumination, the convolution of \cref{eq:coherent} should be expressed with respect to intensity:
\begin{equation}
\label{eq:incoherent}
I_2(\bm{x}; z, \lambda) = \int_{\mathbb{R}^2}d\bm{r}\, p(\bm{x} - \bm{r}; \lambda) \Bigg[ \frac{1}{|M|^2} I_0 \Big( \frac{\bm{r}}{M} ; \lambda\Big)\Bigg],
\end{equation}
where $I_0$ and $I_2$ are the intensities at the scene and image planes respectively, with image intensity defined as the average instantaneous intensity:
\begin{equation}
I(\bm{x}; z,\lambda) = \mathbb{E} \big[ | U(\bm{x}; z,\lambda, t) |^2  \big]   ,
\end{equation}
and the intensity PSF is equal to the squared magnitude of the amplitude PSF~\cite{Goodman2005}, \ie of \cref{eq:coherent}:
\begin{equation}
\label{eq:intensity_psf}
p(\bm{x}; z, \lambda) = | h(\bm{x}; z, \lambda)|^2.
\end{equation}

For polychromatic simulation, each wavelength has to be simulated independently. Converting this multispectral data to RGB is typically done in two steps: (1) mapping each wavelength to the XYZ coordinates defined by the International Commission on Illumination\footnote{\url{https://en.wikipedia.org/wiki/CIE_1931_color_space}} and (2)  converting to red-green-blue (RGB) values based on a reference white.\footnote{\url{http://www.brucelindbloom.com/index.html?Eqn_RGB_XYZ_Matrix.html}}

\subsection{Point spread function modeling of a mask-based encoder}
\label{sec:psf_modeling}

We model a programmable mask as a superposition of apertures for each adjustable pixel:
\begin{align}
	\label{eq:mask_gen_app}
	M(\bm{x}) = \sum_{k}^K w_{k} A(\bm{x} - \bm{x}_k),
\end{align}
where the complex-valued weights $\{w_{k}\}_{k=1}^K$ satisfy $|w_{k}| \leq 1$, the coordinates $ \{(\bm{x}_k)\}_{k=1}^{K} $ are the centers of the mask pixels, and the aperture function $A(\bm{x})$ is assumed to be identical for each pixel. The weights $\{w_{k}\}_{k=1}^K$ correspond to the mask pattern we would like to determine for a given task. 

Our modeling of the PSF for propagation \textit{through} the mask is similar to that of~\cite{sitzmann2018}, namely for each wavelength $ \lambda $, we simulate the propagation (see \cref{fig:propagation} for the wave field variable at different planes):
\begin{enumerate}
	\item \textit{From the scene to the optical element}: propagation is modeled by spherical wavefronts. Assuming a point source at the scene plane $ U_0 $,  we have the following wave field at the aperture plane:
	\begin{equation}
		U_1^-(\bm{x}; z=d_1, \lambda) = \exp \Big(j \frac{2\pi}{\lambda} \sqrt{\|\bm{x}\|_2^2 +  d_1^2} \Big),
	\end{equation}
	where $ d_1 $ is the distance between the scene and the camera aperture.
	\item\textit{At the optical element}: the wave field is multiplied with the mask pattern:
	\begin{equation}
		\label{eq:after_mask}
		U_1^+(\bm{x}; z=d_1, \lambda) = U_1^-(\bm{x}; z=d_1,\lambda)  M(\bm{x}).
	\end{equation}
	Note that an infinitesimally small distance is assumed between the opening of the aperture $ U_1^- $ and the exit of the mask $ U_1^+ $.
	\item \textit{To the sensor}: free-space propagation according to scalar diffraction theory~\cite{Goodman2005} as light is diffracted by the optical element. 
	This yields the following wave field at the sensor plane:
	\begin{align}
		\label{eq:output}
		U_2(\bm{x}; z=d_1 + d_2, \lambda) = \mathcal{F}^{-1}  \Big(& \mathcal{F} \big( U_1^+(\bm{x}; z=d_1, \lambda) \big) \nonumber\\
		&  H(\bm{u}; z=d_2, \lambda) \Big),
	\end{align}
	where $ \mathcal{F}$ and $\mathcal{F}^{-1} $ denote the spatial Fourier transform and its inverse,  $\bm{u} \in \mathbb{R}^2$ are spatial frequencies of $\bm{x}$, and the free-space frequency response 
	according BLAS is given by:
	\begin{align}
		\label{eq:freespace}
		H(\bm{u}; z=d_2, \lambda ) = e^ {j \frac{2 \pi}{\lambda} d_2 \sqrt{1 - \|\lambda \bm{u}\|_2^2} } \,\text{rect2d}\Big(\frac{\bm{u}}{2\bm{u}_{\text{limit}}}\Big),
	\end{align}
	where $ \text{rect2d} $ is a 2D rectangular function for bandlimiting the frequency response and the bandlimiting frequencies are given by
	\begin{align}
		\bm{u}_{\text{limit}} = \dfrac{\sqrt{(d_2 / \bm{S} )^2  + 1} }{\lambda},
	\end{align}
	where $ \bm{S} \in \mathbb{R}^2 $ are the physical dimensions of the propagation region, in our case the physical dimensions of the sensor.
	\item \textit{Intensity at sensor}: as we are simulating incoherent light, we require the squared modulus of the wave field:
	\begin{equation}
		p(\bm{x}; z = d_1 + d_2, \lambda) = \big| U_2(\bm{x}; z=d_1 + d_2, \lambda)  \big|^2.
	\end{equation}
	
	
\end{enumerate}

\subsection{More on modeling a programmable mask} 
\label{sec:modeling_slm}
A key component in the above PSF simulation is modeling the complex-valued mask $M(\bm{x})$ associated with the programmable mask, which could be an LCD as in our setup or a spatial light modulator (SLM). Two assumptions are commonly made in its modeling:
\begin{itemize}
	\item The mask is assumed to be either a phase transformation, \ie $|M(\bm{x})| = 1$, or an amplitude transformation, \ie $M(\bm{x}) \in [0, 1]$. 
	\item The mask is discretized according to its resolution. This approximation neglects \emph{deadspace} (or equivalently the fill factor) of individual pixels, namely the regions of the mask that are not programmable.
\end{itemize}


Our modeling of the programmable mask in \cref{eq:mask_gen_app} takes into account deadspace, whereas others simply discretize the mask according to its resolution~\mbox{\cite{Peng:2020:NeuralHolography}}.
However, this model assumes that no stray light passes between the pixels. 

While numerical discretization may not be able to perfectly sample $M(\bm{x})$ to account for arbitrary shifts of $A(\cdot)$ in \cref{eq:mask_gen_app},
these shifts can be accounted for in the spatial frequency domain:
\begin{align}
	\mathcal{F} (M(\bm{x})) = M(\bm{u}) = A(\bm{u}) \sum_{k}^K w_{k}  e^{j (\bm{u} \cdot \bm{x}_k)}.
\end{align}
\cref{eq:after_mask} (for the wave field at the exit of the mask) can therefore be written as
\begin{align}
	U_1^+(\bm{x}; z=d_1, \lambda) = U_1^-(\bm{x}; z=d_1,\lambda)  \mathcal{F}^{-1} (M(\bm{u})).
\end{align}
While this allows for arbitrary shifts, it requires an additional FFT and can be expensive when tracking gradients in order to optimize the mask weights $w_{k}$. 

A cheaper way to account for deadspace is to discretize $M(\bm{x})$ at a finer resolution than of the mask, and only modulate those pixels which fall within the individual mask-pixel apertures (by setting the appropriate $ w_k $ value). While optimizing the programmable mask weights in our end-to-end approach, we adopt this latter simplification which is much more tractable when tracking gradients.

Other parameters that have been modeled for SLMs in holography and that are applicable to the context of imaging with programmable masks include: non-linear mapping between voltage-to-phase/amplitude of the individual mask pixels, Zernike coefficients to model deviations from theoretical propagation models, and content-dependent undiffracted (stray) light~\cite{Peng:2020:NeuralHolography}.

\begin{figure*}[t!]
	\centering
	\begin{subfigure}{.38\textwidth}
		\centering
		\includegraphics[width=0.99\linewidth]{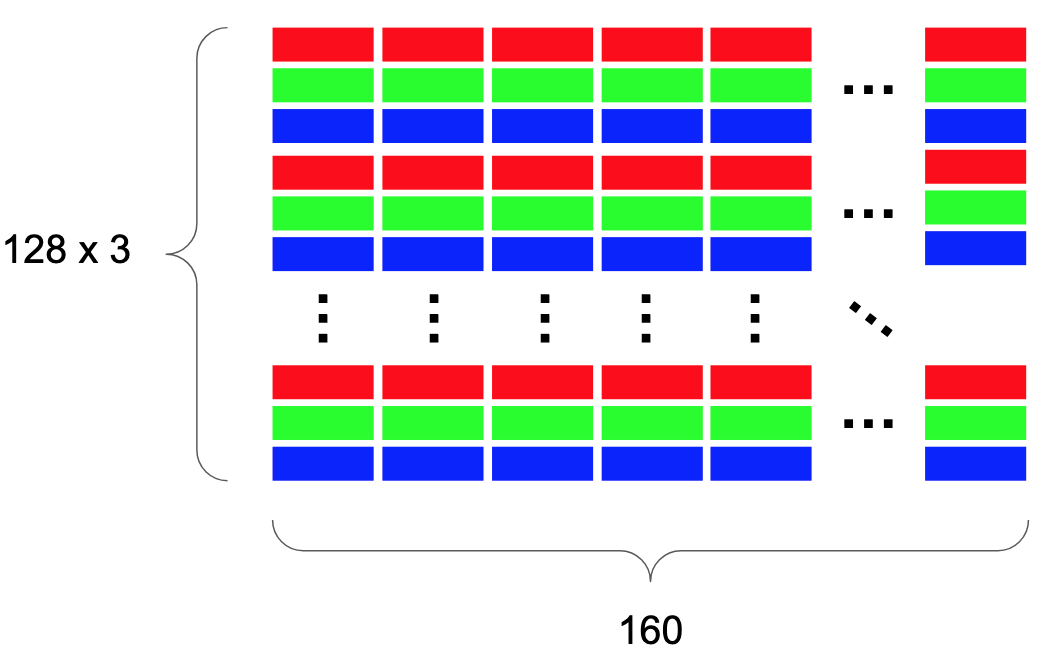}
		\caption{}
		\label{fig:slm_pixel_layout}
	\end{subfigure}
	\begin{subfigure}{.32\textwidth}
		\centering
		\includegraphics[width=0.99\linewidth]{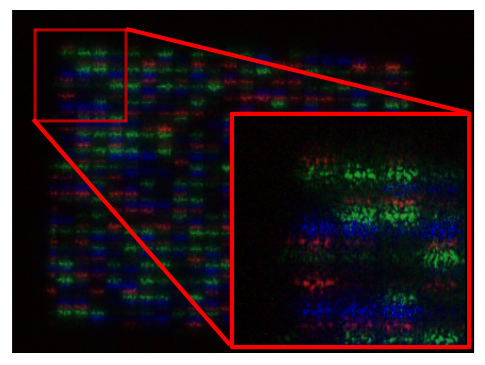}
		\caption{}
		\label{fig:adafruit_section}
	\end{subfigure}	
	\caption{Visualizing pixel layout of the ST7735R component. (a) Red, green, blue color filter arrangement. (b) Zooming into section of a measured point spread function for a random pattern.}
\end{figure*}

\subsection{Specifics for the ST7735R LCD component and the Raspberry Pi High Quality camera}
\label{sec:ST7735R}

As the ST7735R component is originally intended to serve as a color display, it has an interleaved pattern of red, green, and blue filters as shown in \cref{fig:slm_pixel_layout},\footnote{More information can be found on the device driver datasheet: \url{https://cdn-shop.adafruit.com/datasheets/ST7735R_V0.2.pdf}} which can be modeled as a wavelength-dependent version of \cref{eq:mask_gen_app}
\begin{align}
	\label{eq:adafruit_mask}
	M(\bm{x}; \lambda) &= \sum_{c\in\{R,G,B\}} \!\!\!\!\!\! F_c(\lambda)  \, \sum_{k_c}^{K_c}
	w_{k_c}
	A(\bm{x} -\bm{ x}_{k_c}),
\end{align}
where:
\begin{itemize}
	\item $ \{ F_c(\cdot)\}_{c\in\{R,G,B\}} $ is the wavelength-response of each color filter,
	\item $ \Big\{  \{w_{k_c}\}_{k_c=1}^{K_c} , c\in\{R,G,B\}  \Big\} $ are real-valued weights for the red, green, and blue sub-pixels, and $ \Big\{  \{\bm{x}_{k_c}\}_{{k_c}=1}^{K_c}, c\in\{R,G,B\} \Big\}  $ are their respective centers.
\end{itemize}

The display has a resolution of $(128\times 160)$ color pixels, with three sub-pixels per color pixel as shown in \cref{fig:slm_pixel_layout}. The dimension of each sub-pixel is $ (\SI{0.06}{\milli\meter} \times \SI{0.18}{\milli\meter}) $ and the dimension of the entire screen is $ (\SI{28.03}{\milli\meter} \times \SI{35.04}{\milli\meter}) $.\footnote{ST7735R breakout board datasheet: \url{https://cdn-shop.adafruit.com/datasheets/JD-T1800.pdf}} If we assume a uniform spacing of sub-pixels, this corresponds to a pixel pitch of roughly $ (\SI{0.073}{\milli\meter} \times \SI{0.22}{\milli\meter}) $ 
and a fill-factor of $82\%$, namely deadspace of $18\%$ around each sub-pixel. 

The Raspberry Pi High Quality Camera\footnote{Raspberry Pi High Quality Camera datasheet: \url{https://cdn-shop.adafruit.com/product-files/4561/4561+Raspberry+Pi+HQ+Camera+Product+Brief.pdf}} uses the Sony IMX477R back-illuminated sensor which has the following specifications: $(3040 \times 4056)$ pixel resolution, $7.9\si{\milli\meter}$ sensor diagonal, and a pixel size of $ (1.55\si{\micro\meter} \times 1.55\si{\micro\meter} ) $, which corresponds to full sensor dimensions of $ (\SI{4.71}{\milli\meter} \times \SI{6.29}{\milli\meter}) $. As the display of the ST7735R component is larger than the sensor, we use a subset of its pixels that covers the sensor area. From the pixel pitch of the ST7735R component determined above $ (\SI{0.073}{\milli\meter} \times \SI{0.22}{\milli\meter}) $, it can be concluded that the number of sub-pixels that overlap the sensor is around $ (64 \times 22) $. Moreover, for enforcing the LSI assumption described in \cref{sec:conv}, we crop the LCD such that $ 80\% $ of sensor surface is exposed, which corresponds to $ 51 \times 22  = 1122 $ LCD sub-pixels. This is the number of sub-pixels that we optimize in our experiments.

\section{Simulating an image of the desired scene}
\label{sec:simulation}

\begin{table}[t!]
	\centering
	\caption{Dataset distribution and simulation parameters.}
	\label{tab:dataset}
	\begin{tabular}{lccc}
		\toprule
		\textit{Dataset} $\rightarrow $ & MNIST & CelebA & CIFAR10 \\ 
		\midrule
		Scene-to-encoder & \SI{40}{\centi\meter}   & \SI{55}{\centi\meter}  & \SI{40}{\centi\meter}  \\ 
		Object height & \SI{12}{\centi\meter}  & \SI{27}{\centi\meter} & \SI{25}{\centi\meter} \\ 
		Signal-to-noise ratio & \SI{40}{\decibel}  & \SI{40}{\decibel} & \SI{40}{\decibel}  \\
		\bottomrule 
	\end{tabular} 
\end{table}

When simulating the propagation between two planes, as shown in \cref{fig:propagation}, one may wish that the scene in the $ I_0 $ plane corresponds to the content of a digital image. This section describes how to pre-process an image such that its output corresponds to (1) a  scene at a distance $ d_1 $ from the camera, (2) content from the original image having a height $h_{\text{obj}}$ at the plane $ d_1 $ from the camera, and (3) a measurement taken by a camera of a known PSF.
While \cref{alg:sim} describes this simulation procedure succinctly, we dive into more detail in this section. 

An RGB image can be interpreted as image intensities at three wavelengths: red, green, and blue~\cite{sitzmann2018}. We use the following wavelengths for red, green, and blue respectively: \SI{640}{\nano\meter}, \SI{550}{\nano\meter}, and \SI{460}{\nano\meter}. For a grayscale image, such as images from MNIST~\cite{lecun1998mnist}, the same data can be used across channels, or it can be convolved with a grayscale version of the PSF.


Given an image $\bm{x} \in \mathbb{R}^{C\times H\times W}$ with $ C $ channels and an intensity PSF $\bm{p} \in \mathbb{R}^{C \times H_{\text{PSF}}\times W_{\text{PSF}}}$, the simulation of a sensor measurement $\bm{v} \in \mathbb{R}^{C \times DH_{\text{PSF}}\times DW_{\text{PSF}}}$ (with an optional downsampling factor $ D \geq 1 $) can be summarized by the following steps:
\begin{enumerate}
	\item Resize $\bm{x} $ to the PSF's dimension to obtain $\bm{x}_\text{scene} \in \mathbb{R}^{C\times H_{\text{PSF}}\times W_{\text{PSF}}}$, while preserving $\bm{x} $'s original aspect ratio and scaling it to correspond to a desired object height (or width). The details of this rescaling are explained in \cref{sec:rescale}.
	\item Convolve each channel of $\bm{x}_\text{scene} $ with the corresponding PSF channel to obtain $\bm{v} \in \mathbb{R}^{C \times H_{ \text{PSF}}\times W_{\text{PSF}}}$:
	\begin{equation}
	\bm{v}[c] = \bm{x}_\text{scene}[c] \ast \bm{p}[c], \quad c=1, 2, \ldots , C.
	\end{equation}
	Due to large convolution kernels, this is typically best done in the spatial frequency domain, where convolution corresponds to an element-wise multiplication, and the FFT algorithm can be used to efficiently project between spatial and spatial frequency domains.
	\item If $D > 1$, downsample the convolution output to the sensor resolution.  We apply bilinear interpolation for this resizing.
	\begin{equation}
		\bm{v}[c] = \text{Downsample}(\bm{v}[c], D), \quad c=1, 2, \ldots , C.
	\end{equation}
	\item Add noise at a desired signal-to-noise ratio (SNR). More on this in \cref{sec:noise}. 
	\begin{equation}
		\bm{v}[c] = \bm{v}[c] + \bm{n}[c], \quad c=1, 2, \ldots , C.
	\end{equation}
\end{enumerate}

Having a faithful estimate of the intensity PSF $\bm{p} $ is the most vital part of the above simulation. For a fixed optical encoder, if a physical setup is available, the best approach may be to simply measure the PSF by placing a point source (\ie a white LED behind a pinhole as shown in \cref{fig:measuring_psf}) at the desired distance and taking the resulting measurement as the intensity PSF. Some post-processing may be necessary to remove sensor noise and balance color channels. For encoders that have no parametric function, such as pseudo-random diffusers~\cite{Antipa:18}, measuring the PSF may be the only viable option.

For encoders that have a parametric function, \eg lenses or programmable masks, it is possible to simulate the intensity PSF. This is in fact necessary for most end-to-end optimization techniques, unless forward-/back-propagation are done directly with hardware~\cite{Zhou2020}. Even for a parametric encoder, it can be useful to measure the PSF (if a physical setup is available) in order to reduce mismatch due to model assumptions / simplifications~\cite{phlatcam}. In \cref{sec:psf_modeling}, we describe our modeling of the PSF of a programmable mask at a particular wavelength, and explain how we account for the specifics of the ST7735R component and the Raspberry Pi High Quality Camera in our proposed system. 

\paragraph{Using physical measurements.} The above simulation of \cref{fig:propagation} seeks to replicate the physical measurement setup shown in \cref{fig:measurement_setup}, namely projecting the image of the desired scene on a display at a distance $ d_1 $ from the camera. While such a measurement would produce more realistic results, it can be very time-consuming for an \textit{entire dataset} of images. If this dataset is to be used for a task with a fixed optical encoder,  a lens or diffuser, it  may be worth the time and effort as the measurement only has to be done once. For a task that seeks to optimize the optical encoder in an end-to-end fashion, new measurements would have to be performed during training whenever updates are made to the optical encoder. This is highly impracticable for optical encoders that require precise fabrication~\cite{sitzmann2018,phlatcam,markley2021physicsbased}. In the case of programmable optical encoders, alternating between physical measurements and updating the optical encoder lends itself to hardware-in-the-loop / physics-aware training~\cite{Peng:2020:NeuralHolography,wright2022deep}. This has the potential to reduce model-mismatch but at the cost of longer training, due to acquisition time and a lack of parallelization. 

\subsection{Rescaling image to PSF resolution for a desired object height}
\label{sec:rescale}

Note that in this section we use capital letters to denote dimensions in pixels or scalar factors, and lower case letters  for dimensions in meters.
The goal of this step in the simulation of \cref{sec:simulation} is to rescale a digital image $\bm{x} \in \mathbb{R}^{C\times H\times W}$ such that its convolution with a digital PSF $\bm{p} \in \mathbb{R}^{C \times H_{\text{PSF}}\times W_{\text{PSF}}}$ corresponds to the setup in \cref{fig:propagation} for an object of height $h_{\text{obj}}$ at the scene plane. For such a configuration,
namely a scene-to-encoder distance of $d_1$ and an encoder-to-image distance of $d_2$,
the object height \emph{at the sensor} is given by:
\begin{equation}
h_{\text{sensor}} = h_{\text{obj}} (d_2 / d_1) = h_{\text{obj}} \cdot |M|.
\end{equation}
For our simulation we are interested in the number of pixels that this height corresponds to. If our PSF was measured for the above distances with a sensor resolution of $ (H_{\text{sensor}} \times W_{\text{sensor}} ) $ pixels and a pixel pitch of $ p $,
the sensor will have captured a PSF for a scene of the following physical dimensions: 
\begin{equation}
(h_{\text{scene}} \times w_{\text{scene}} ) = \Big(\dfrac{p H_{\text{sensor}} }{|M|} \times \dfrac{p W_{\text{sensor}}  }{|M|}  \Big).
\end{equation}
Consequently, the object height \textit{in pixels} at the sensor is approximately given by:
\begin{equation}
H_{\text{pixel}} = \text{round}\Big( \dfrac{h_{\text{obj}} H_{\text{PSF}}}{h_{\text{scene}}} \Big).
\end{equation}
Therefore, to rescale the original input image $\bm{x} \in \mathbb{R}^{C\times H\times W}$ to the PSF resolution, while preserving its aspect ratio and scaling it such that it corresponds to the desired object height, we need to perform the following steps:
\begin{enumerate}
	\item Resize  $\bm{x} $ to $ \big( C \times \text{round}(SH) \times \text{round}(SW) \big) $ where $S = (H_{\text{pixel}} / H)$ .
	\item Pad above to $(C \times H_{\text{PSF}}\times W_{\text{PSF}})$.
\end{enumerate}
The resulting image $\bm{x}_\text{scene} \in \mathbb{R}^{C\times H_{\text{PSF}}\times W_{\text{PSF}}}$ can then be convolved with the PSF to simulate a propagation as in \cref{fig:propagation}.

\subsection{Adding noise at a desired signal-to-noise ratio}
\label{sec:noise}

Different types of noise can be added during simulation. In practice, read noise at a sensor follows a Poisson distribution with respect to the input. However, as this distribution is not differentiable with respect to the optical encoder parameters, Gaussian noise is used instead~\cite{sitzmann2018,markley2021physicsbased}.

In order to add generated noise to a signal and obtain a desired signal-to-noise ratio (SNR), the generated noise must be scaled appropriately. SNR (in dB) is defined as:
\begin{equation}
\text{SNR} = 10 \log_{10} (\sigma_S^2 / \sigma_N^2),
\end{equation}
where $\sigma_S^2$ is the clean image variance and $\sigma_N^2$ is the generated noise variance. For a target SNR $T$, the generated noise can be scaled with the following factor:
\begin{equation}
k = \sqrt{ \dfrac{\sigma_S^2 }{\sigma_N^2 10 ^{(T/10)}}}.
\end{equation}

In our simulation, we generate noise following a Poisson distribution. As we do not backpropagate through noise generation to the optical encoder parameters, we opt for this more realistic realization of noise.

\section{Baseline and proposed cameras}
\label{sec:baseline}

\begin{figure}[t!]
	\centering
	\begin{subfigure}[b]{.22\textwidth}
		\centering
		\includegraphics[width=0.99\linewidth]{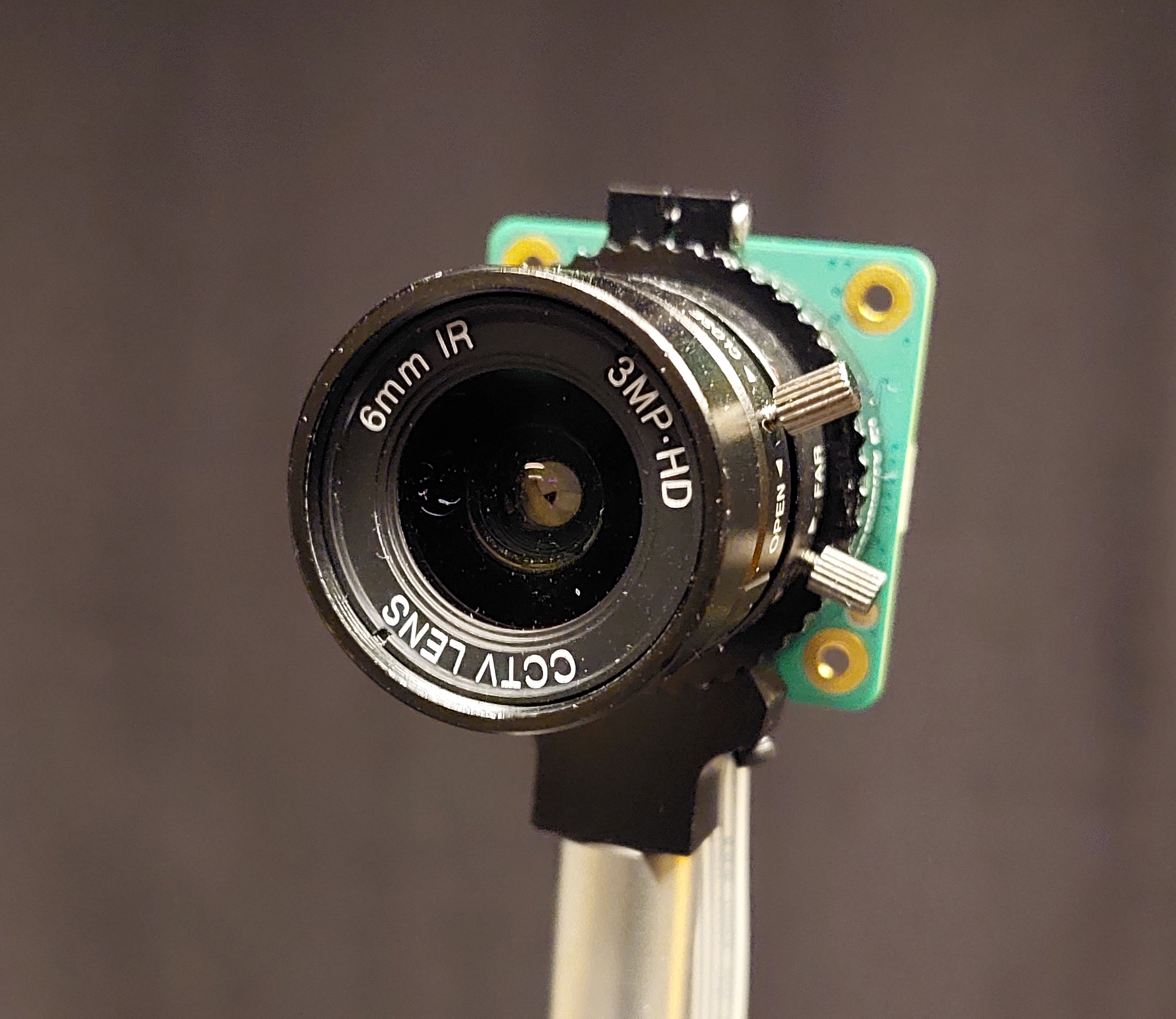}
		\caption{}
		\label{fig:lensed_camera}
	\end{subfigure}
	\begin{subfigure}[b]{.22\textwidth}
		\centering
		\includegraphics[width=0.99\linewidth]{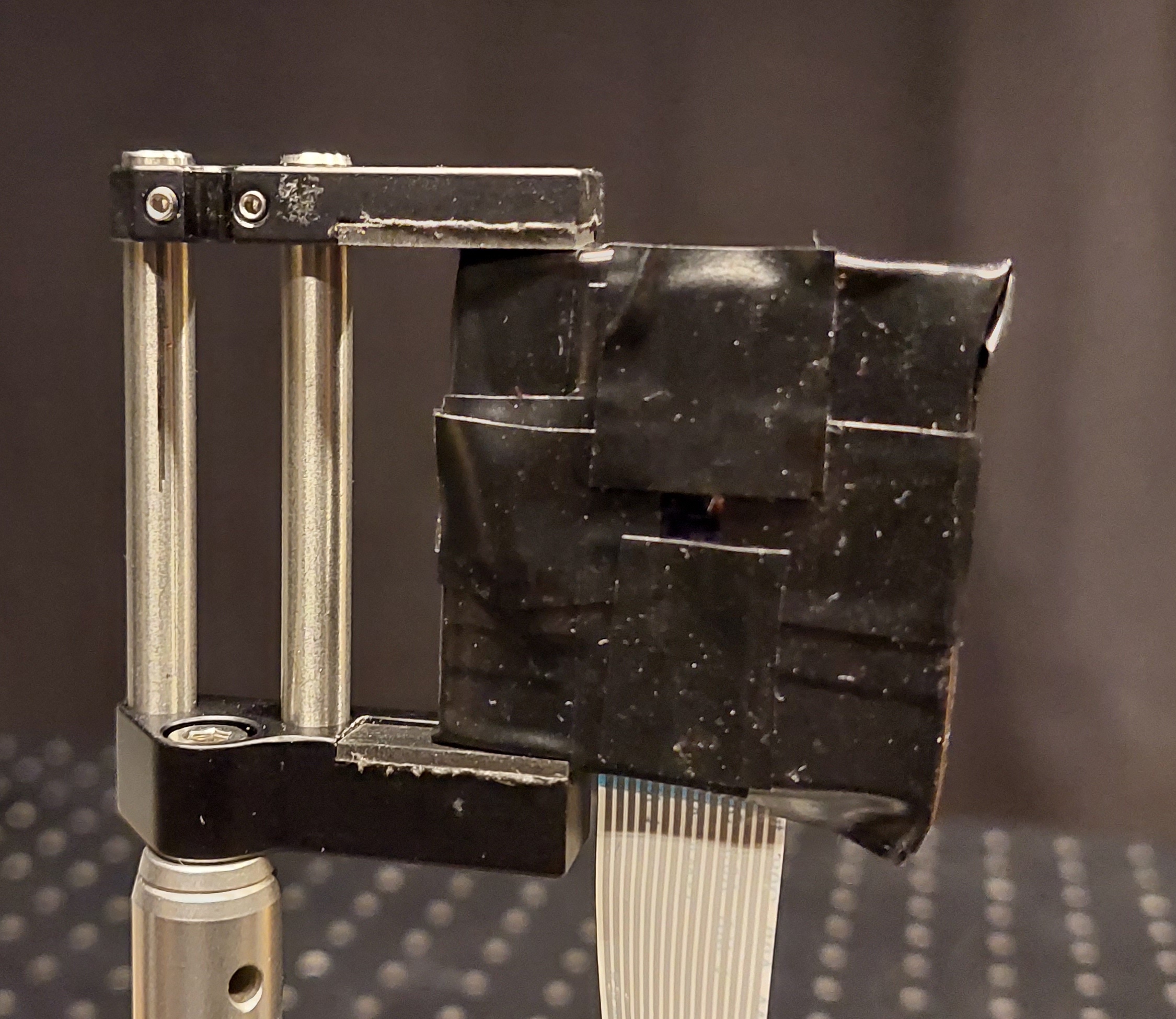}
		\caption{}
		\label{fig:diffuser}
	\end{subfigure}
	\begin{subfigure}[b]{.25\textwidth}
		\centering
		\includegraphics[width=0.99\linewidth]{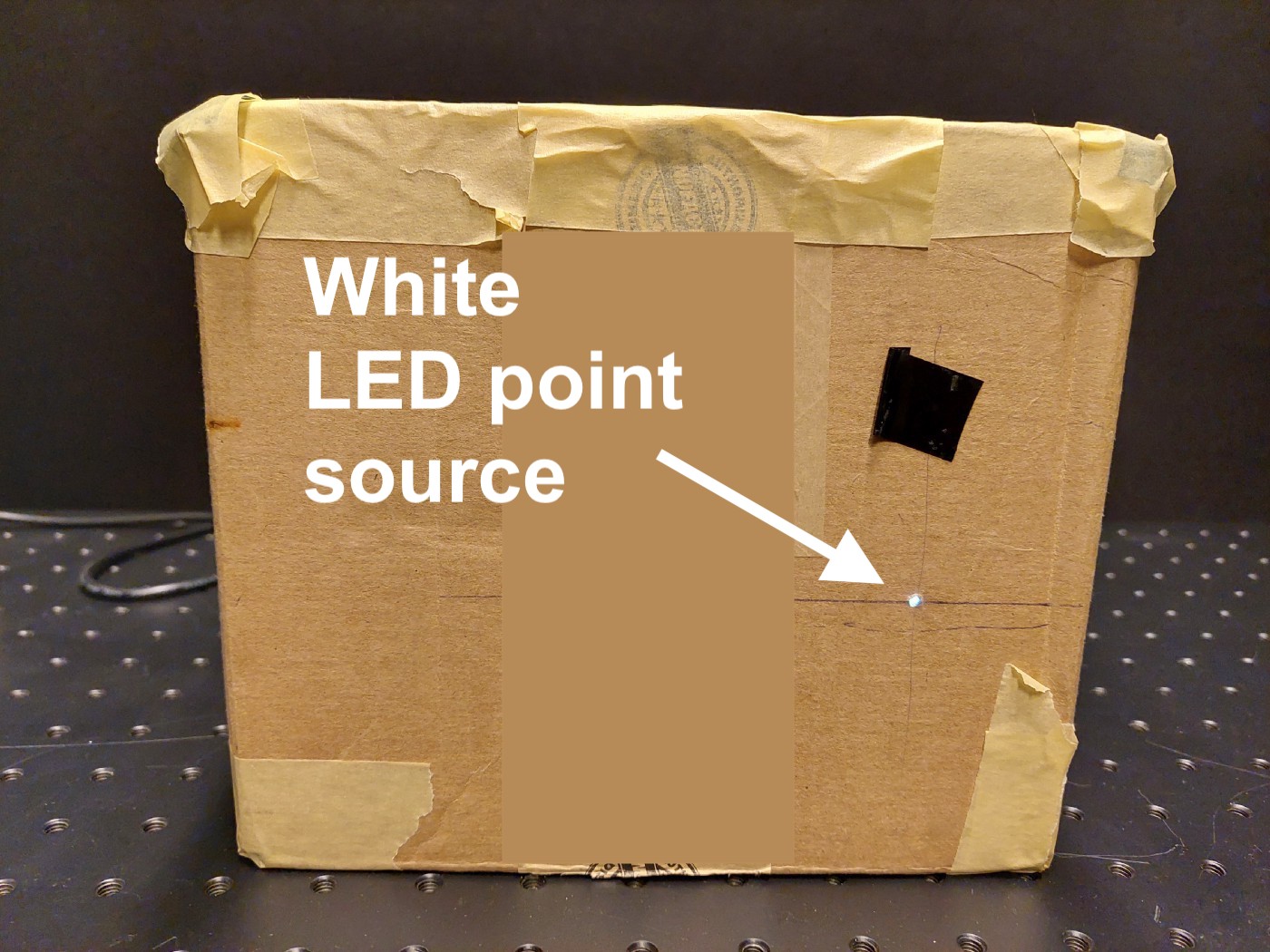}
		\caption{}
		\label{fig:measuring_psf}
	\end{subfigure}
	\caption{Baseline cameras: (a) lensed and (b) diffuser-based. (c) Light source (white LED behind a hole in cardboard box) for measuring the point spread function.}
\end{figure}

As mentioned in \cref{sec:simulation}, a faithful estimate of the intensity PSF is an important part of the simulation for our baseline and proposed cameras: \textit{Lens}, \textit{Coded aperture}~\cite{flatcam}, \textit{Diffuser}~\cite{Antipa:18}, \textit{Fixed mask (m)}, \textit{Fixed mask (s)}, and \textit{Learned mask}. For our experiments in \cref{sec:experiments}, we use the Raspberry Pi High Quality Camera for both the PSF measurement and in simulating the PSF.
Measured PSFs are obtained by placing a white LED behind a pinhole aperture, as shown in \cref{fig:measuring_psf}, at the target distance, and measuring the response in an environment with no external light.
Simulated PSFs are obtained with the approach described in \cref{sec:psf_modeling}. 
Except for \textit{Lens}, the scene, encoder, and image planes ($ U_0, U_1^+, U_2 $ respectively in \cref{fig:propagation}) for simulating the PSFs take on the size and resolution of the Raspberry Pi High Quality Camera downsampled by a factor of $ D = 8 $, namely a pixel resolution of $(380 \times 507)$ and a pixel size of $ (12.4\si{\micro\meter} \times 12.4\si{\micro\meter} ) $. Due to the very compact support of its PSF, we use the full resolution of the Raspberry Pi High Quality Camera, namely $(3040 \times 4056)$.

Below are technical details regarding each PSF:
\begin{itemize}
	\item \textit{Lens}: measured PSF for the camera shown in \cref{fig:lensed_camera}, which has a $6\si{\milli\meter}$ wide angle lens\footnote{6mm Wide Angle Lens for Raspberry Pi HQ Camera datasheet: \url{https://cdn-shop.adafruit.com/product-files/4563/4563-datasheet.pdf}} focused at \SI{40}{\centi\meter}. The lens and its objective have a thickness of \SI{34}{\milli\meter}, and the lens is \SI{7.53}{\milli\meter} from the sensor.
	\item \textit{Coded aperture}: a binary mask is generated by (1) generating a length-$63$ maximum length sequence (MLS) binary array,\footnote{Using the SciPy function \texttt{max\textunderscore len\textunderscore seq}: \url{https://docs.scipy.org/doc/scipy/reference/generated/scipy.signal.max_len_seq.html}} (2) repeating the sequence to create a $ 126 $-length sequence, and (3) computing the outer product with itself to create a $ 126\times126 $ matrix. 
	These are the same steps for generating a coded aperture mask as for FlatCam~\cite{flatcam}, except that we use a shorter MLS sequence ($ 63 $ instead of $ 255 $) to obtain a feature size of $ \SI{30}{\micro\meter} $ (as in~\cite{flatcam}). The mask covers $ 80\% $ of the sensor height (as \textit{Fixed SLM (s)} and \textit{Learned SLM} below). For the PSF, we simulate the mask's diffraction pattern for a distance of $ d_2 = \SI{0.5}{\milli\meter} $, matching the distance in~\cite{flatcam}. 
	\item \textit{Diffuser}: measured PSF for the camera shown in \cref{fig:diffuser}, where the diffuser is placed roughly \SI{4}{\milli\meter} from the sensor. The diffuser is double-sided tape as in the DiffuserCam tutorial~\cite{diffusercam_tut}. In~\cite{lenslesspicam}, the authors demonstrate the effectiveness of this simple diffuser when used with the Raspberry Pi High Quality Camera. It is less than \SI{1}{\milli\meter} thick and is placed roughly \SI{4}{\milli\meter} from the sensor.
	\item \textit{Fixed mask (m)}: measured PSF for the proposed camera shown in \cref{fig:prototype_labeled} for a random pattern (uniform distribution). With a stepper motor, the mask-to-sensor distance is programmatically set to \SI{4}{\milli\meter} to match the distance of the diffuser-based camera.
	\item \textit{Fixed mask (s)}: simulated PSF for the proposed camera, using the approach described in \cref{sec:psf_modeling} for a random set of amplitude values (uniform distribution) and a mask-to-sensor distance of \SI{4}{\milli\meter}. The aperture is set such that mask pixels covering $ 80\% $ of the sensor surface area are exposed. This corresponds to  $ 51 \times 22 = 1122 $ sub-pixels as described in \cref{sec:ST7735R}.
	\item \textit{Learned mask}: simulated PSF for the proposed camera that is obtained by optimizing \cref{eq:optimization} for the mask weights, and simulating the corresponding PSF with  the approach described in \cref{sec:psf_modeling} for a mask-to-sensor distance of \SI{4}{\milli\meter}. Like \textit{Fixed mask (s)}, pixels that cover $ 80\% $ of the sensor surface area are used, corresponding to  $ 51 \times 22 = 1122 $ pixels. As the mask values are updated after backpropagation during training, the resulting PSF is different for each batch. 
\end{itemize}


\section{Training details and classification architectures}
\label{sec:arch}

All experiments in \cref{sec:experiments} were run on a Dell Precision 5820 Tower X-Series (08B1) machine  with an Intel i9-10900X \SI{3.70}{\giga\hertz} CPU and two NVIDIA RTX A5000 GPUs. PyTorch~\cite{paszke2017automatic} was used for dataset preparation and training.

\subsection{Loss functions} For the multi-label classification tasks (MNIST and CIFAR10), we use a cross entropy loss in optimizing \cref{eq:optimization}:
\begin{equation}
	\mathcal{L} (\bm{y}, \bm{\hat{y}}) = -  \sum_{c=1}^{C} \log \dfrac{\exp(\bm{\hat{y}}_{c})}{\sum_{i=1}^{C} \exp(\bm{\hat{y}}_i)} \bm{y}_{c} ,
\end{equation}
where $ \bm{y} \in \mathbb{R}^C$ are one-hot encoded vectors corresponding to the ground-truth labels, and $ \bm{\hat{y}} =  D_{\bm{\theta}_D} \big( O_{\bm{\theta}_E}  ( \bm{x}) \big) \in \mathbb{R}^{C}$ are the predicted scores for a given input $ \bm{x} $ that passes through the optical encoder $ O_{\bm{\theta}_E} (\cdot) $ and the digital decoder $ D_{\bm{\theta}_D}  (\cdot) $. For both MNIST and CIFAR10, $ C = 10 $.

For the binary classification tasks (gender and smiling classification with CelebA), we use a binary cross entropy loss in optimizing \cref{eq:optimization}:
\begin{equation}
	\mathcal{L} (y, \hat{y}) = - \Big[ y \log \hat{y} + (1 - y) \log (1 - \hat{y})  \Big].
\end{equation}

\subsection{Normalization}

For the fixed optical encoders, the embeddings $ \{ \bm{v}_i \}_{i=1}^{N} $ that are inputted to the classifiers are pre-computed with the approach described in \cref{sec:simulation}. The resulting augmented dataset is normalized (according to the augmented training set statistics). For \textit{Learned mask}, we apply batch normalization~\cite{10.5555/3045118.3045167} and a ReLU activation to the sensor embedding prior to passing it to the classifier.

\subsection{Classification architectures}

\begin{figure*}[t!]
	\centering
	\includegraphics[width=0.6\linewidth]{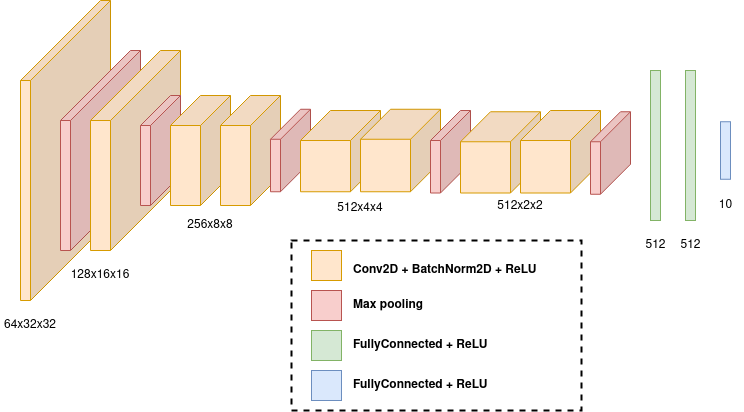}
	\caption{VGG11 architecture for CIFAR10.}
	\label{fig:vgg11}
\end{figure*}

The following classification architectures are used in the experiments of  \cref{sec:experiments}.

\paragraph{Logistic regression (LR).}
\label{sec:logistic}
\noindent The classifier performs the following steps:
\begin{enumerate}
	\tightlist
	\item Flatten sensor embedding. 
	\item Fully connected linear layer to $ 10 $ classes if MNIST or CIFAR10, else a single unit for CelebA.
	\item Softmax decision layer if MNIST or CIFAR10, else sigmoid activation for CelebA. 
\end{enumerate}

\paragraph{Two-layer fully connected neural network (FC).}
\label{sec:nn}

\noindent The classifier performs the following steps:
\begin{enumerate}
	\tightlist
	\item Flatten sensor embedding. 
	\item Fully connected linear layer to hidden layer of $ 800 $ units, as in~\cite{1227801}.
	\item Batch normalization.
	\item ReLU activation.
	\item Fully connected linear layer to $ 10 $ classes if MNIST or CIFAR10,  else single unit for CelebA.
	\item Softmax decision layer if MNIST or CIFAR10, else sigmoid activation for CelebA. 
\end{enumerate}

\paragraph{VGG11.}
\label{sec:vgg11}

\noindent The classifier performs the following steps:
\begin{enumerate}
	\tightlist
	\item Eight convolutional layers using $ (3\times 3) $ filters with max-pooling in between some layers.
	\item Flatten.
	\item Three-layer fully-connected network. The first two layers are preceded by \SI{50}{\percent} dropout (in training) and followed by a ReLU activation.
\end{enumerate}

The architecture can be seen in \cref{fig:vgg11}, and is inspired by that of the original VGG paper~\cite{simonyan2014very}).\footnote{Adapted code from this CIFAR10 architecture: \url{https://github.com/kuangliu/pytorch-cifar/blob/master/models/vgg.py}}

\subsection{Training hyperparameters} All classifiers are trained for $ 50 $ epochs. 
Unless noted otherwise, the Adam optimizer~\cite{adam} is used with an initial learning rate of $ 0.001 $.
Depending on the experiment, different batch sizes and learning rate schedules were used in order to avoid over-fitting.
\cref{sec:mnist_vary_app,sec:mnist_robust_app,sec:celeba_app,sec:cifar10_app} describe the specific training hyperparameters for the end-to-end training experiments in \cref{sec:mnist,sec:celeba,sec:cifar10}.

For \textit{Learned mask} we tend to a larger batch size for speeding up training time, as the simulation of the new PSF (after updating the mask values) and of all the train / test examples results in a significantly larger training time. For the fixed encoders (the rest) the train and test examples can be simulated offline, \ie as a pre-processing step.
For our hardware and $ 50 $ epochs, it took around \SI{20}{\minute} for training fixed encoders and \SI{280}{\minute} for \textit{Learned mask}.

\section{MNIST -- varying sensor dimension and architecture}
\label{sec:mnist_vary_app}

\newcommand{\curvesize}{0.45}
\begin{figure*}[h!]
	\begingroup
	\centering
	\renewcommand{\arraystretch}{1} 
	\setlength{\tabcolsep}{0.2em} 
	\scalebox{1.0}{
		\begin{tabular}{ccc}
			& LR & FC \\
			
			24$\times$32 &
			\includegraphics[width=\curvesize\linewidth,valign=m]{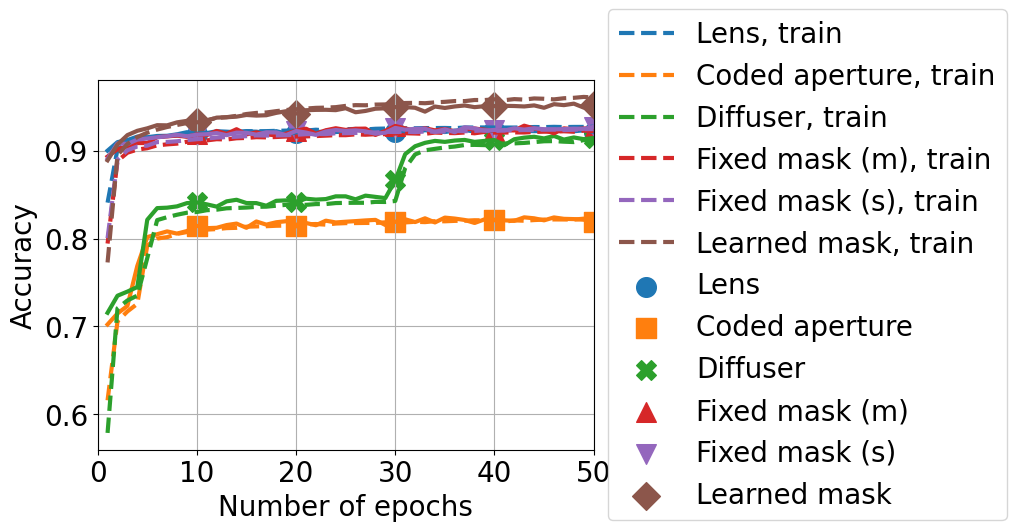}  &
			\includegraphics[width=\curvesize\linewidth,valign=m]{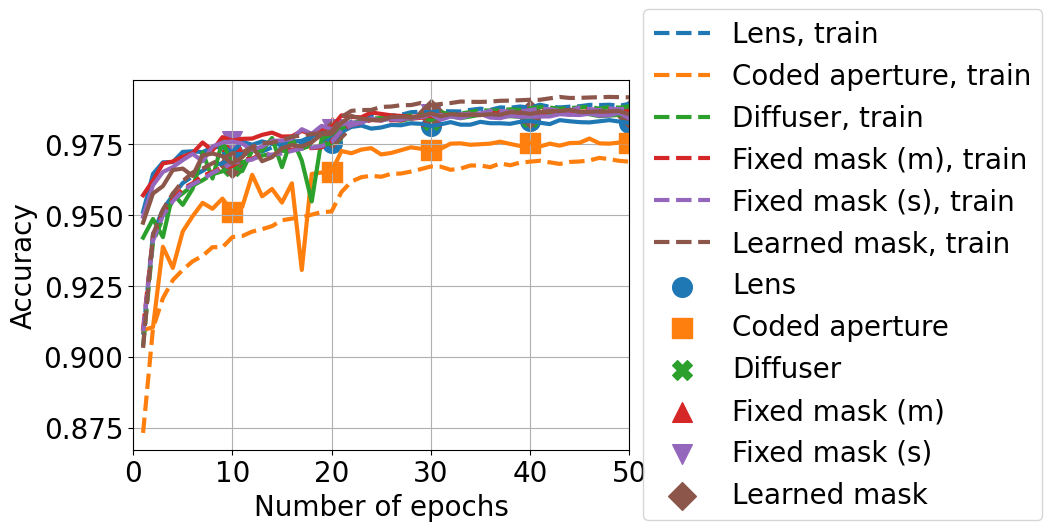}\\[25pt]
			
			12$\times$16&
			\includegraphics[width=\curvesize\linewidth,valign=m]{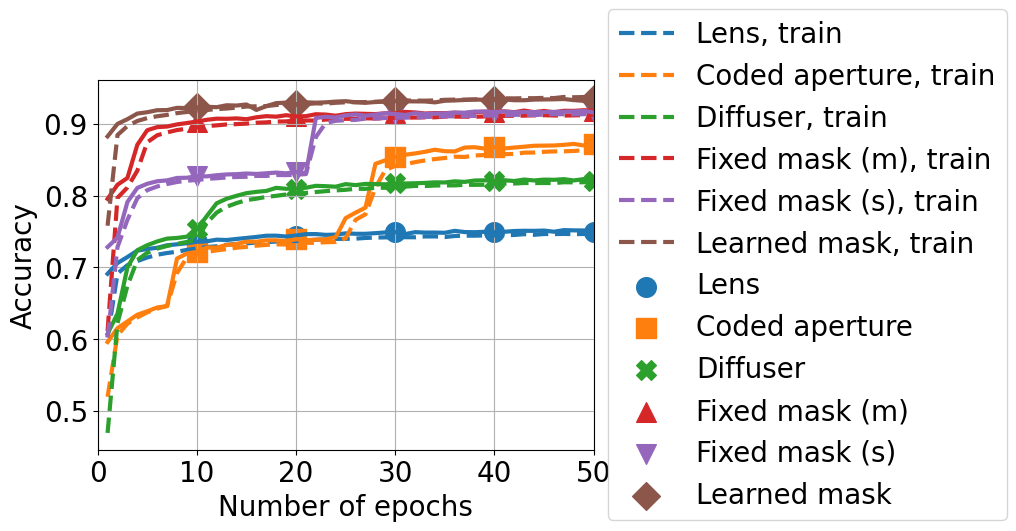}  &
			\includegraphics[width=\curvesize\linewidth,valign=m]{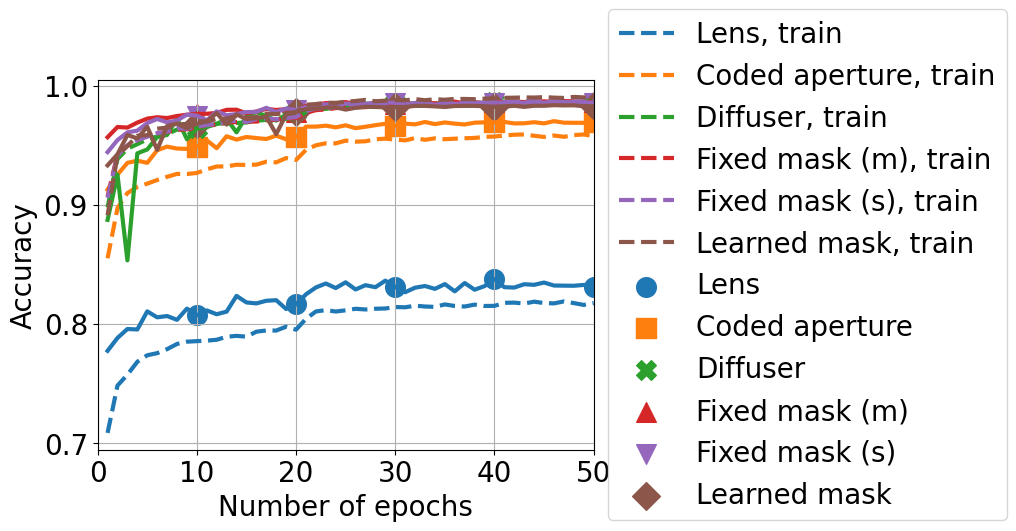}\\[25pt]
			
			6$\times$8 &
			\includegraphics[width=\curvesize\linewidth,valign=m]{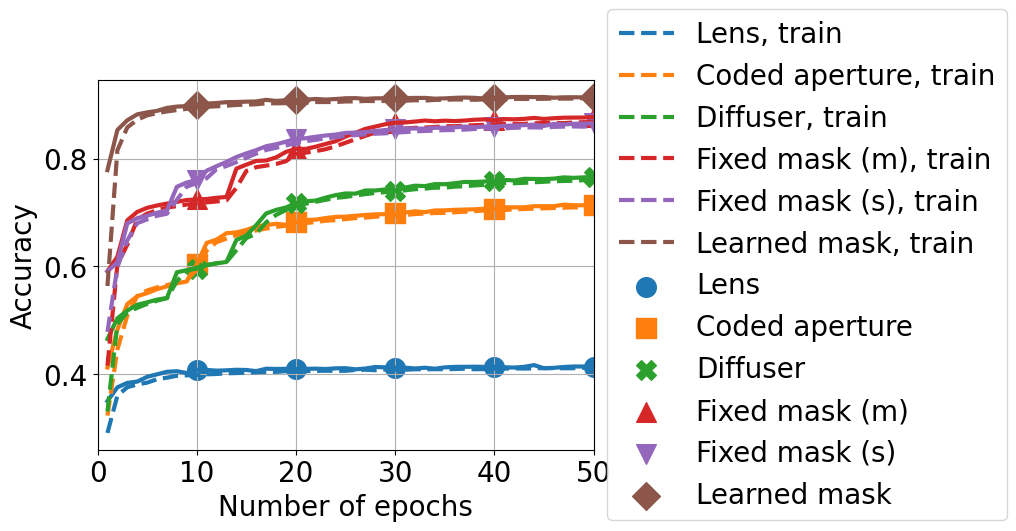}  &
			\includegraphics[width=\curvesize\linewidth,valign=m]{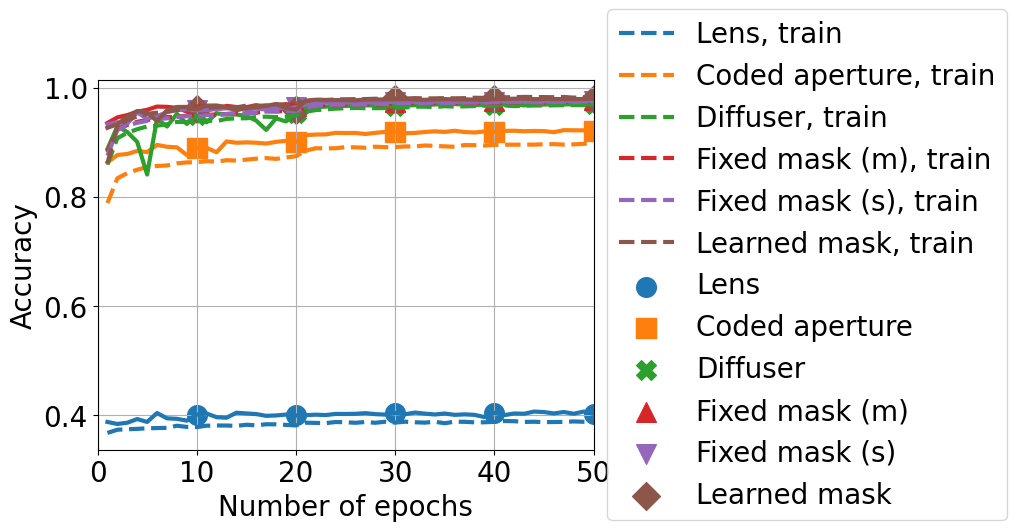}\\[25pt]
			
			3$\times$4&
			\includegraphics[width=\curvesize\linewidth,valign=m]{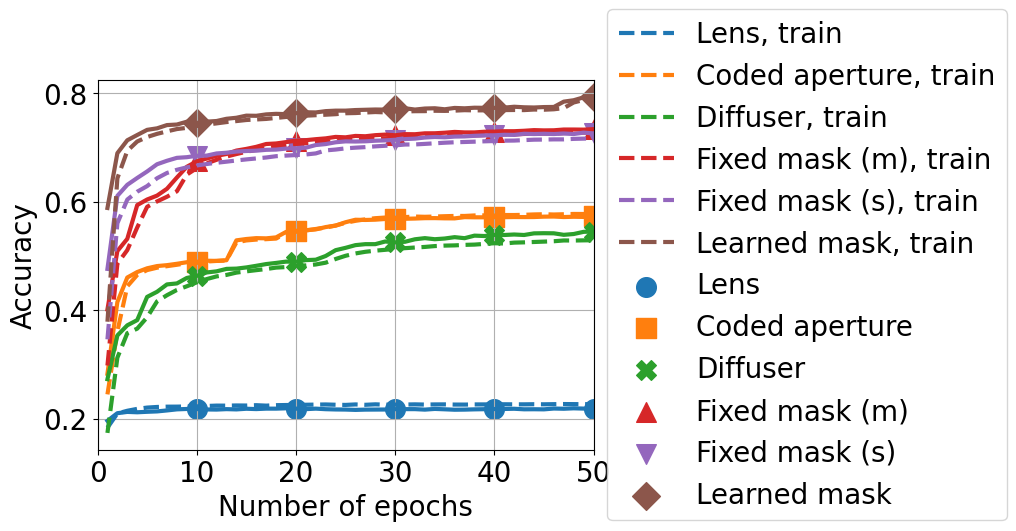}  &
			\includegraphics[width=\curvesize\linewidth,valign=m]{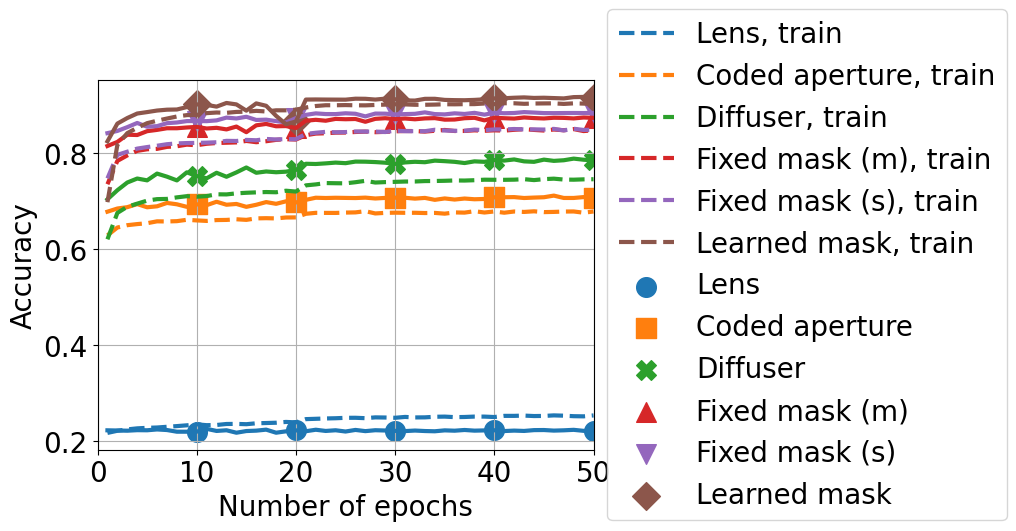}
			
		\end{tabular}
	}
	\endgroup
	\caption{Train and test curves for MNIST.}
	\label{tab:mnist_vary_curves}
\end{figure*}

\cref{tab:vary_dim_hparam} details the training hyperparameters for the experiment in \cref{sec:vary_dimension}  that varies the embedding dimension and architecture for handwritten digit classification (MNIST). 
Train and test accuracy curves can be see in \cref{tab:mnist_vary_curves}. 
Example sensor embeddings can be found in \cref{tab:mnist_examples}.
\begin{table}[h]
	\centering
	\caption{Training hyperparameters for experiment in \cref{sec:mnist} for handwritten digit classification (MNIST). LR and FC architectures are described in \cref{sec:logistic} and \cref{sec:nn} respectively. \textit{Schedule} denotes (if applied) after how many epochs the learning rate is reduced by a factor of $ 0.1 $. All models are trained for 50 epochs.}
	\label{tab:vary_dim_hparam}
	\begin{tabular}{lcc}
		\toprule
		\textit{Configuration} $\downarrow $ & \textit{Batch size} & \textit{Schedule}  \\ 
		\midrule
		LR, fixed encoders & 64  & NA  \\ 
		LR, \textit{Learned mask} & 100 & NA\\ 
		FC, fixed encoders & 32 & 20 \\
		FC, \textit{Learned mask} & 64 & 20 \\
		\bottomrule 
	\end{tabular} 
\end{table}

\newpage
~
\newpage
~
\newpage

\newcommand{\figsizeex}{0.17}
\newcommand{\newlineex}{10pt}
\begin{figure}[h!]
	\begingroup
	\centering
	\renewcommand{\arraystretch}{1} 
	\setlength{\tabcolsep}{0.2em} 
	\scalebox{1.0}{
		\begin{tabular}{ccccc}
			& 24$\times$32 & 12$\times$16 & 6$ \times $8 & 3$\times$4  \\
			
			Lens &
			\includegraphics[width=\figsizeex\linewidth,valign=m]{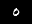}  &
			\includegraphics[width=\figsizeex\linewidth,valign=m]{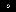} &
			
			\includegraphics[width=\figsizeex\linewidth,valign=m]{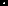}  &
			\includegraphics[width=\figsizeex\linewidth,valign=m]{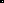}\\[\newlineex]
			
			\makecell{Coded \\aperture\\\cite{flatcam}}
			&
			\includegraphics[width=\figsizeex\linewidth,valign=m]{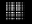}  & \includegraphics[width=\figsizeex\linewidth,valign=m]{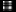} &
			\includegraphics[width=\figsizeex\linewidth,valign=m]{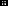} & \includegraphics[width=\figsizeex\linewidth,valign=m]{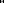}\\[\newlineex]
			
			\makecell{Diffuser~\cite{Antipa:18}}
			&
			\includegraphics[width=\figsizeex\linewidth,valign=m]{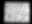} & \includegraphics[width=\figsizeex\linewidth,valign=m]{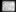} &
			\includegraphics[width=\figsizeex\linewidth,valign=m]{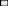} & \includegraphics[width=\figsizeex\linewidth,valign=m]{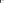}\\[\newlineex]
			
			\makecell{Fixed\\mask (m)}
			&
			\includegraphics[width=\figsizeex\linewidth,valign=m]{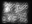} & \includegraphics[width=\figsizeex\linewidth,valign=m]{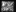} &
			\includegraphics[width=\figsizeex\linewidth,valign=m]{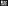} & \includegraphics[width=\figsizeex\linewidth,valign=m]{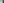}\\[\newlineex]
			
			\makecell{Fixed\\mask (s)}
			&
			\includegraphics[width=\figsizeex\linewidth,valign=m]{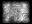}  & \includegraphics[width=\figsizeex\linewidth,valign=m]{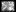} &
			\includegraphics[width=\figsizeex\linewidth,valign=m]{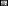} & \includegraphics[width=\figsizeex\linewidth,valign=m]{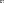}\\[\newlineex]
			
			\makecell{Learned\\mask}
			&
			\includegraphics[width=\figsizeex\linewidth,valign=m]{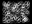}  & \includegraphics[width=\figsizeex\linewidth,valign=m]{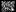} &
			\includegraphics[width=\figsizeex\linewidth,valign=m]{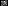} & \includegraphics[width=\figsizeex\linewidth,valign=m]{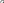}
			
		\end{tabular}
	}
	\endgroup
	\caption{Example sensor embeddings of the baseline and proposed camera systems.}
	\label{tab:mnist_examples}
\end{figure}

\section{MNIST -- robustness to common image transformations}
\label{sec:mnist_robust_app}

\begin{figure*}[h!]
	\centering
	\begin{subfigure}[b]{.45\textwidth}
		\centering
		\includegraphics[width=0.99\linewidth]{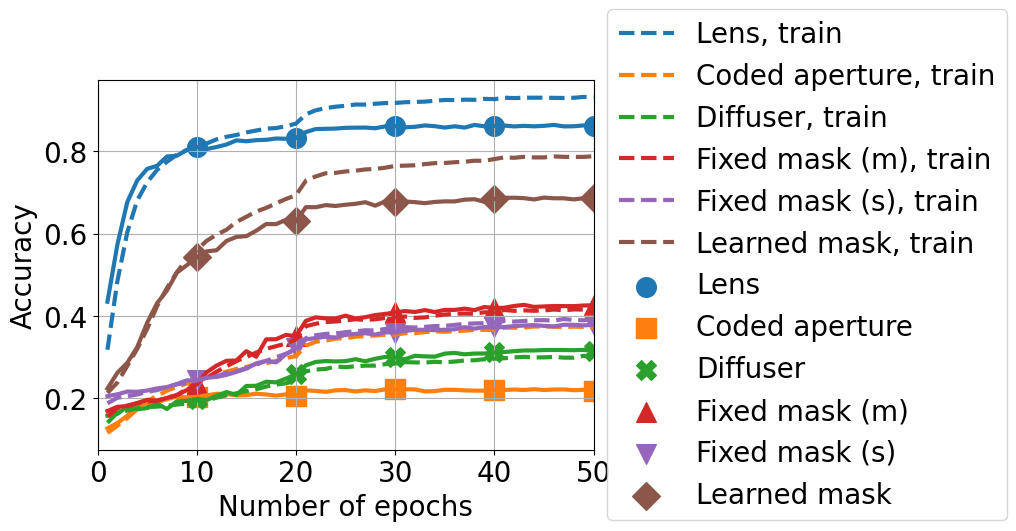}
		\caption{Random shifts.}
		\label{fig:FCNN_24x32_Shift}
	\end{subfigure}
	\begin{subfigure}[b]{.45\textwidth}
		\centering
		\includegraphics[width=0.99\linewidth]{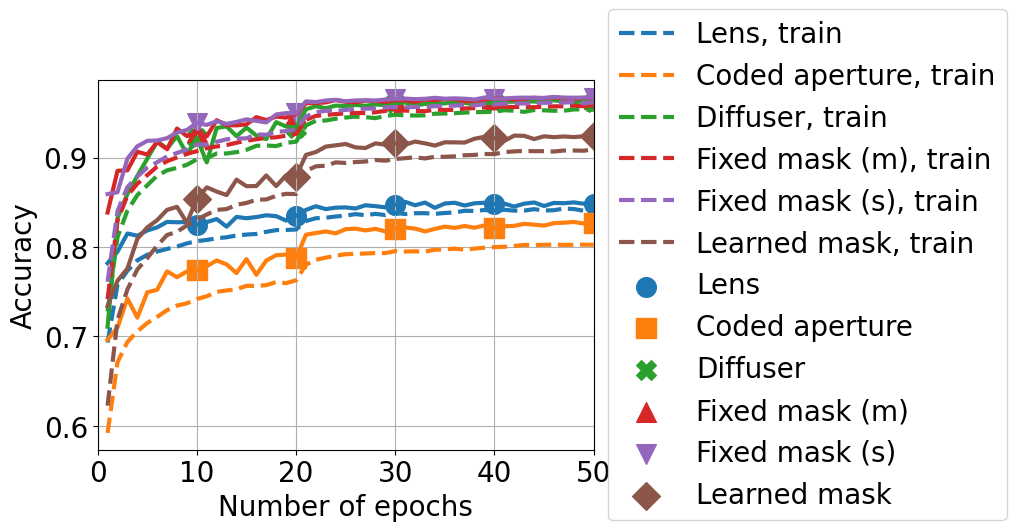}
		\caption{Random rescaling.}
		\label{fig:FCNN_24x32_Rescale}
	\end{subfigure}\\
	\begin{subfigure}[b]{.45\textwidth}
		\centering
		\includegraphics[width=0.99\linewidth]{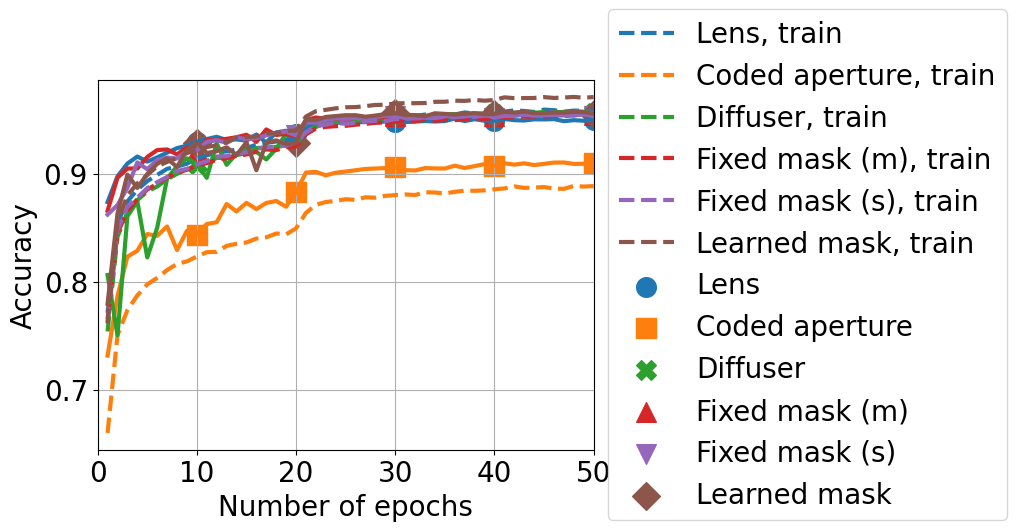}
		\caption{Random rotation.}
		\label{fig:FCNN_24x32_Rotate}
	\end{subfigure}
	\begin{subfigure}[b]{.45\textwidth}
		\centering
		\includegraphics[width=0.99\linewidth]{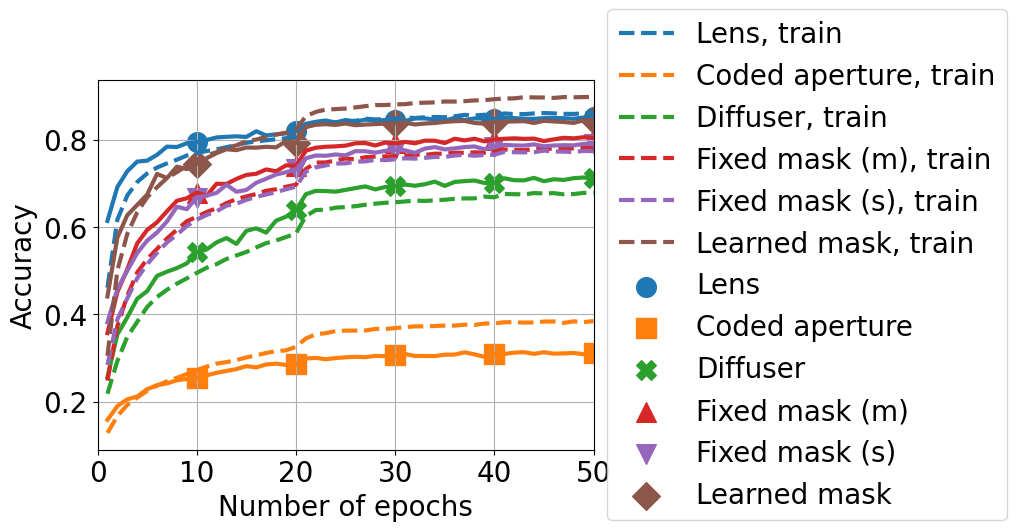}
		\caption{Random perspective changes.}
		\label{fig:FCNN_24x32_Perspective}
	\end{subfigure}
	\caption{Train and test curves for MNIST simulated with perturbations.}
	\label{fig:mnist_perturb_curves}
\end{figure*}

Only FC is used as a digital classifier in this experiment; the same hyperparameters shown in \cref{tab:vary_dim_hparam} are used for the batch size and learning rate schedule. Train and test accuracy curves can be see in \cref{fig:mnist_perturb_curves}. 
The classification accuracy under each random perturbation can be seen in \cref{tab:mnist_robustness_abs}.

\begin{table}[h]
	\caption{MNIST accuracy for two-layer fully connected network with 800 hidden units and an embedding dimension of $ (24\times 32) $ for a randomly transformed data.}
	\label{tab:mnist_robustness_abs}
	\centering
	\scalebox{0.8}{
		\begin{tabular}{l ccccc}
			\toprule
			\textit{Encoder}  $\downarrow $     &  Original & Shift & Rescale   &Rotate   &  Perspective \\
			\midrule
			\textit{Lens}   &$ \mathit{98.4\%} $&
			$ \mathit{86.4\%} $  & $ \mathit{85.0\%} $& $ \mathit{95.1\%} $& $\mathit{85.1\%}$ \\ \hdashline\noalign{\vskip 0.5ex}
			Coded aperture   & $ 97.7\% $&
			$ 22.2\% $  & $ 82.8\% $& $ 91.0\% $& $31.3\%$\\
			Diffuser   & $ 98.6\% $&
			$ 31.8\% $  & $96.4\% $& $ 95.7\% $& $71.4\%$\\
			Fixed mask (m)   & $ \bm{98.7\%} $&
			$ 42.7\% $  & $ 96.7\% $& $ 95.6\% $& $80.6\%$\\
			Fixed mask (s)   & $ 98.6\% $&
			$ 37.9\% $  & $ \bm{96.8\%}  $& $ 95.5\% $& $79.1\%$\\
			Learned mask   & $ \bm{98.7\%} $&
			$ \bm{68.8\%} $  & $92.5\% $& $\bm{95.8\%} $& $\bm{84.5\%}$  \\
			\bottomrule
	\end{tabular}}
\end{table}

In the following subsections, we detail the distribution of perturbations applied to the train and test sets \textit{offline}, \ie as a pre-processing step when preparing the data for both fixed and learned encoders.
Example sensor embeddings and reconstructions are shown to demonstrate how the different imaging systems cope with the random perturbations. For comparison, \cref{tab:mnist_original} are example sensor embeddings and reconstructions when there are no perturbations (\textit{Original}).

\newcommand{\figsizeperturb}{0.18}
\begin{figure}[h!]
	\begingroup
	\centering
	\renewcommand{\arraystretch}{1} 
	\setlength{\tabcolsep}{0.2em} 
	\scalebox{1.0}{
		\begin{tabular}{ccccc}
			
			\includegraphics[width=\figsizeperturb\linewidth,valign=m]{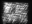}  &
			\includegraphics[width=\figsizeperturb\linewidth,valign=m]{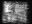}  &
			\includegraphics[width=\figsizeperturb\linewidth,valign=m]{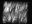} &
			\includegraphics[width=\figsizeperturb\linewidth,valign=m]{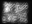}  &
			\includegraphics[width=\figsizeperturb\linewidth,valign=m]{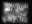}\\[\newlineex]

			\includegraphics[width=\figsizeperturb\linewidth,valign=m]{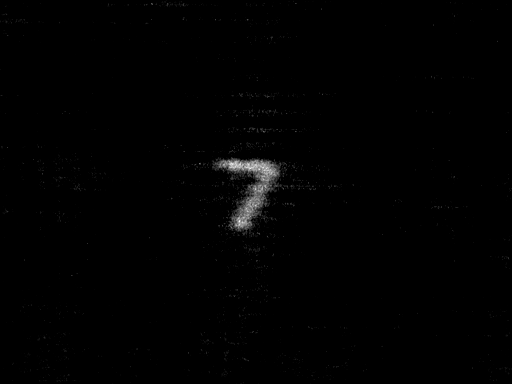} &
			\includegraphics[width=\figsizeperturb\linewidth,valign=m]{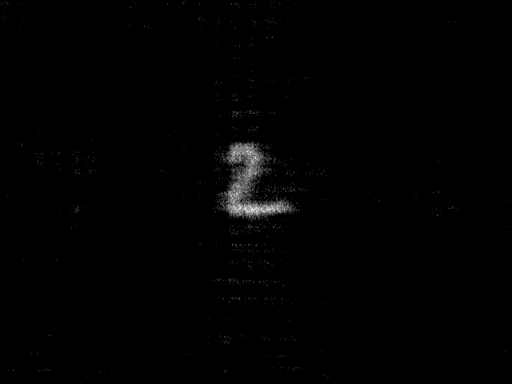}  & \includegraphics[width=\figsizeperturb\linewidth,valign=m]{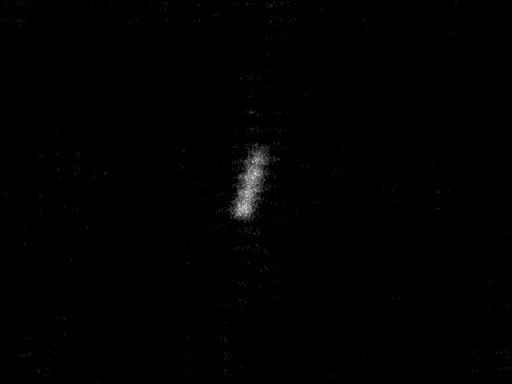} &
			\includegraphics[width=\figsizeperturb\linewidth,valign=m]{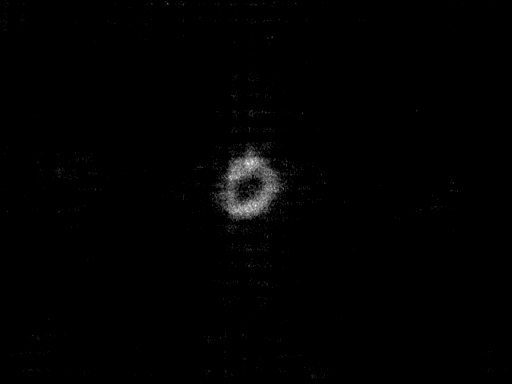} & \includegraphics[width=\figsizeperturb\linewidth,valign=m]{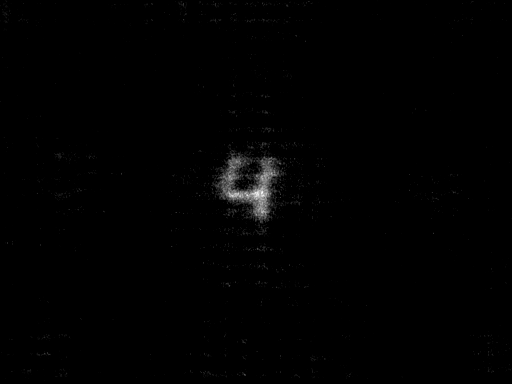}\\
			
		\end{tabular}
	}
	\endgroup
	\caption{Example sensor embeddings for \textit{Fixed mask (m)} with no perturbations (top) and the corresponding reconstruction (bottom) using the convex optimization approach described in \cref{sec:cvx}.}
	\label{tab:mnist_original}
\end{figure}

\newpage
\subsection{Shift}
\label{sec:mnist_shift}

While maintaining an object height of \SI{12}{\centi\meter} and an object-to-camera distance of \SI{40}{\centi\meter}, shift the image in any direction along the object plane such that it is still fully captured by the sensor.

\begin{figure}[h!]
	\begingroup
	\centering
	\renewcommand{\arraystretch}{1} 
	\setlength{\tabcolsep}{0.2em} 
	\scalebox{1.0}{
		\begin{tabular}{ccccc}
			
			\includegraphics[width=\figsizeperturb\linewidth,valign=m]{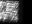}  &
			\includegraphics[width=\figsizeperturb\linewidth,valign=m]{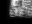}  &
			\includegraphics[width=\figsizeperturb\linewidth,valign=m]{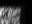} &
			\includegraphics[width=\figsizeperturb\linewidth,valign=m]{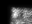}  &
			\includegraphics[width=\figsizeperturb\linewidth,valign=m]{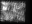}\\[\newlineex]

			\includegraphics[width=\figsizeperturb\linewidth,valign=m]{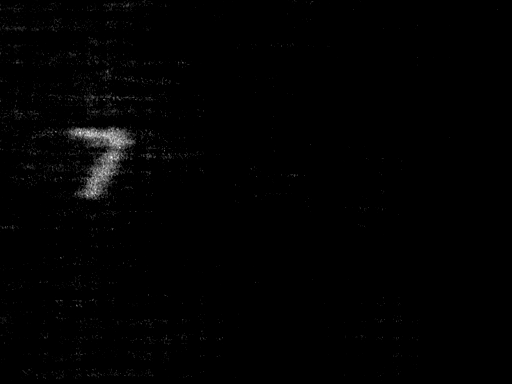} &
			\includegraphics[width=\figsizeperturb\linewidth,valign=m]{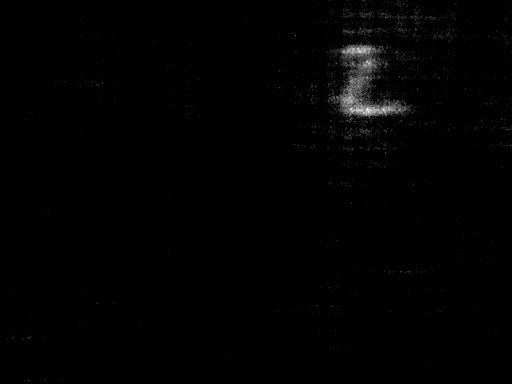}  & \includegraphics[width=\figsizeperturb\linewidth,valign=m]{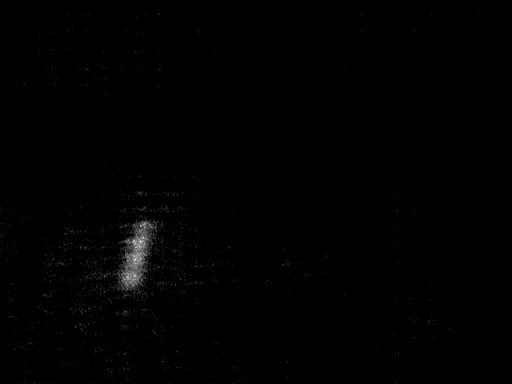} &
			\includegraphics[width=\figsizeperturb\linewidth,valign=m]{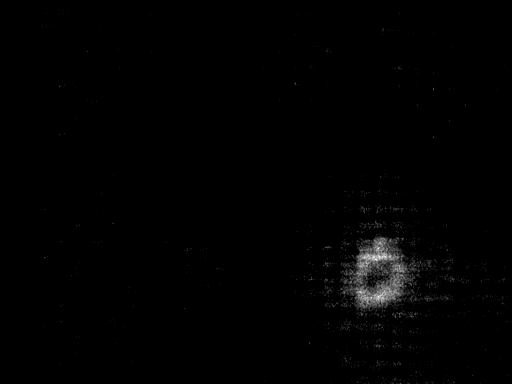} & \includegraphics[width=\figsizeperturb\linewidth,valign=m]{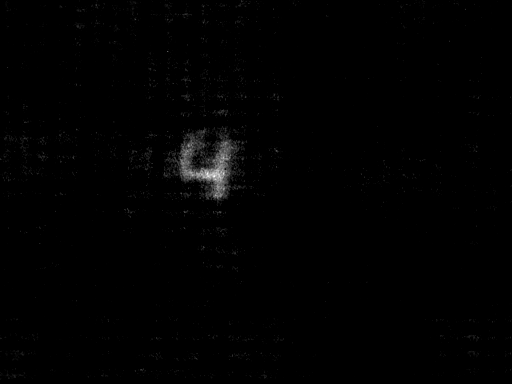}\\
			
		\end{tabular}
	}
	\endgroup
	\caption{Example sensor embeddings for \textit{Fixed mask (m)} in the presence of shifting (top) and the corresponding reconstruction (bottom) using the convex optimization approach described in \cref{sec:cvx}.}
	\label{tab:mnist_shift}
\end{figure}

\subsection{Rescale}

While maintaining an object-to-camera distance of \SI{40}{\centi\meter}, set a random height uniformly drawn from $ [\SI{2}{\centi\meter}, \SI{20}{\centi\meter}] $.

\newcommand{\figsizerescale}{0.23}
\begin{figure}[h!]
	\begingroup
	\centering
	\renewcommand{\arraystretch}{1} 
	\setlength{\tabcolsep}{0.2em} 
	\scalebox{1.0}{
		\begin{tabular}{cccc}
			\stackunder[4pt]{\SI{2}{\centi\meter}}{\includegraphics[width=\figsizerescale\linewidth,valign=m]{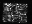}}  &
			\stackunder[4pt]{\SI{8}{\centi\meter}}{\includegraphics[width=\figsizerescale\linewidth,valign=m]{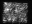}}  &
			\stackunder[4pt]{\SI{14}{\centi\meter}}{\includegraphics[width=\figsizerescale\linewidth,valign=m]{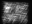}} &
			\stackunder[4pt]{\SI{20}{\centi\meter}}{\includegraphics[width=\figsizerescale\linewidth,valign=m]{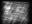}}\\[45pt]
			
			\includegraphics[width=\figsizerescale\linewidth,valign=m]{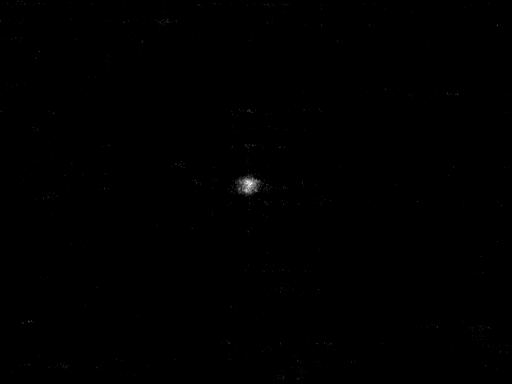}  &
			\includegraphics[width=\figsizerescale\linewidth,valign=m]{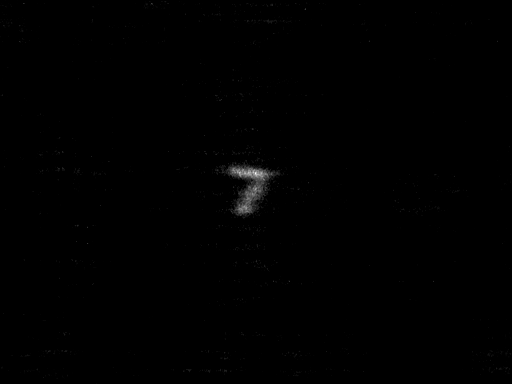}  &
			\includegraphics[width=\figsizerescale\linewidth,valign=m]{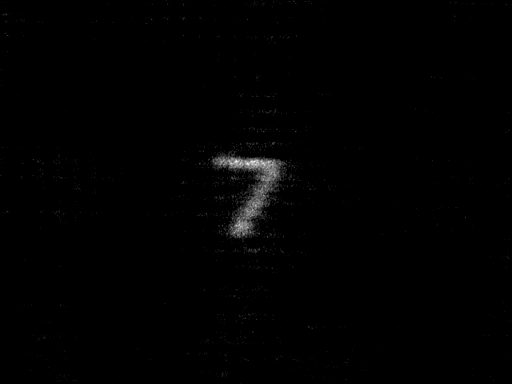} &
			\includegraphics[width=\figsizerescale\linewidth,valign=m]{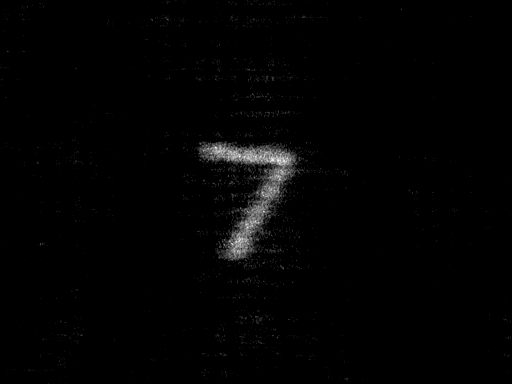}\\
			
		\end{tabular}
	}
	\endgroup
	\caption{Example sensor embeddings for \textit{Fixed mask (m)} in the presence of rescaling (top) and the corresponding reconstruction (bottom) using the convex optimization approach described in \cref{sec:cvx}.}
	\label{tab:mnist_rescale}
\end{figure}

\newpage
\subsection{Rotate}

While maintaining an object height of \SI{12}{\centi\meter} and an object-to-camera distance of \SI{40}{\centi\meter}, uniformly draw a rotation angle from $ [\SI{-90}{\deg}, \SI{90}{\deg}] $.

\begin{figure}[h!]
	\begingroup
	\centering
	\renewcommand{\arraystretch}{1} 
	\setlength{\tabcolsep}{0.2em} 
	\scalebox{1.0}{
		\begin{tabular}{ccccc}
			
			\includegraphics[width=\figsizeperturb\linewidth,valign=m]{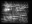}  &
			\includegraphics[width=\figsizeperturb\linewidth,valign=m]{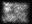}  &
			\includegraphics[width=\figsizeperturb\linewidth,valign=m]{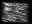} &
			\includegraphics[width=\figsizeperturb\linewidth,valign=m]{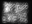}  &
			\includegraphics[width=\figsizeperturb\linewidth,valign=m]{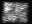}\\[\newlineex]

			\includegraphics[width=\figsizeperturb\linewidth,valign=m]{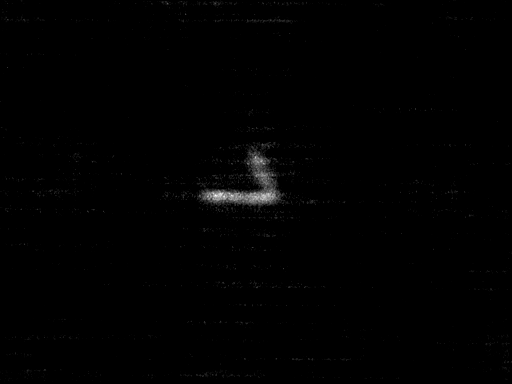} &
			\includegraphics[width=\figsizeperturb\linewidth,valign=m]{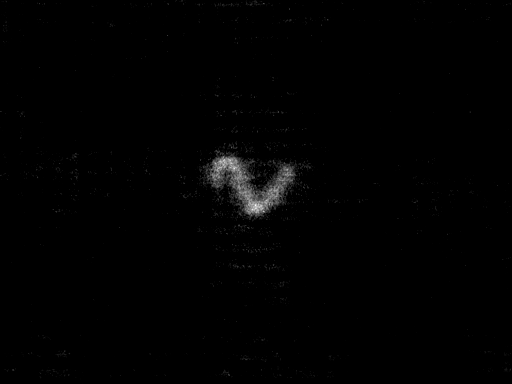}  & \includegraphics[width=\figsizeperturb\linewidth,valign=m]{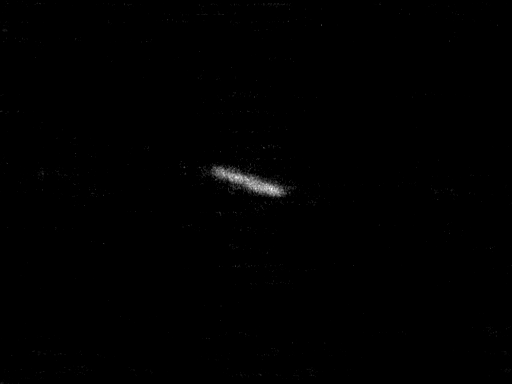} &
			\includegraphics[width=\figsizeperturb\linewidth,valign=m]{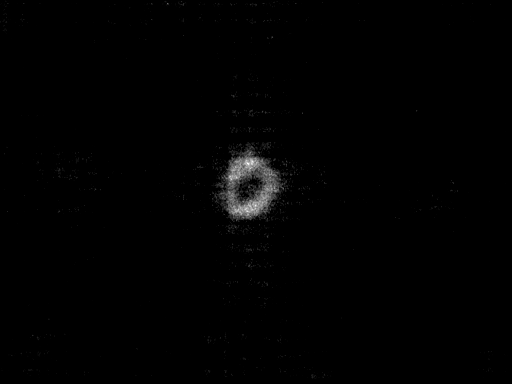} & \includegraphics[width=\figsizeperturb\linewidth,valign=m]{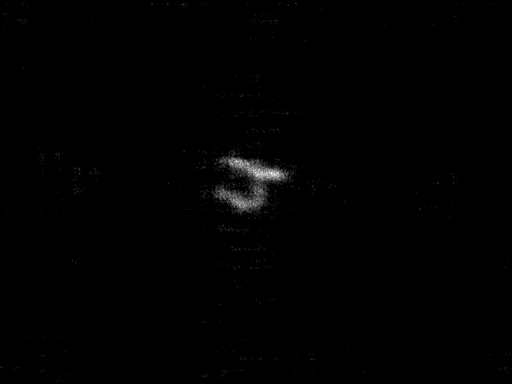}\\
			
		\end{tabular}
	}
	\endgroup
	\caption{Example sensor embeddings for \textit{Fixed mask (m)} in the presence of random rotation (top) and the corresponding reconstruction (bottom) using the convex optimization approach described in \cref{sec:cvx}.}
	\label{tab:mnist_rotate}
\end{figure}

\subsection{Perspective}

While maintaining an object-to-camera distance of \SI{40}{\centi\meter}, perform a random perspective transformation via PyTorch's \texttt{RandomPerspective} with \SI{100}{\percent} probability and a distortion factor of $ 0.5 $.\footnote{\texttt{RandomPerspective} documentation: \url{https://pytorch.org/vision/main/generated/torchvision.transforms.RandomPerspective.html}}

\begin{figure}[h!]
	\begingroup
	\centering
	\renewcommand{\arraystretch}{1} 
	\setlength{\tabcolsep}{0.2em} 
	\scalebox{1.0}{
		\begin{tabular}{ccccc}
			
			\includegraphics[width=\figsizeperturb\linewidth,valign=m]{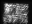}  &
			\includegraphics[width=\figsizeperturb\linewidth,valign=m]{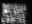}  &
			\includegraphics[width=\figsizeperturb\linewidth,valign=m]{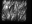} &
			\includegraphics[width=\figsizeperturb\linewidth,valign=m]{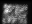}  &
			\includegraphics[width=\figsizeperturb\linewidth,valign=m]{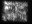}\\[\newlineex]

			\includegraphics[width=\figsizeperturb\linewidth,valign=m]{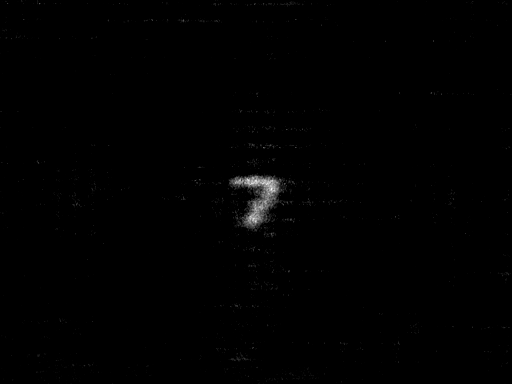} &
			\includegraphics[width=\figsizeperturb\linewidth,valign=m]{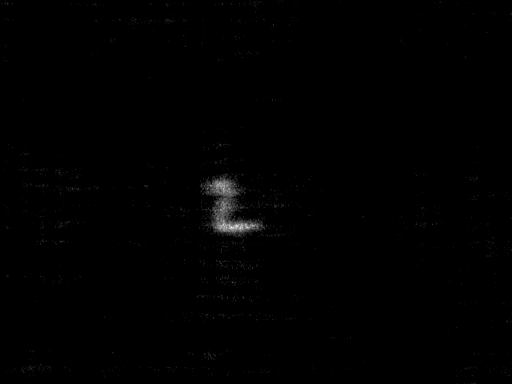}  & \includegraphics[width=\figsizeperturb\linewidth,valign=m]{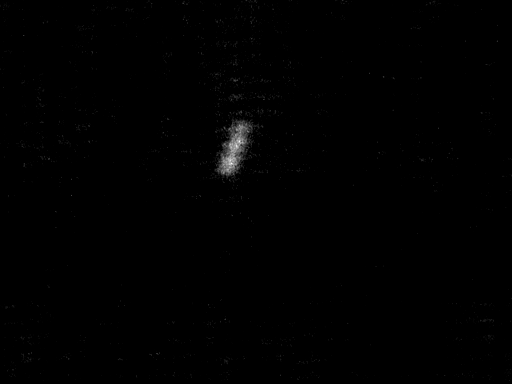} &
			\includegraphics[width=\figsizeperturb\linewidth,valign=m]{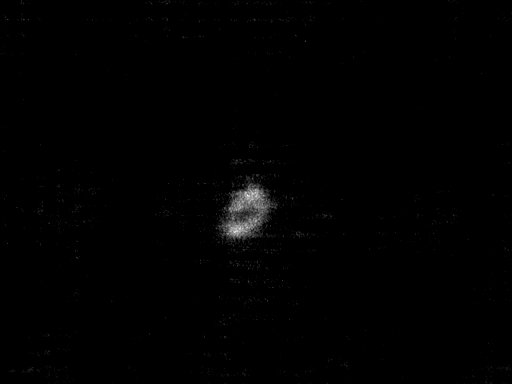} & \includegraphics[width=\figsizeperturb\linewidth,valign=m]{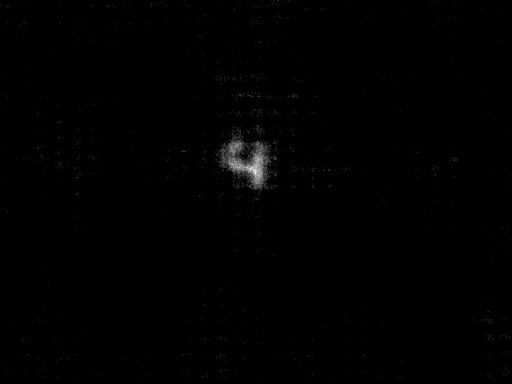}\\
			
		\end{tabular}
	}
	\endgroup
	\caption{Example sensor embeddings for \textit{Fixed mask (m)} in the presence of random perspective distortions  (top) and the corresponding reconstruction (bottom) using the convex optimization approach described in \cref{sec:cvx}.}
	\label{tab:mnist_perspective}
\end{figure}

\section{CelebA -- face attribute binary classification}
\label{sec:celeba_app}

\begin{figure*}[h!]
	\centering
	\begin{subfigure}[b]{.45\textwidth}
		\centering
		\includegraphics[width=0.99\linewidth]{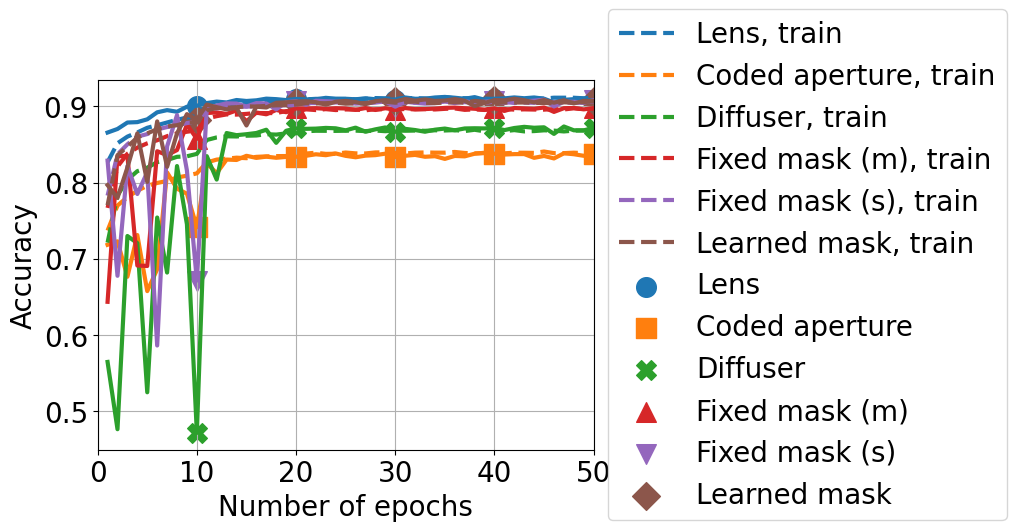}
		\caption{Gender classification, 24$ \times $32.}
		\label{fig:Gender_FCNN_24x32}
	\end{subfigure}
	\begin{subfigure}[b]{.45\textwidth}
		\centering
		\includegraphics[width=0.99\linewidth]{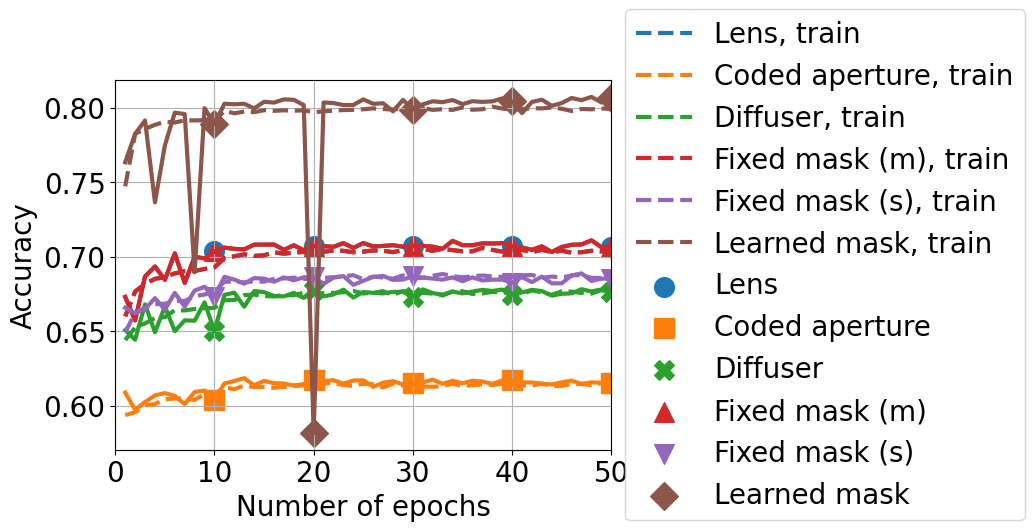}
		\caption{Gender classification, 3$ \times $4.}
		\label{fig:Gender_FCNN_3x4}
	\end{subfigure}\\
	\begin{subfigure}[b]{.45\textwidth}
		\centering
		\includegraphics[width=0.99\linewidth]{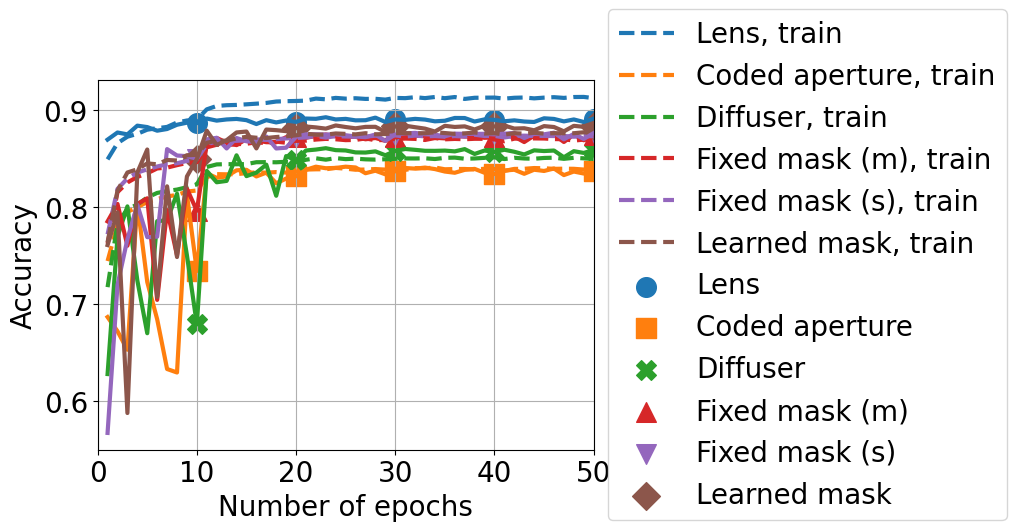}
		\caption{Smiling classification, 24$ \times $32.}
		\label{fig:Smiling_FCNN_24x32}
	\end{subfigure}
	\begin{subfigure}[b]{.45\textwidth}
		\centering
		\includegraphics[width=0.99\linewidth]{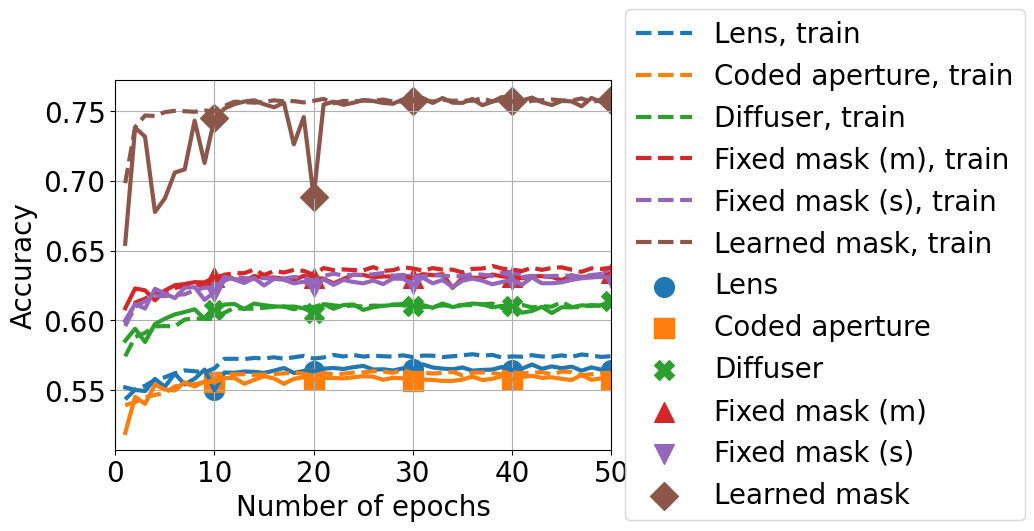}
		\caption{Smiling classification, 3$ \times $4.}
		\label{fig:Smiling_FCNN_3x4}
	\end{subfigure}
	\caption{Train and test curves for CelebA face attribute classification.}
	\label{fig:celeba_curves}
\end{figure*}

\cref{tab:celeba_hparam} details the training hyperparameters for the experiment in \cref{sec:celeba}  that varies the embedding dimension for face attribute classification (CelebA). Example sensor embeddings can be found in \cref{tab:celeba_examples}. Train and test accuracy curves can be see in \cref{fig:celeba_curves}. 

\begin{table}[h!]
	\centering
	\caption{Training hyperparameters for experiment in \cref{sec:celeba} for face attribute classification (CelebA). The FC architecture -- described in \cref{sec:nn} -- was used for all imaging systems and embedding dimensions. \textit{Schedule} denotes after how many epochs the learning rate is reduced by a factor of $ 0.1 $. All models are trained for 50 epochs.}
	\label{tab:celeba_hparam}
	\begin{tabular}{lcc}
		\toprule
		\textit{Encoder} $\downarrow $ & \textit{Batch size} & \textit{Schedule}  \\ 
		\midrule
		Fixed encoders & 64 & 10 \\
		\textit{Learned mask} & 64 & 20 \\
		\bottomrule 
	\end{tabular} 
\end{table}

\newcommand{\figsizeexceleb}{0.19}
\newcommand{\newlineexceleb}{10pt}
\begin{figure}[h!]
	\begingroup
	\centering
	\renewcommand{\arraystretch}{1} 
	\setlength{\tabcolsep}{0.2em} 
	\begin{tabular}{ccccc}
		& 24$\times$32 & 12$\times$16 & 6$\times$8 & 3$\times$4  \\
		
		Lens
		&
		\includegraphics[width=\figsizeexceleb\linewidth,valign=m]{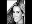}  &
		\includegraphics[width=\figsizeexceleb\linewidth,valign=m]{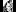}  & \includegraphics[width=\figsizeexceleb\linewidth,valign=m]{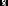} & \includegraphics[width=\figsizeexceleb\linewidth,valign=m]{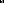}\\[\newlineexceleb]
		\makecell{Coded \\aperture\\\cite{flatcam}} &
		
		\includegraphics[width=\figsizeexceleb\linewidth,valign=m]{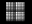}  &
		\includegraphics[width=\figsizeexceleb\linewidth,valign=m]{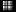}  & \includegraphics[width=\figsizeexceleb\linewidth,valign=m]{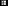} & \includegraphics[width=\figsizeexceleb\linewidth,valign=m]{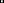}\\[\newlineexceleb]
		\makecell{Diffuser\\\cite{Antipa:18}}
		& 
		
		\includegraphics[width=\figsizeexceleb\linewidth,valign=m]{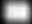}  &
		\includegraphics[width=\figsizeexceleb\linewidth,valign=m]{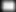}  & \includegraphics[width=\figsizeexceleb\linewidth,valign=m]{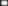} & \includegraphics[width=\figsizeexceleb\linewidth,valign=m]{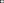}\\[\newlineexceleb]
		\makecell{Fixed\\mask (m)}
		& 
		\includegraphics[width=\figsizeexceleb\linewidth,valign=m]{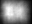}  &
		\includegraphics[width=\figsizeexceleb\linewidth,valign=m]{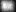}  & \includegraphics[width=\figsizeexceleb\linewidth,valign=m]{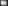} & \includegraphics[width=\figsizeexceleb\linewidth,valign=m]{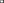}\\[\newlineexceleb]
		\makecell{Fixed\\mask (s)}
		& 
		\includegraphics[width=\figsizeexceleb\linewidth,valign=m]{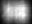}  &
		\includegraphics[width=\figsizeexceleb\linewidth,valign=m]{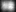} & \includegraphics[width=\figsizeexceleb\linewidth,valign=m]{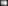} & \includegraphics[width=\figsizeexceleb\linewidth,valign=m]{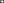}\\[\newlineexceleb]
		\makecell{Learned\\mask}
		&
		\includegraphics[width=\figsizeexceleb\linewidth,valign=m]{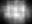}  &
		\includegraphics[width=\figsizeexceleb\linewidth,valign=m]{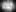} & \includegraphics[width=\figsizeexceleb\linewidth,valign=m]{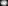} & \includegraphics[width=\figsizeexceleb\linewidth,valign=m]{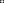}\\
	\end{tabular}
	\endgroup
	\caption{Example sensor embeddings for CelebA.}
	\label{tab:celeba_examples}
\end{figure}

\newpage
\section{CIFAR10 - RGB object classification}
\label{sec:cifar10_app}

\cref{tab:cifar10_hparam} details the training hyperparameters for the experiment in \cref{sec:cifar10}  that varies the embedding dimension for RGB object classification (CIFAR10). Stochastic gradient descent is used as an optimizer, with an initial learning rate of $ 0.01 $. We also apply online augmentation to minimize over-fitting, namely (1) random horizontal flipping and (2) padding and random cropping back to the original embedding dimensions (essentially random shifts). We apply different amount of padding depending on the embedding dimension and encoder, as shown in \cref{tab:cifar10_hparam}. For all embedding dimensions, we resize the embedding to $ (32 \times 32) $ as the VGG architecture typically operates on square inputs that can be downsampled at multiple steps until the classification layer. 

\begin{figure*}[t!]
	\centering
	\begin{subfigure}[b]{.45\textwidth}
		\centering
		\includegraphics[width=0.99\linewidth]{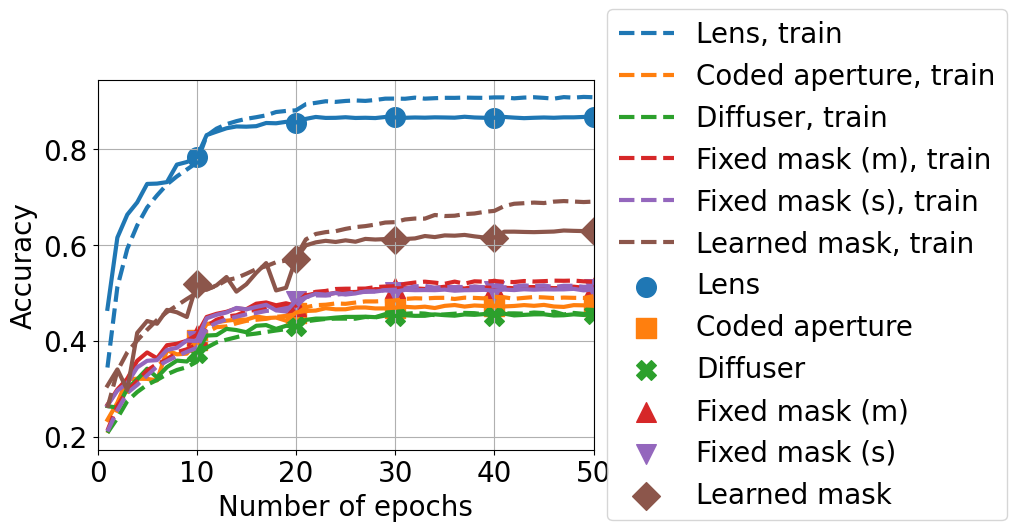}
		\caption{3$ \times $27$ \times $36.}
		\label{fig:CIFAR10_VGG11_27x36}
	\end{subfigure}
	\begin{subfigure}[b]{.45\textwidth}
		\centering
		\includegraphics[width=0.99\linewidth]{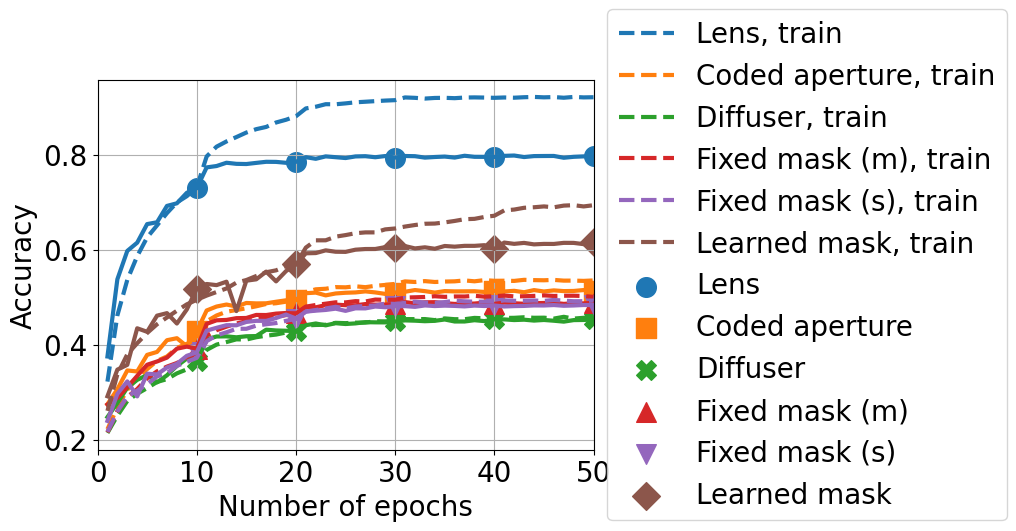}
		\caption{3$ \times $13$ \times $17.}
		\label{fig:CIFAR10_VGG11_13x17}
	\end{subfigure}\\
	\begin{subfigure}[b]{.45\textwidth}
		\centering
		\includegraphics[width=0.99\linewidth]{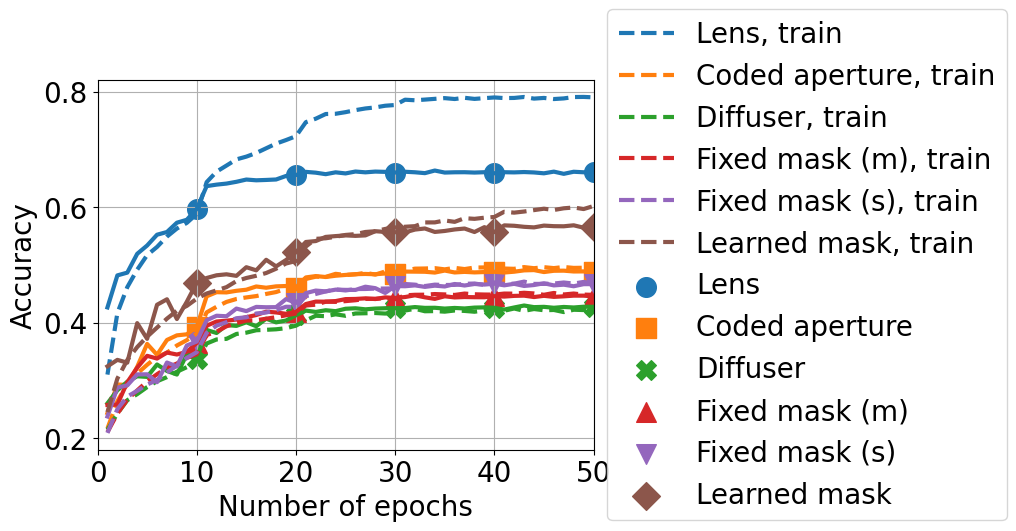}
		\caption{3$ \times $6$ \times $8.}
		\label{fig:CIFAR10_VGG11_6x8}
	\end{subfigure}
	\begin{subfigure}[b]{.45\textwidth}
		\centering
		\includegraphics[width=0.99\linewidth]{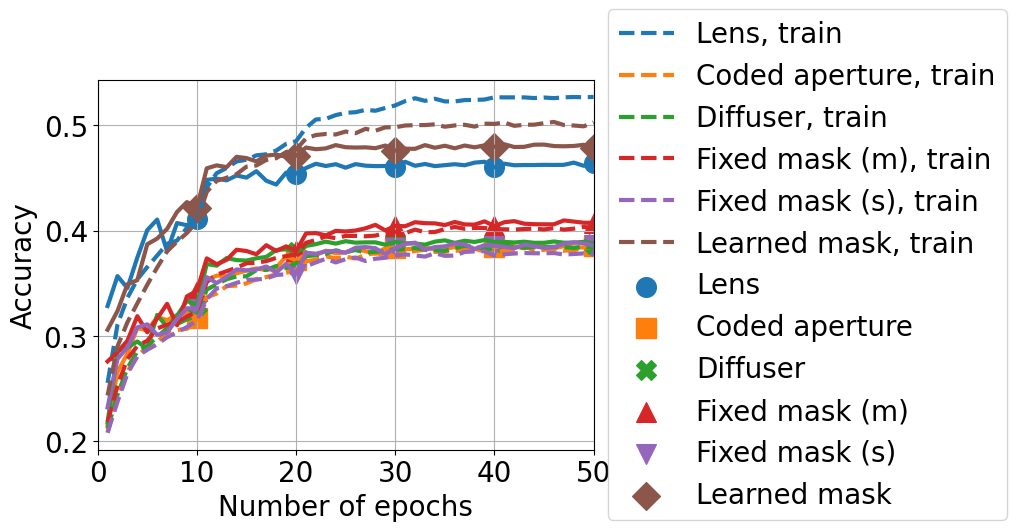}
		\caption{3$ \times $3$ \times $4.}
		\label{fig:CIFAR10_VGG11_3x4}
	\end{subfigure}
	\caption{Train and test curves for CIFAR10 classification, varying embedding dimension.}
	\label{fig:cifar10_curves}
\end{figure*}

\begin{figure*}[t!]
	\centering
	\begin{subfigure}{.35\textwidth}
		\centering
		\includegraphics[width=0.99\linewidth]{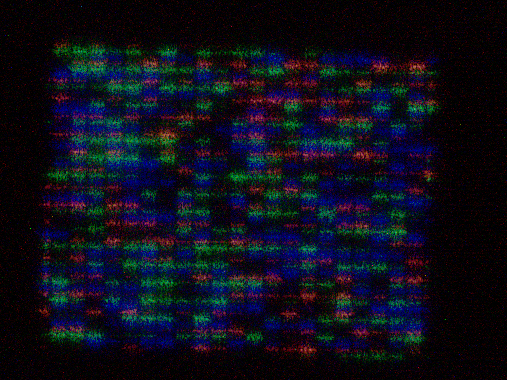}
		\caption{\textit{Fixed mask (m)}.}
		\label{fig:adafruit_psf_rgb}
	\end{subfigure}
	\begin{subfigure}{.35\textwidth}
		\centering
		\includegraphics[width=0.99\linewidth]{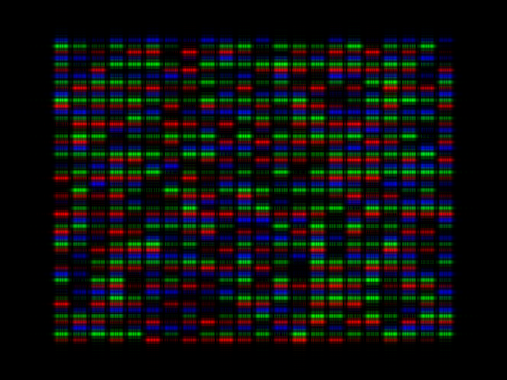}
		\caption{\textit{Fixed mask (s)}.}
		\label{fig:adafruit_sim_psf_rgb}
	\end{subfigure}	
	\caption{RGB point spread functions of \textit{Fixed mask (m)} and \textit{Fixed mask (s)}. Note that different mask patterns are set.}
	\label{fig:rgb_psfs}
\end{figure*}

Train and test accuracy curves can be see in \cref{fig:cifar10_curves}. Example sensor embeddings can be found in \cref{tab:cifar10_examples}. Note that the proposed system has a color filter, such that the measurements of \textit{Fixed mask (m)}, \textit{Fixed mask (s)}, and \textit{Learned mask} do not resemble the ``average'' color of the scene. \cref{tab:learned_psfs} (bottom row) shows the PSFs of the learned masks for each embedding dimension. \cref{fig:rgb_psfs} shows the color PSFs of \textit{Fixed mask (m)} and \textit{Fixed mask (s)}.

\begin{figure*}[t!]
	\centering
	\begin{subfigure}{.45\textwidth}
		\centering
		\includegraphics[width=0.99\linewidth]{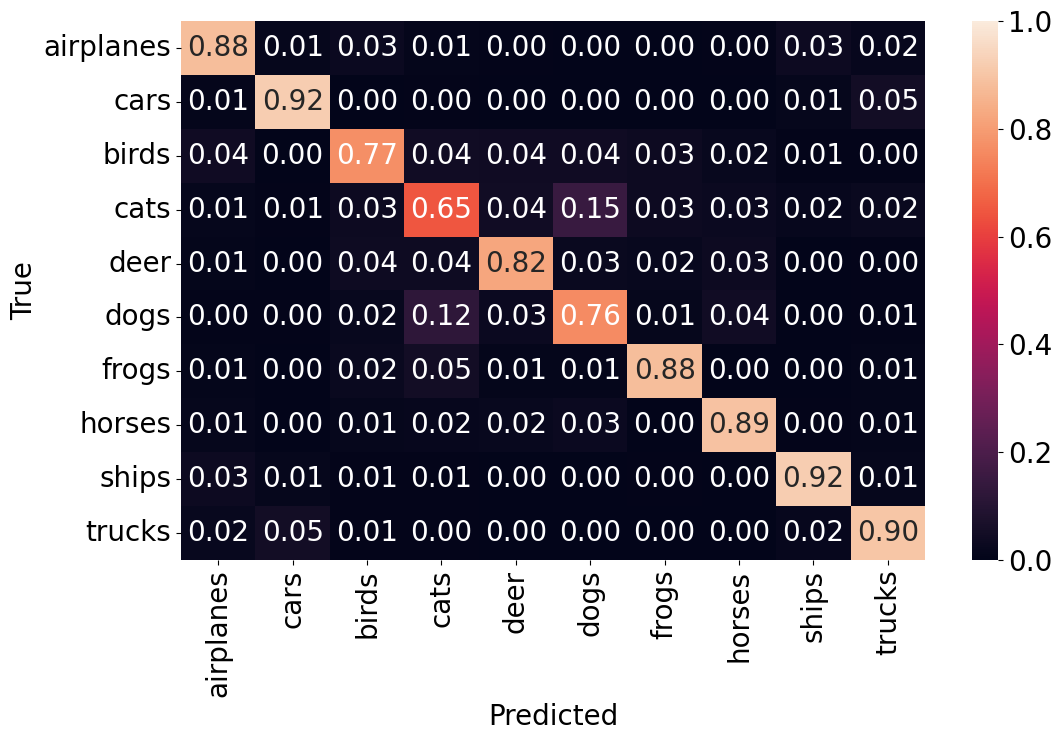}
		\caption{\textit{Lens}.}
		\label{fig:cifar10_lens_confusion}
	\end{subfigure}
	\begin{subfigure}{.45\textwidth}
		\centering
		\includegraphics[width=0.99\linewidth]{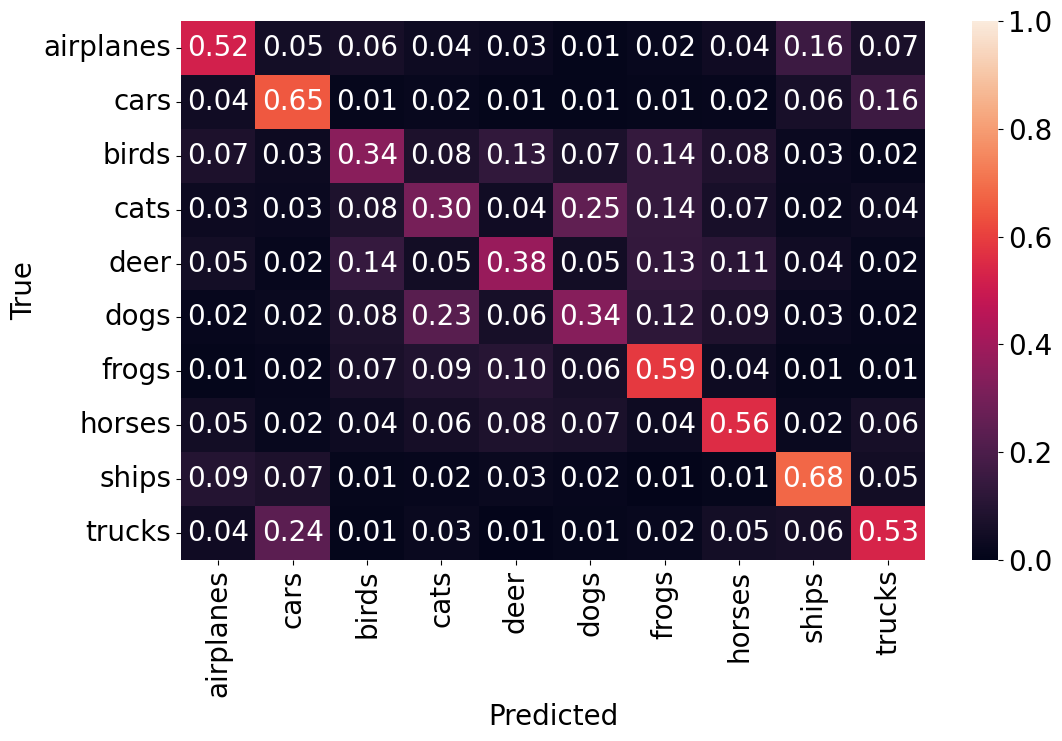}
		\caption{\textit{Fixed mask (m)}.}
		\label{fig:cifar10_lfixed_mask_confusion}
	\end{subfigure}	
	\hfill
	\begin{subfigure}{.45\textwidth}
		\centering
		\includegraphics[width=0.99\linewidth]{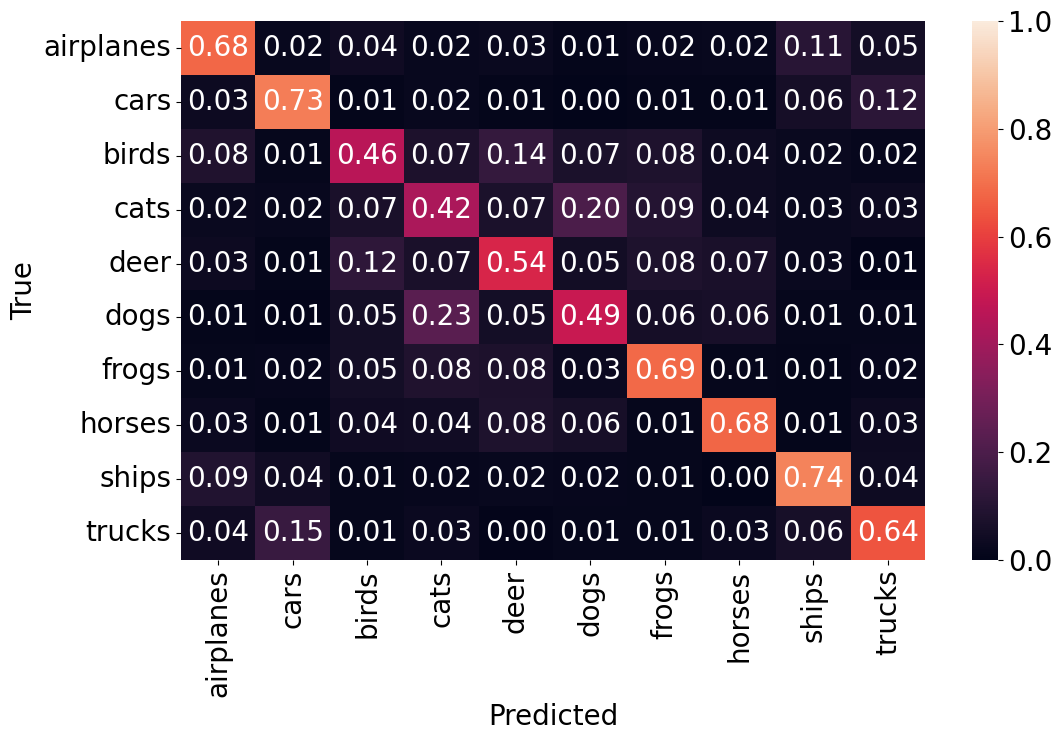}
		\caption{\textit{Learned mask}.}
		\label{fig:cifar10_learned_confusion}
	\end{subfigure}	
	\caption{CIFAR10 confusion matrices for an embedding dimension of $ (27\times 36) $. See \cref{tab:harder_tasks} (right side) for average accuracy.}
	\label{fig:cifar_confusion}
\end{figure*}

\cref{fig:cifar_confusion} shows the confusion matrices for an embedding dimension of $ (26\times 37) $ for \textit{Lens}, \textit{Fixed mask (m)} (the best fixed lensless encoder), and \textit{Learned mask}. For the lensless encoders, there is difficulty in distinguishing \textit{cars-trucks}, \textit{cats-dogs}, and \textit{birds-deer}. The first two mistakes are understandable, the last ones less so. Perhaps it stems from the outdoor setting; although \textit{deer-horse} would have been more understandable.

\begin{table}[h]
	\centering
	\caption{Training hyperparameters for experiment in \cref{sec:cifar10} for RGB object classification (CIFAR10). The VGG11 architecture -- described in \cref{sec:vgg11} -- was used for all imaging systems and embedding dimensions. \textit{Schedule} denotes after how many epochs the learning rate is reduced by a factor of $ 0.1 $. All models are trained for 50 epochs. \textit{Pad} denotes how much the input image was padded prior to random cropping for (3$ \times $ 27$ \times $36), (3$ \times $ 13$ \times $17), (3$ \times $ 6$ \times $8), and (3$ \times $ 3$ \times $4) respectively.}
	\label{tab:cifar10_hparam}
	\begin{tabular}{lccc}
		\toprule
		\textit{Encoder} $\downarrow $ & \textit{Batch size} & \textit{Schedule}  & \textit{Pad}  \\ 
		\midrule
		\textit{Lens} & 32 & 10 & 4-4-2-0 \\
		Fixed lensless & 32 & 10  & 1-1-1-0\\
		\textit{Learned mask} & 32 & 20  & 2-1-1-0 \\
		\bottomrule 
	\end{tabular} 
\end{table}

\newcommand{\figsizeexcifar}{0.1}
\newcommand{\newlineexcifar}{20pt}
\begin{figure*}[t!]
	\begingroup
	\centering
	\renewcommand{\arraystretch}{1} 
	\setlength{\tabcolsep}{0.2em} 
	\begin{tabular}{c cccc cccc}
		& \multicolumn{2}{c}{27$\times$36}  & \multicolumn{2}{c}{13$\times$17} & \multicolumn{2}{c}{6$\times$8} & \multicolumn{2}{c}{3$\times$4}  \\
		\cmidrule(r){2-3} \cmidrule(r){4-5} \cmidrule(r){6-7} \cmidrule(r){8-9} 
		\newline
		\multirow{2}{*}{Lens}
		&
		\includegraphics[width=\figsizeexcifar\linewidth,valign=m]{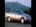}  &
		\includegraphics[width=\figsizeexcifar\linewidth,valign=m]{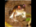}  & \includegraphics[width=\figsizeexcifar\linewidth,valign=m]{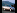} & \includegraphics[width=\figsizeexcifar\linewidth,valign=m]{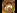} &
		\includegraphics[width=\figsizeexcifar\linewidth,valign=m]{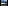}  &
		\includegraphics[width=\figsizeexcifar\linewidth,valign=m]{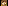}  & \includegraphics[width=\figsizeexcifar\linewidth,valign=m]{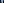} & \includegraphics[width=\figsizeexcifar\linewidth,valign=m]{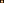}\\[12pt]
		& \includegraphics[width=\figsizeexcifar\linewidth,valign=m]{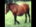}  &
		\includegraphics[width=\figsizeexcifar\linewidth,valign=m]{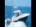}  & \includegraphics[width=\figsizeexcifar\linewidth,valign=m]{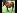} & \includegraphics[width=\figsizeexcifar\linewidth,valign=m]{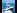} &
		\includegraphics[width=\figsizeexcifar\linewidth,valign=m]{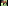}  &
		\includegraphics[width=\figsizeexcifar\linewidth,valign=m]{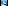}  & \includegraphics[width=\figsizeexcifar\linewidth,valign=m]{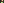} & \includegraphics[width=\figsizeexcifar\linewidth,valign=m]{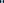}\\[\newlineexcifar]
		
		\multirow{2}{*}{\makecell{Coded \\aperture\\\cite{flatcam}}} &
		\includegraphics[width=\figsizeexcifar\linewidth,valign=m]{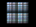}  &
		\includegraphics[width=\figsizeexcifar\linewidth,valign=m]{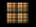}  & \includegraphics[width=\figsizeexcifar\linewidth,valign=m]{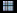} & \includegraphics[width=\figsizeexcifar\linewidth,valign=m]{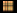} &
		\includegraphics[width=\figsizeexcifar\linewidth,valign=m]{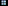}  &
		\includegraphics[width=\figsizeexcifar\linewidth,valign=m]{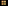}  & \includegraphics[width=\figsizeexcifar\linewidth,valign=m]{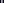} & \includegraphics[width=\figsizeexcifar\linewidth,valign=m]{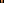}\\[12pt]
		&
		\includegraphics[width=\figsizeexcifar\linewidth,valign=m]{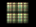}  &
		\includegraphics[width=\figsizeexcifar\linewidth,valign=m]{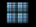}  & \includegraphics[width=\figsizeexcifar\linewidth,valign=m]{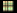} & \includegraphics[width=\figsizeexcifar\linewidth,valign=m]{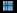} &
		\includegraphics[width=\figsizeexcifar\linewidth,valign=m]{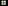}  &
		\includegraphics[width=\figsizeexcifar\linewidth,valign=m]{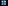}  & \includegraphics[width=\figsizeexcifar\linewidth,valign=m]{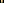} & \includegraphics[width=\figsizeexcifar\linewidth,valign=m]{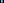}\\[\newlineexcifar]
		
		\multirow{2}{*}{\makecell{Diffuser~\cite{Antipa:18}}} & 
		\includegraphics[width=\figsizeexcifar\linewidth,valign=m]{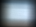}  &
		\includegraphics[width=\figsizeexcifar\linewidth,valign=m]{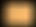}  & \includegraphics[width=\figsizeexcifar\linewidth,valign=m]{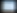} & \includegraphics[width=\figsizeexcifar\linewidth,valign=m]{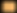} &
		\includegraphics[width=\figsizeexcifar\linewidth,valign=m]{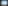}  &
		\includegraphics[width=\figsizeexcifar\linewidth,valign=m]{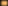}  & \includegraphics[width=\figsizeexcifar\linewidth,valign=m]{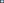} & \includegraphics[width=\figsizeexcifar\linewidth,valign=m]{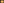}\\[12pt]
		& 
		\includegraphics[width=\figsizeexcifar\linewidth,valign=m]{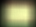}  &
		\includegraphics[width=\figsizeexcifar\linewidth,valign=m]{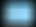}  & \includegraphics[width=\figsizeexcifar\linewidth,valign=m]{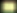} & \includegraphics[width=\figsizeexcifar\linewidth,valign=m]{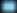} &
		\includegraphics[width=\figsizeexcifar\linewidth,valign=m]{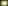}  &
		\includegraphics[width=\figsizeexcifar\linewidth,valign=m]{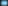}  & \includegraphics[width=\figsizeexcifar\linewidth,valign=m]{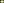} & \includegraphics[width=\figsizeexcifar\linewidth,valign=m]{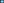}\\[\newlineexcifar]
		
		\multirow{2}{*}{\makecell{Fixed\\mask (m)}}
		& 
		\includegraphics[width=\figsizeexcifar\linewidth,valign=m]{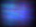}  &
		\includegraphics[width=\figsizeexcifar\linewidth,valign=m]{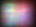}  & \includegraphics[width=\figsizeexcifar\linewidth,valign=m]{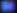} & \includegraphics[width=\figsizeexcifar\linewidth,valign=m]{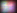} &
		\includegraphics[width=\figsizeexcifar\linewidth,valign=m]{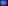}  &
		\includegraphics[width=\figsizeexcifar\linewidth,valign=m]{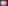}  & \includegraphics[width=\figsizeexcifar\linewidth,valign=m]{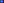} & \includegraphics[width=\figsizeexcifar\linewidth,valign=m]{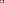}\\[12pt]
		& 
		\includegraphics[width=\figsizeexcifar\linewidth,valign=m]{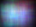}  &
		\includegraphics[width=\figsizeexcifar\linewidth,valign=m]{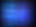}  & \includegraphics[width=\figsizeexcifar\linewidth,valign=m]{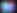} & \includegraphics[width=\figsizeexcifar\linewidth,valign=m]{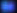} &
		\includegraphics[width=\figsizeexcifar\linewidth,valign=m]{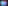}  &
		\includegraphics[width=\figsizeexcifar\linewidth,valign=m]{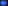}  & \includegraphics[width=\figsizeexcifar\linewidth,valign=m]{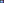} & \includegraphics[width=\figsizeexcifar\linewidth,valign=m]{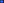}\\[\newlineexcifar]
		
		\multirow{2}{*}{\makecell{Fixed\\mask (s)}}
		& 
		\includegraphics[width=\figsizeexcifar\linewidth,valign=m]{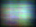}  &
		\includegraphics[width=\figsizeexcifar\linewidth,valign=m]{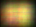}  & \includegraphics[width=\figsizeexcifar\linewidth,valign=m]{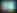} & \includegraphics[width=\figsizeexcifar\linewidth,valign=m]{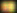} &
		\includegraphics[width=\figsizeexcifar\linewidth,valign=m]{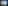}  &
		\includegraphics[width=\figsizeexcifar\linewidth,valign=m]{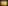}  & \includegraphics[width=\figsizeexcifar\linewidth,valign=m]{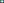} & \includegraphics[width=\figsizeexcifar\linewidth,valign=m]{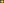}\\[12pt]
		& 
		\includegraphics[width=\figsizeexcifar\linewidth,valign=m]{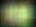}  &
		\includegraphics[width=\figsizeexcifar\linewidth,valign=m]{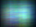}  & \includegraphics[width=\figsizeexcifar\linewidth,valign=m]{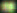} & \includegraphics[width=\figsizeexcifar\linewidth,valign=m]{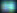} &
		\includegraphics[width=\figsizeexcifar\linewidth,valign=m]{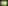}  &
		\includegraphics[width=\figsizeexcifar\linewidth,valign=m]{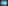}  & \includegraphics[width=\figsizeexcifar\linewidth,valign=m]{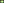} & \includegraphics[width=\figsizeexcifar\linewidth,valign=m]{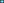}\\[\newlineexcifar]
		
		\multirow{2}{*}{\makecell{Learned\\mask}}
		&
		\includegraphics[width=\figsizeexcifar\linewidth,valign=m]{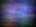}  &
		\includegraphics[width=\figsizeexcifar\linewidth,valign=m]{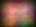}  & \includegraphics[width=\figsizeexcifar\linewidth,valign=m]{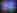} & \includegraphics[width=\figsizeexcifar\linewidth,valign=m]{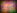} &
		\includegraphics[width=\figsizeexcifar\linewidth,valign=m]{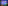}  &
		\includegraphics[width=\figsizeexcifar\linewidth,valign=m]{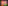}  & \includegraphics[width=\figsizeexcifar\linewidth,valign=m]{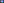} & \includegraphics[width=\figsizeexcifar\linewidth,valign=m]{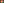}\\[12pt]
		&
		\includegraphics[width=\figsizeexcifar\linewidth,valign=m]{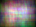}  &
		\includegraphics[width=\figsizeexcifar\linewidth,valign=m]{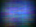}  & \includegraphics[width=\figsizeexcifar\linewidth,valign=m]{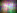} & \includegraphics[width=\figsizeexcifar\linewidth,valign=m]{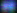} &
		\includegraphics[width=\figsizeexcifar\linewidth,valign=m]{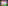}  &
		\includegraphics[width=\figsizeexcifar\linewidth,valign=m]{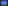}  & \includegraphics[width=\figsizeexcifar\linewidth,valign=m]{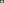} & \includegraphics[width=\figsizeexcifar\linewidth,valign=m]{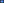}\\[\newlineexcifar]
	\end{tabular}
	\endgroup
	\caption{Example sensor embeddings for CIFAR10; \textit{car}, \textit{frog}, \textit{ship}, and \textit{ship} examples (clockwise from top-left). Note that the proposed system has a color filter, such that the measurements of \textit{Fixed mask (m)}, \textit{Fixed mask (s)}, and \textit{Learned mask} do not resemble the ``average'' color of the scene. \cref{tab:learned_psfs} (bottom row) shows the PSFs of the learned masks for each embedding dimension. \cref{fig:rgb_psfs} shows the color PSFs of \textit{Fixed mask (m)} and \textit{Fixed mask (s)}.
	}
	\label{tab:cifar10_examples}
\end{figure*}

\newpage
~
\newpage
~
\newpage
~
\newpage
~
\newpage
~
\newpage
~
\newpage
~
\newpage
\section{Adversarial attacks}

\subsection{Convex optimization to recover object from downsampled lensless camera measurement}
\label{sec:cvx}

\begin{equation}\label{eq:cvx}
	\hat{\bm{x}}	= \argmin_{\bm{x}\geq 0}\dfrac{1}{2}	\|\bm{y} - \bm{D} \bm{H}\bm{x} \|_2^2,
\end{equation}
where $ \bm{y} $ is the raw measurement, $ \bm{D} $ is a downsampling operation whose output size corresponds to the sensor resolution, $ \bm{H} $ is the high-resolution point spread function (PSF) of the system, $ \bm{x} $ is the underlying object. This problem can be solved through projected gradient descent~\cite{Combettes2011}.

A common approach in inverse modeling is to add a prior to the loss. This also serves to regularize the solution. We tried adding the total variation (TV) norm \eg $ \|\Psi \bm{x}\|_1 $ where $ \Psi $ is the finite-difference operator (as in~\cite{Antipa:18}), but found results to be slightly worse (a bit blurrier) for non-sparse data such as faces.

For the first adversary experiment in \cref{sec:defense} (example attacks in \cref{fig:example_cvx_recon} and results in \cref{tab:cvx_attack}): we simulate examples with \cref{alg:sim} to obtain several $ \bm{y} $, and use the Pycsou library~\cite{matthieu_simeoni_2021_4715243} to solve \cref{eq:cvx} numerically. For the correct PSF scenario, we use the same PSF used in simulating the example. For the bad PSF scenario, we randomly set the mask values and simulate a new PSF with the approach described in \cref{sec:sim_psf}. This new PSF is used in solving \cref{eq:cvx}.

\subsection{Plaintext attack generator}
\label{sec:generator}

In this section, we describe the training details for the second adversary experiment in \cref{sec:defense} that attempts to reconstruct embedding dimensions of faces through a trained generator.

\paragraph{Dataset.}

We use the CelebA dataset~\cite{liu2015faceattributes} to train the decoder, namely a separate 100K files than were used to train the masks as in \cref{sec:celeba}. All files are passed through the proposed imaging system with 1, 10, and 100 varying masks to produce 100K embeddings.
When varying the number of plaintext attacks to produce the results in \cref{tab:celeba_decoder}, a subset of these 100K embeddings are used to train the generator with a train-test split of 85-15.

\begin{figure}[t!]
	\centering
	\includegraphics[width=0.99\linewidth]{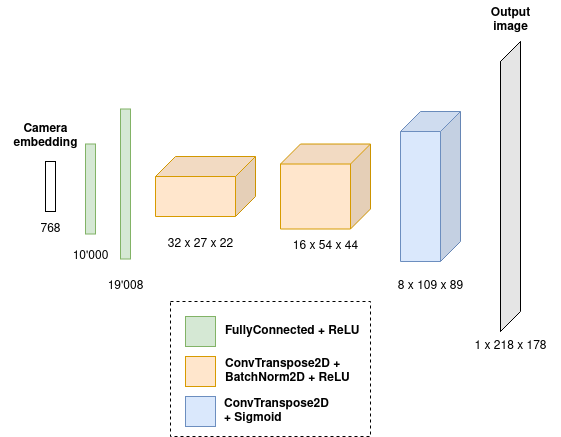}
	\caption{Architecture for plaintext generator.}
	\label{fig:cvpr22_generator}
\end{figure}

\paragraph{Architecture.}
\cref{fig:cvpr22_generator} shows the architecture used for generating faces, which consists of:
\begin{enumerate}
	\tightlist
	\item Flatten sensor embedding.
	\item Two-layer fully connected network with ReLU activations ($ 10'000 $ and $ 19'008 $ hidden units).
	\item Reshaping output to a cube of dimension $ (32\times 27\times 22) $.
	\item Two 2D transposed convolution operators~\cite{5539957} -- upsampling followed by convolution with a set of $ (3\times 3) $ filters -- followed by 2D batch normalization and ReLU activation.
	\item A final 2D transposed convolution operator with a sigmoid activation to restrict the output image pixels to $ [0, 1] $.
\end{enumerate}

Our architecture is inspired by that of autoencoders and generative adversarial networks~\cite{karras2017progressive} (fully connected layers followed by convolutional layers).\footnote{\url{https://medium.com/dataseries/convolutional-autoencoder-in-pytorch-on-mnist-dataset-d65145c132ac\#63b2}; \url{https://machinelearningmastery.com/upsampling-and-transpose-convolution-layers-for-generative-adversarial-networks/}}

We also tried the following but they did not produce better results:
\begin{itemize}
	\tightlist
	\item Increasing the number of (ConvTranspose2D + BatchNorm2d + ReLU) layers.
	\item Training a fully-connected layer prior to a (frozen and unfrozen) pre-trained StyleGAN2~\cite{Karras2020ada}.
\end{itemize}

\paragraph{Training hyperparameters.}

Due to the significant difference in number of training examples, we use different batch sizes: 4 for the 100 plaintext attacks scenario, 16 for 1K, and 32 for the 10K and 100K. All models were training with stochastic gradient descent and a learning rate of $ 0.01 $.

\begin{figure*}[t!]
	\centering
	\begin{subfigure}[b]{.45\textwidth}
		\centering
		\includegraphics[width=0.99\linewidth]{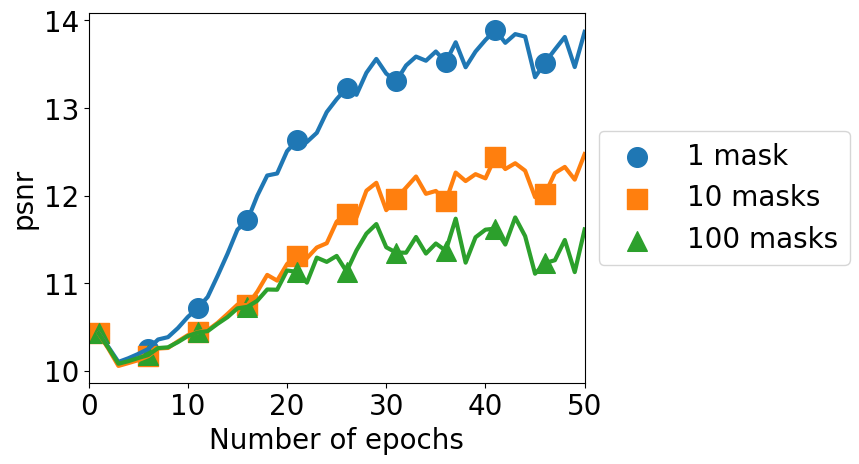}
		\caption{PSNR, 100 plaintext attacks.}
		\label{fig:decoder_psnr_100examples}
	\end{subfigure}
	\begin{subfigure}[b]{.45\textwidth}
		\centering
		\includegraphics[width=0.99\linewidth]{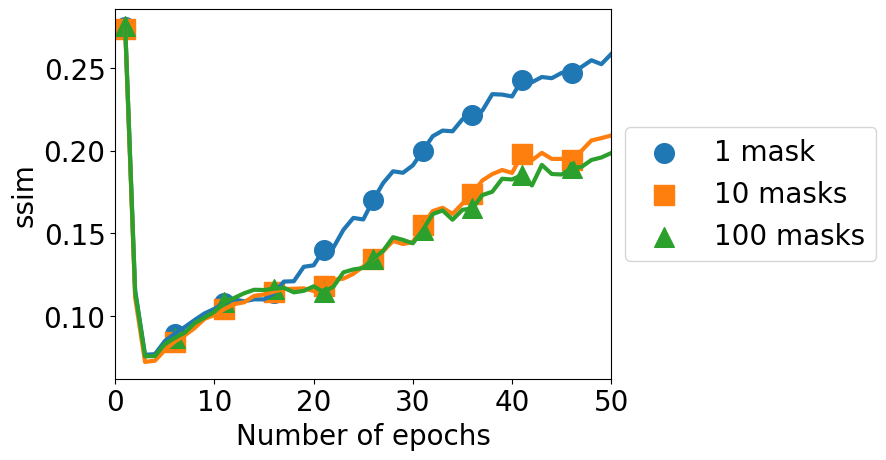}
		\caption{SSIM, 100 plaintext attacks.}
		\label{fig:decoder_ssim_100examples}
	\end{subfigure}\\
	
	\begin{subfigure}[b]{.45\textwidth}
		\centering
		\includegraphics[width=0.99\linewidth]{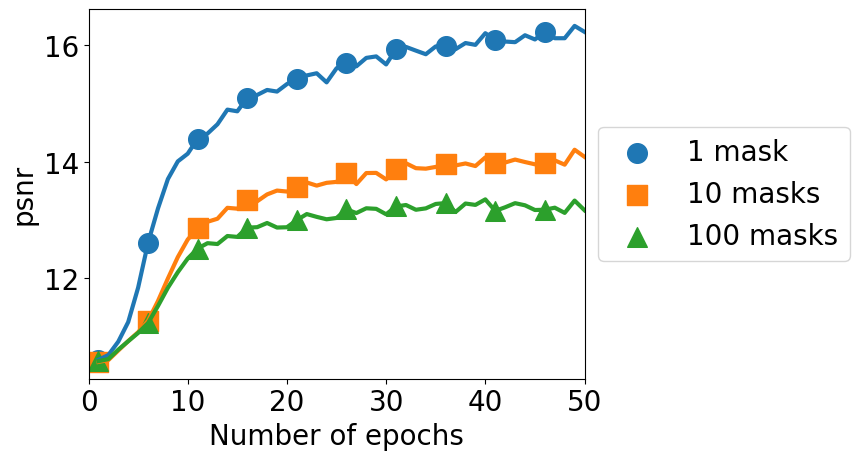}
		\caption{PSNR, 1K plaintext attacks.}
		\label{fig:decoder_psnr_1000examples}
	\end{subfigure}
	\begin{subfigure}[b]{.45\textwidth}
		\centering
		\includegraphics[width=0.99\linewidth]{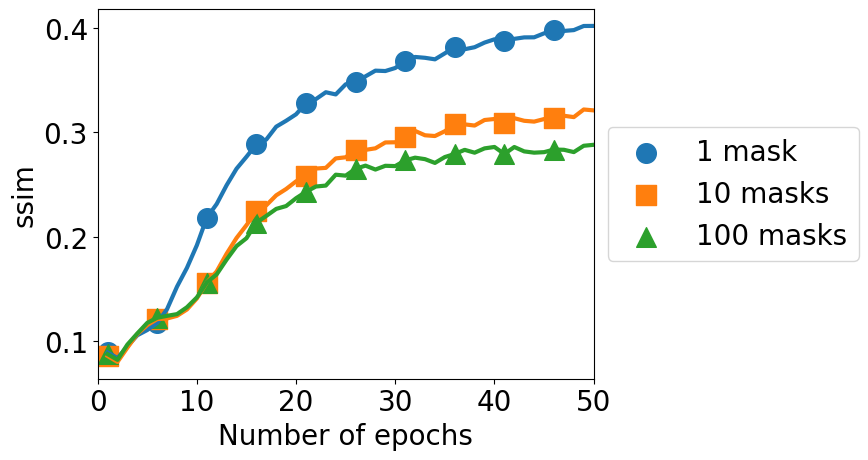}
		\caption{SSIM, 1K plaintext attacks.}
		\label{fig:decoder_ssim_1000examples}
	\end{subfigure}\\
	
	\begin{subfigure}[b]{.45\textwidth}
		\centering
		\includegraphics[width=0.99\linewidth]{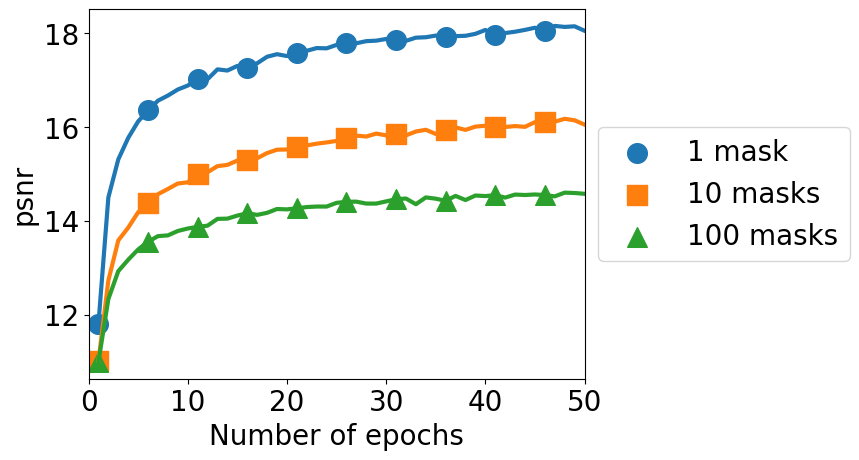}
		\caption{PSNR, 10K plaintext attacks.}
		\label{fig:decoder_psnr_10000examples}
	\end{subfigure}
	\begin{subfigure}[b]{.45\textwidth}
		\centering
		\includegraphics[width=0.99\linewidth]{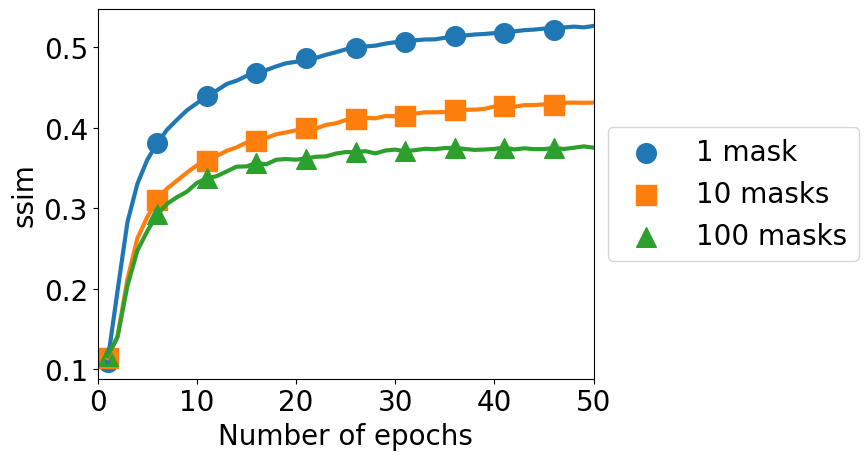}
		\caption{SSIM, 10K plaintext attacks.}
		\label{fig:decoder_ssim_10000examples}
	\end{subfigure}\\

	\begin{subfigure}[b]{.45\textwidth}
		\centering
		\includegraphics[width=0.99\linewidth]{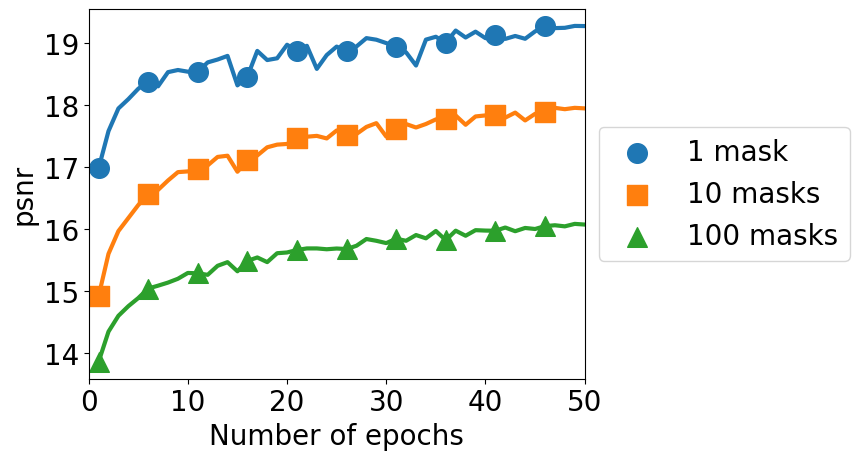}
		\caption{PSNR, 100K plaintext attacks.}
		\label{fig:decoder_psnr_100000examples}
	\end{subfigure}
	\begin{subfigure}[b]{.45\textwidth}
		\centering
		\includegraphics[width=0.99\linewidth]{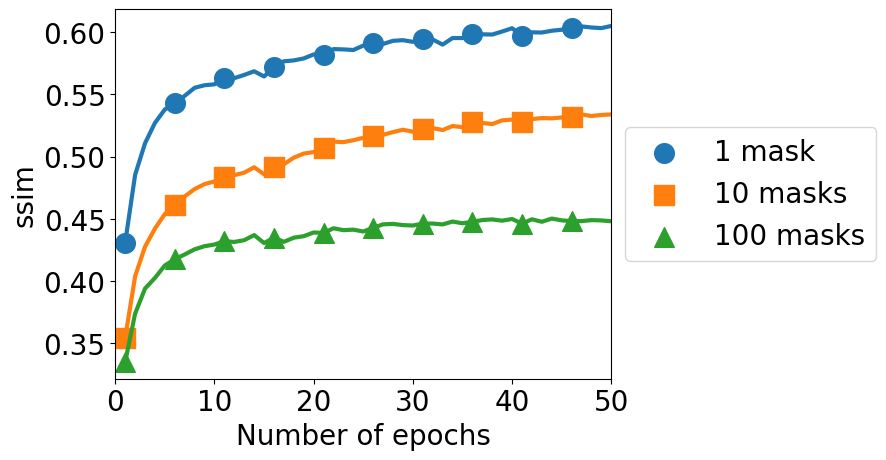}
		\caption{SSIM, 100K plaintext attacks.}
		\label{fig:decoder_ssim_100000examples}
	\end{subfigure}
	
	\caption{PSNR and SSIM curves for training CelebA generator.}
	\label{fig:celeba_decoder_curves}
\end{figure*}

PSNR and SSIM test curves can be seen in \cref{fig:celeba_decoder_curves}, demonstrating the gap in performance as the number of varying masks increases. Note that the large peak in SSIM at the start of 100 plaintext attacks --\cref{fig:decoder_ssim_100examples} -- occurs as the network has not been exposed to enough training examples due to the small dataset size.

\paragraph{For 100 masks.}

\begin{figure*}[t!]
	\centering
	\begin{subfigure}[b]{.45\textwidth}
		\centering
		\includegraphics[width=0.99\linewidth]{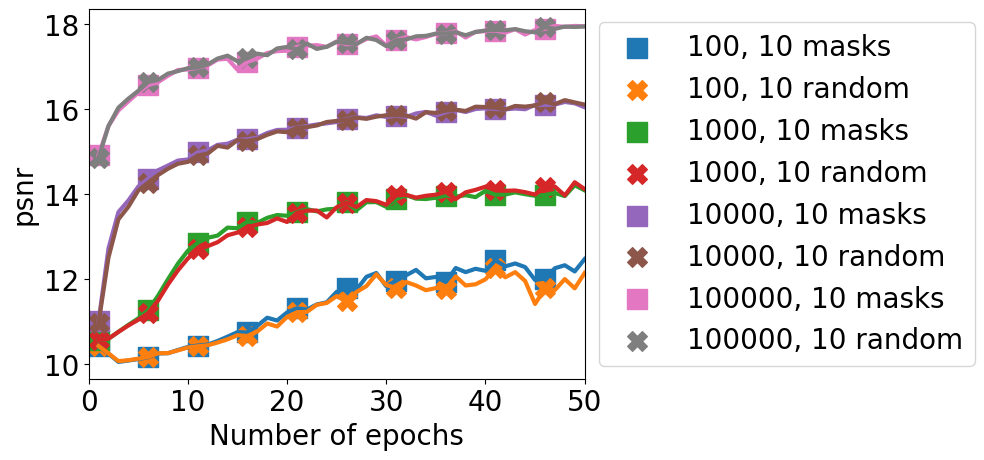}
		\caption{PSNR.}
		\label{fig:decoder_psnr}
	\end{subfigure}
	\begin{subfigure}[b]{.45\textwidth}
		\centering
		\includegraphics[width=0.99\linewidth]{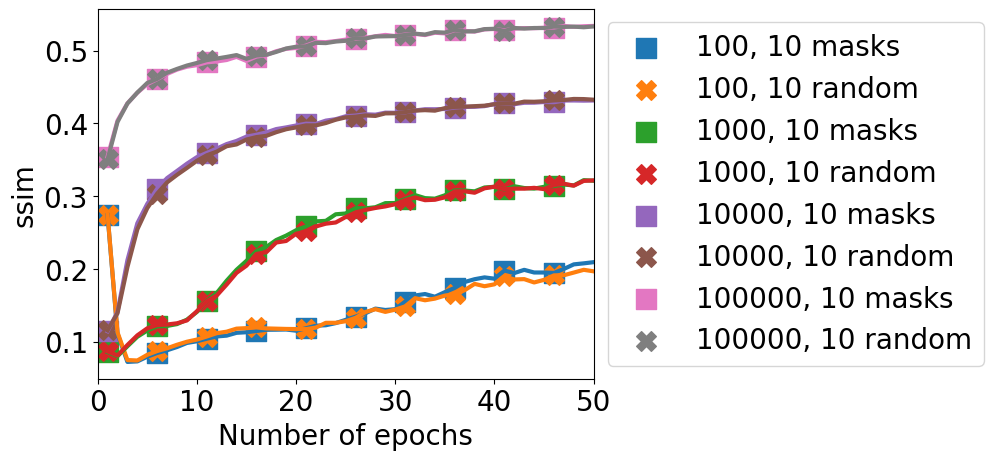}
		\caption{SSIM.}
		\label{fig:decoder_ssim}
	\end{subfigure}
	\caption{Peak signal-to-noise ratio (PSNR) and structural similarity (SSIM) test curves for generators of CelebA; trained with varying number of masks in the imaging system.}
	\label{fig:10_masks}
\end{figure*}

Due to long training times for the hybrid approach, we use a set of random masks for the 100 mask case. From our experiements we have observed similar decoder performance for learned or random masks. \cref{tab:decoder_10_mask} compares the image quality metrics for 10 learned and random masks that are varying. As the results and test curves (see \cref{fig:10_masks}) are very similar, we expect the same for 100 masks.
\begin{table}[h!]
	\caption{Image quality (PSNR / SSIM) of trained decoder for varying number of plaintext attacks and number of varying masks. Higher PSNR/SSIM is better.}
	\label{tab:decoder_10_mask}
	\centering
	\scalebox{0.9}{
		\begin{tabular}{l c c}
			\toprule
			\textit{\# attacks}   $\downarrow $ & 10 (learned) & 10 (random) \\
			\midrule
			$ 100 $    &  $12.5 $  / $  0.21 $& $ 12.3 $ / $ 0.20 $   \\
			$ 1'000 $  & $ 14.2$  / $ 0.32 $& $ 14.3 $ / $ 0.32$ \\
			$ 10'000  $  & $16.2$  / $ 0.43 $& $ 16.2 $ / $0.43 $ \\
			$ 100'000  $ & $ 18.0$  / $ 0.53 $& $ 17.9 $ / $ 0.53$ \\
			\bottomrule
	\end{tabular}}
\end{table}

\paragraph{Example generated outputs.}
\label{sec:example_generated}

\begin{figure}[h]
	\centering
	\stackunder[4pt]{Original}{\includegraphics[width=\celebfig\linewidth,valign=m]{figs/celeba_decoder/original_0.png}}
	\stackunder[6pt]{Fixed}{\includegraphics[width=\celebfig\linewidth,valign=m]{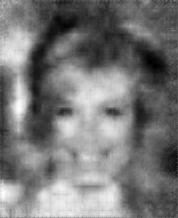}}
	\stackunder[6pt]{10 masks}{\includegraphics[width=\celebfig\linewidth,valign=m]{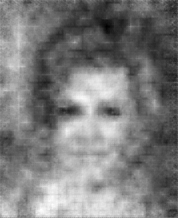}}
	\stackunder[6pt]{100 masks}{\includegraphics[width=\celebfig\linewidth,valign=m]{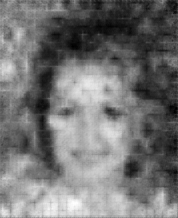}} \\[\newlineceleb]
	
	\includegraphics[width=\celebfig\linewidth,valign=m]{figs/celeba_decoder/original_1.png} \includegraphics[width=\celebfig\linewidth,valign=m]{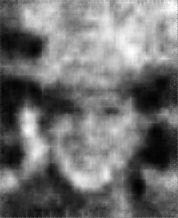}
	\includegraphics[width=\celebfig\linewidth,valign=m]{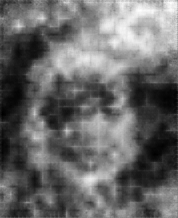}
	\includegraphics[width=\celebfig\linewidth,valign=m]{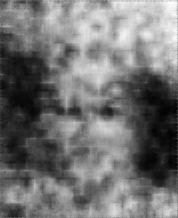}\\[\newlineceleb]
	
	\includegraphics[width=\celebfig\linewidth,valign=m]{figs/celeba_decoder/original_3.png} \includegraphics[width=\celebfig\linewidth,valign=m]{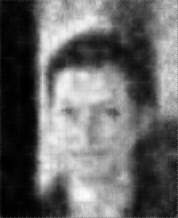}
	\includegraphics[width=\celebfig\linewidth,valign=m]{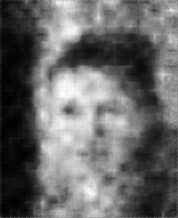}
	\includegraphics[width=\celebfig\linewidth,valign=m]{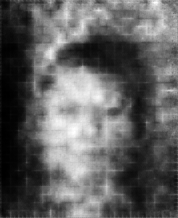}\\
	
	\caption{Example outputs of a decoder that was trained with $ 10'000 $ plaintext attacks of embeddings of resolution $ (24\times 32) $.}
	\label{fig:celeba_decoder_10k}
\end{figure}

\begin{figure}[h]
	\centering
	\stackunder[4pt]{Original}{\includegraphics[width=\celebfig\linewidth,valign=m]{figs/celeba_decoder/original_0.png}}
	\stackunder[6pt]{Fixed}{\includegraphics[width=\celebfig\linewidth,valign=m]{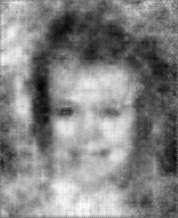}}
	\stackunder[6pt]{10 masks}{\includegraphics[width=\celebfig\linewidth,valign=m]{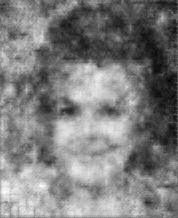}}
	\stackunder[6pt]{100 masks}{\includegraphics[width=\celebfig\linewidth,valign=m]{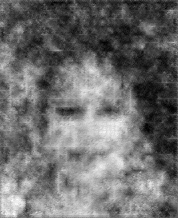}} \\[\newlineceleb]
	
	\includegraphics[width=\celebfig\linewidth,valign=m]{figs/celeba_decoder/original_1.png} \includegraphics[width=\celebfig\linewidth,valign=m]{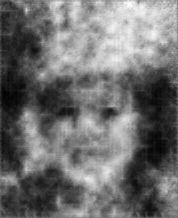}
	\includegraphics[width=\celebfig\linewidth,valign=m]{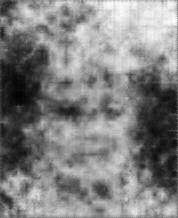}
	\includegraphics[width=\celebfig\linewidth,valign=m]{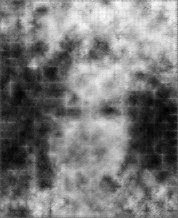}\\[\newlineceleb]
	
	\includegraphics[width=\celebfig\linewidth,valign=m]{figs/celeba_decoder/original_3.png} \includegraphics[width=\celebfig\linewidth,valign=m]{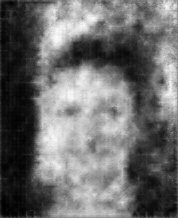}
	\includegraphics[width=\celebfig\linewidth,valign=m]{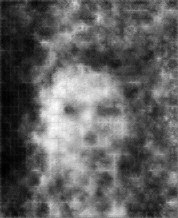}
	\includegraphics[width=\celebfig\linewidth,valign=m]{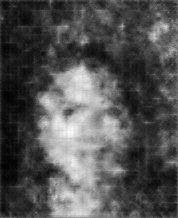}\\
	
	\caption{Example outputs of a decoder that was trained with $ 1'000 $ plaintext attacks of embeddings of resolution $ (24\times 32) $.}
	\label{fig:celeba_decoder_1k}
\end{figure}

\begin{figure}[h]
	\centering
	\stackunder[4pt]{Original}{\includegraphics[width=\celebfig\linewidth,valign=m]{figs/celeba_decoder/original_0.png}}
	\stackunder[6pt]{Fixed}{\includegraphics[width=\celebfig\linewidth,valign=m]{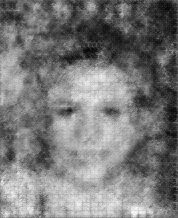}}
	\stackunder[6pt]{10 masks}{\includegraphics[width=\celebfig\linewidth,valign=m]{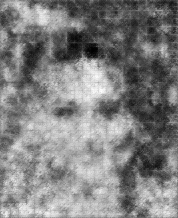}}
	\stackunder[6pt]{100 masks}{\includegraphics[width=\celebfig\linewidth,valign=m]{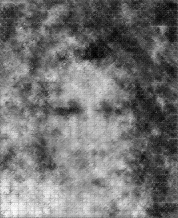}} \\[\newlineceleb]
	
	\includegraphics[width=\celebfig\linewidth,valign=m]{figs/celeba_decoder/original_1.png} \includegraphics[width=\celebfig\linewidth,valign=m]{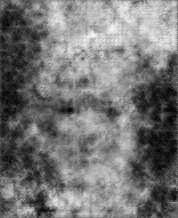}
	\includegraphics[width=\celebfig\linewidth,valign=m]{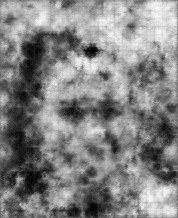}
	\includegraphics[width=\celebfig\linewidth,valign=m]{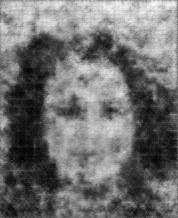}\\[\newlineceleb]
	
	\includegraphics[width=\celebfig\linewidth,valign=m]{figs/celeba_decoder/original_3.png} \includegraphics[width=\celebfig\linewidth,valign=m]{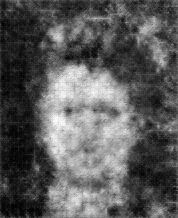}
	\includegraphics[width=\celebfig\linewidth,valign=m]{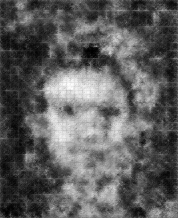}
	\includegraphics[width=\celebfig\linewidth,valign=m]{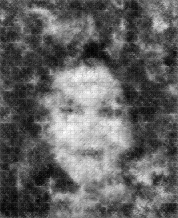}\\
	
	\caption{Example outputs of a decoder that was trained with $ 100 $ plaintext attacks of embeddings of resolution $ (24\times 32) $.}
	\label{fig:celeba_decoder_100}
\end{figure}

\cref{fig:celeba_decoder} shows example generated outputs for $ 100'000 $ plaintext attacks. \cref{fig:celeba_decoder_10k,fig:celeba_decoder_1k,fig:celeba_decoder_100} show the corresponding generated outputs for 10K, 1K, and 100 plaintext attacks respectively.

\section{Point spread functions of learned masks}

\label{sec:learned_psfs}
The point spread functions of the learned masks for the experiments in \cref{sec:mnist,sec:celeba,sec:cifar10} can be found in \cref{tab:learned_psfs}. All of these masks were initialized with the same seed, hence the similarity in pattern.

\begin{figure*}[t!]
	\centering

	\begin{subfigure}[b]{.19\textwidth}
		\centering
		\includegraphics[width=0.99\linewidth]{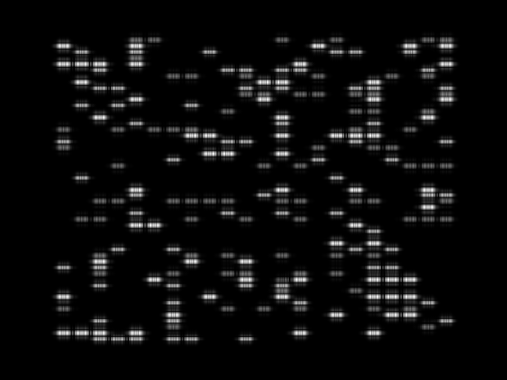}
		\caption{Seed 0.}
		\label{fig:learned_mask_gender_0_psf}
	\end{subfigure}
	\begin{subfigure}[b]{.19\textwidth}
		\centering
		\includegraphics[width=0.99\linewidth]{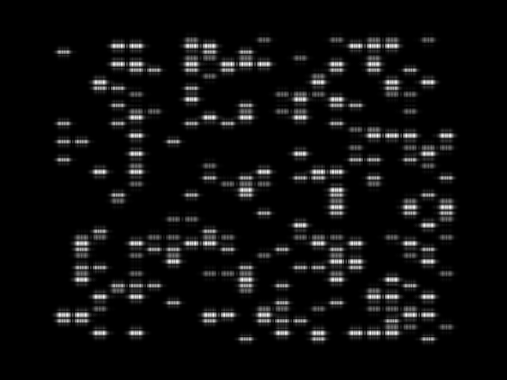}
		\caption{Seed 1.}
		\label{fig:learned_mask_gender_1_psf}
	\end{subfigure}
\begin{subfigure}[b]{.19\textwidth}
	\centering
	\includegraphics[width=0.99\linewidth]{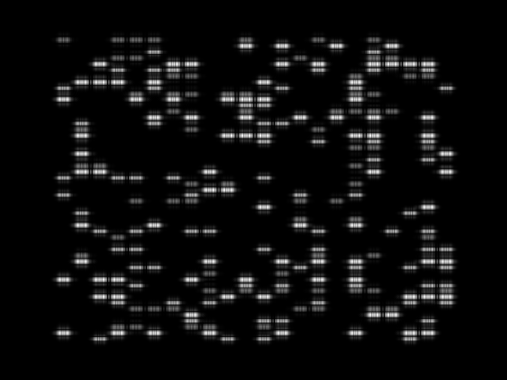}
	\caption{Seed 2.}
	\label{fig:learned_mask_gender_2_psf}
\end{subfigure}
\begin{subfigure}[b]{.19\textwidth}
	\centering
	\includegraphics[width=0.99\linewidth]{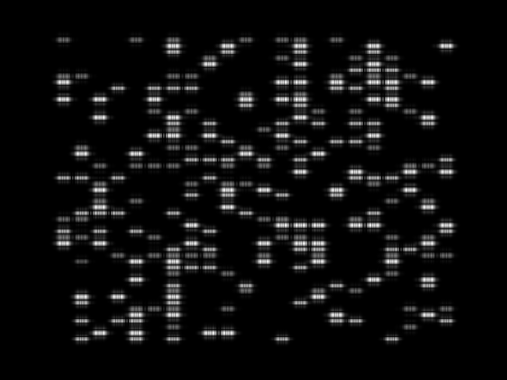}
	\caption{Seed 3.}
	\label{fig:learned_mask_gender_3_psf}
\end{subfigure}
\begin{subfigure}[b]{.19\textwidth}
	\centering
	\includegraphics[width=0.99\linewidth]{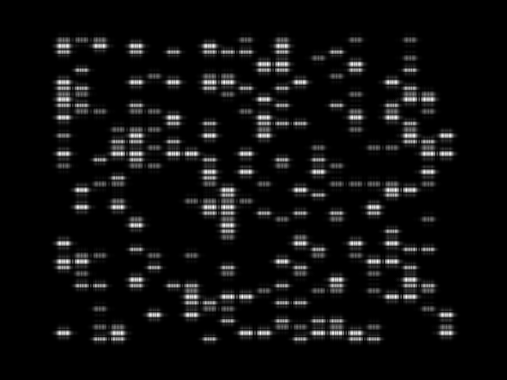}
	\caption{Seed 4.}
	\label{fig:learned_mask_gender_4_psf}
\end{subfigure}\\

	\begin{subfigure}[b]{.19\textwidth}
	\centering
	\includegraphics[width=0.99\linewidth]{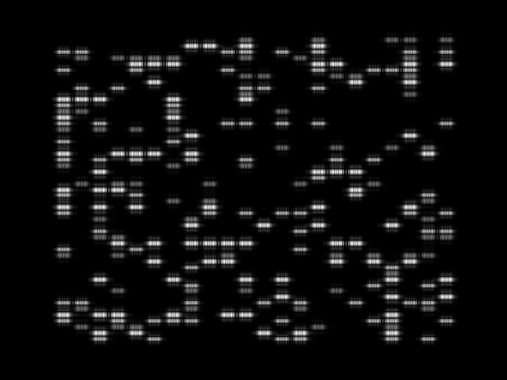}
	\caption{Seed 5.}
	\label{fig:learned_mask_gender_5_psf}
\end{subfigure}
\begin{subfigure}[b]{.19\textwidth}
	\centering
	\includegraphics[width=0.99\linewidth]{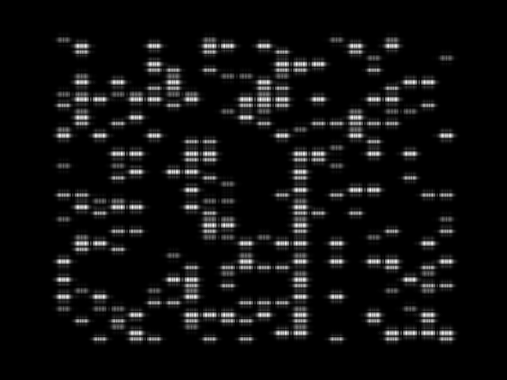}
	\caption{Seed 6.}
	\label{fig:learned_mask_gender_6_psf}
\end{subfigure}
\begin{subfigure}[b]{.19\textwidth}
	\centering
	\includegraphics[width=0.99\linewidth]{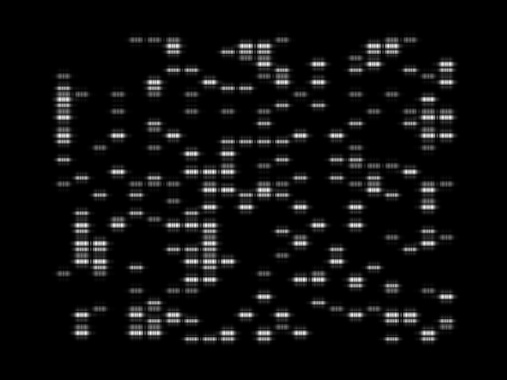}
	\caption{Seed 7.}
	\label{fig:learned_mask_gender_7_psf}
\end{subfigure}
\begin{subfigure}[b]{.19\textwidth}
	\centering
	\includegraphics[width=0.99\linewidth]{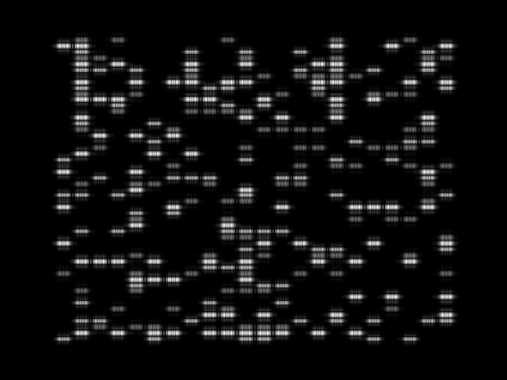}
	\caption{Seed 8.}
	\label{fig:learned_mask_gender_8_psf}
\end{subfigure}
\begin{subfigure}[b]{.19\textwidth}
	\centering
	\includegraphics[width=0.99\linewidth]{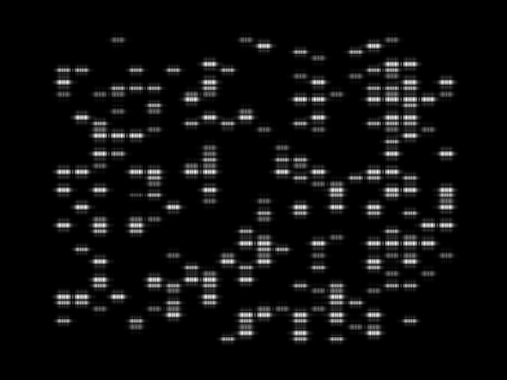}
	\caption{Seed 9.}
	\label{fig:learned_mask_gender_9_psf}
\end{subfigure}\\

	\caption{Point spread functions of learned masks for the plaintext generator experiment in \cref{sec:generator}.}
	\label{tab:learned_psfs_multiseed_gender}
\end{figure*}

\begin{figure*}[h!]
	\centering
	\includegraphics[width=0.65\linewidth]{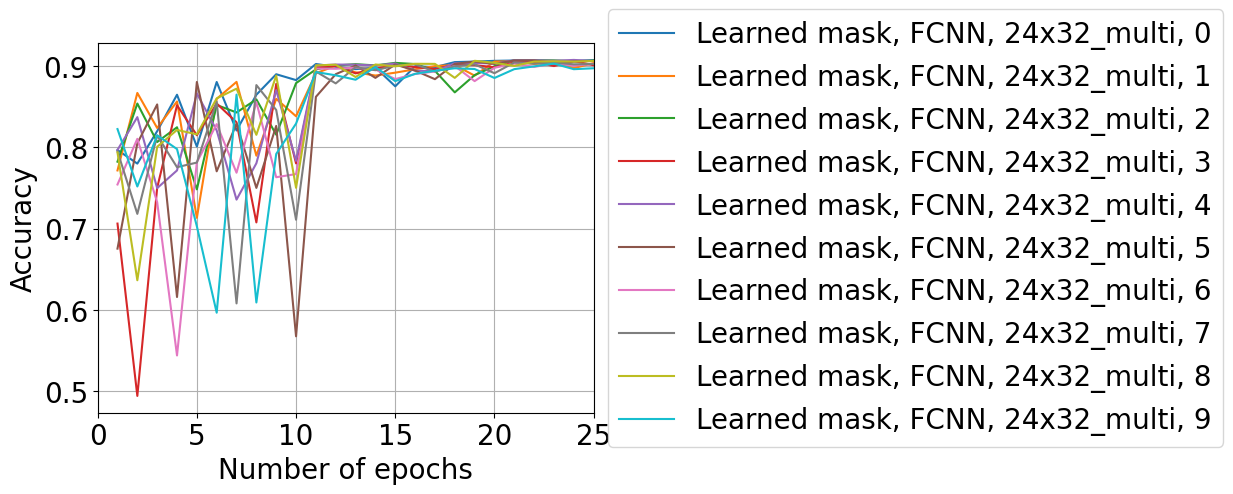}
	\caption{Test accuracy curves for different random initializations of training \textit{Learned mask} for gender classification with a two-layer fully connected neural network -- $ (24\times 32) $ embedding dimension and 800 hidden units. Learning rate is decreased every $ 10 $ epochs.}
	\label{fig:Gender_FCNN_24x32_multi}
\end{figure*}

The point spread functions of the learned masks for the plaintext generator experiment can be found in \cref{tab:learned_psfs_multiseed_gender}. These were initialized with different seeds. The test curves can be found in \cref{fig:Gender_FCNN_24x32_multi}, showing the robustness to different initializations.

\newcommand{\psfsize}{0.21}
\newcommand{\psfnewline}{3pt}
\begin{figure*}[t!]
	\begingroup
	\centering
	\renewcommand{\arraystretch}{1} 
	\setlength{\tabcolsep}{0.2em} 
	\scalebox{1.0}{
		\begin{tabular}{ccccc}
			
			\makecell{MNIST,\\Logistic\\regression,\\\cref{sec:mnist}} &
			\includegraphics[width=\psfsize\linewidth,valign=m]{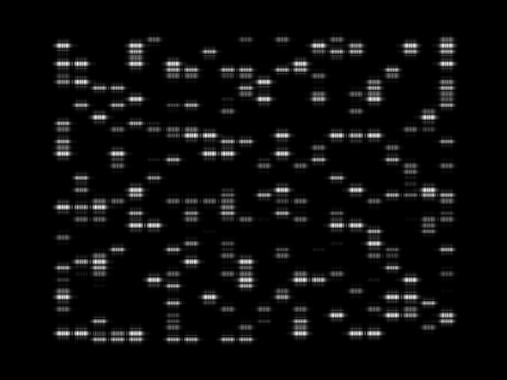}  &
			\includegraphics[width=\psfsize\linewidth,valign=m]{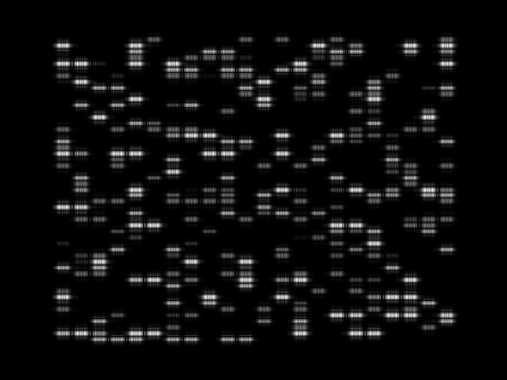} &
			
			\includegraphics[width=\psfsize\linewidth,valign=m]{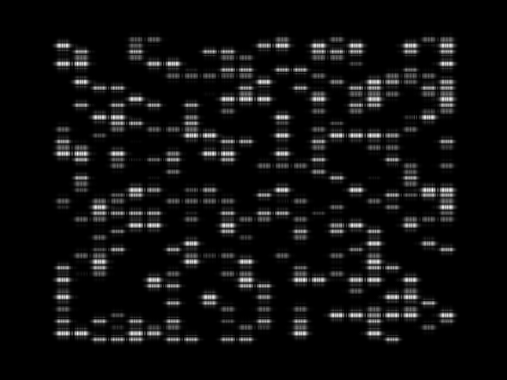}  &
			\includegraphics[width=\psfsize\linewidth,valign=m]{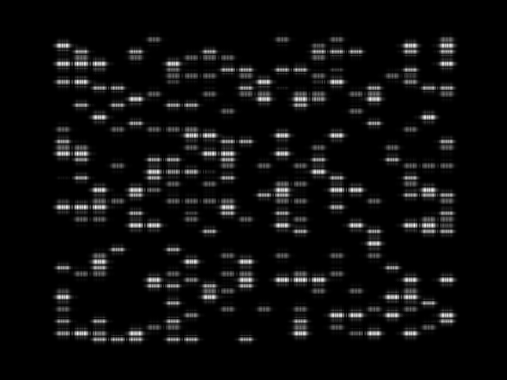}\\[\psfnewline]
			& 24$ \times $32 & 12$ \times $16 & 6$ \times $8 & 3$ \times $4\\[\psfnewline]
			
			\makecell{MNIST,\\Fully\\connected,\\\cref{sec:mnist}}  &
			\includegraphics[width=\psfsize\linewidth,valign=m]{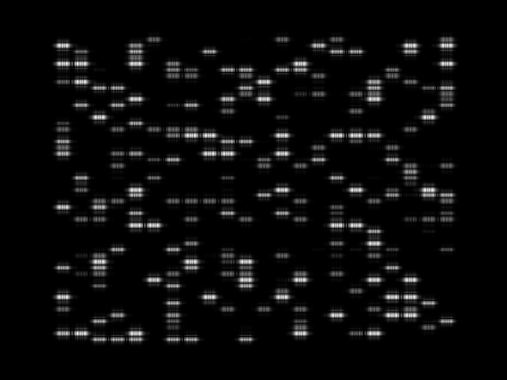}  &
			\includegraphics[width=\psfsize\linewidth,valign=m]{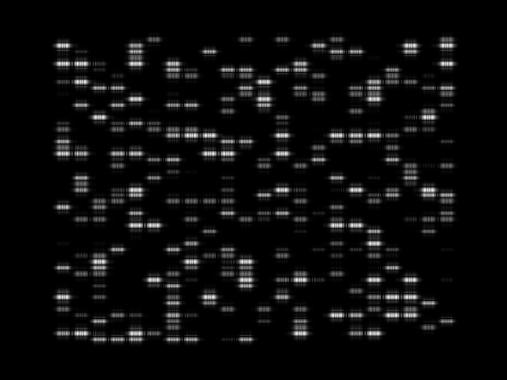} &
			
			\includegraphics[width=\psfsize\linewidth,valign=m]{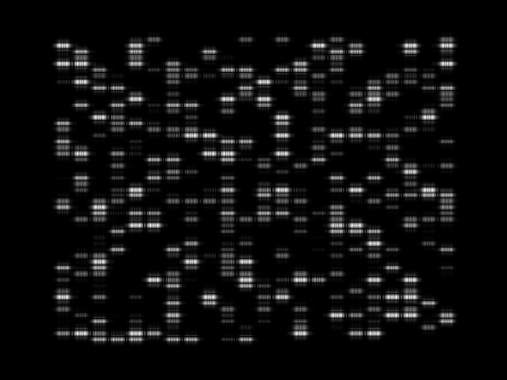}  &
			\includegraphics[width=\psfsize\linewidth,valign=m]{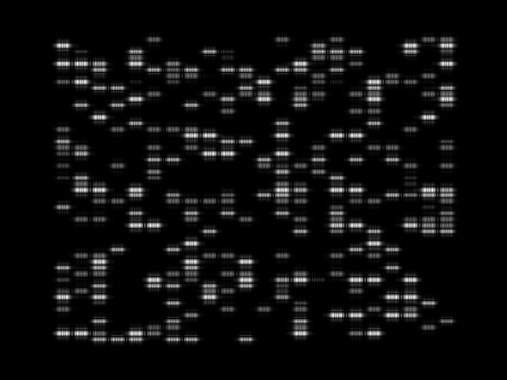}\\[\psfnewline]
			& 24$ \times $32 & 12$ \times $16 & 6$ \times $8 & 3$ \times $4\\[\psfnewline]
			
			\makecell{MNIST,\\Fully\\connected,\\Perturbed,\\\cref{sec:mnist}}  &
			\includegraphics[width=\psfsize\linewidth,valign=m]{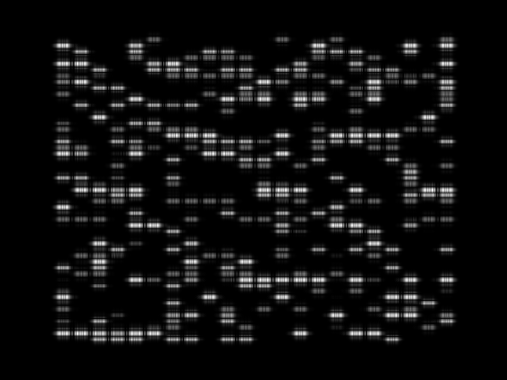}  &
			\includegraphics[width=\psfsize\linewidth,valign=m]{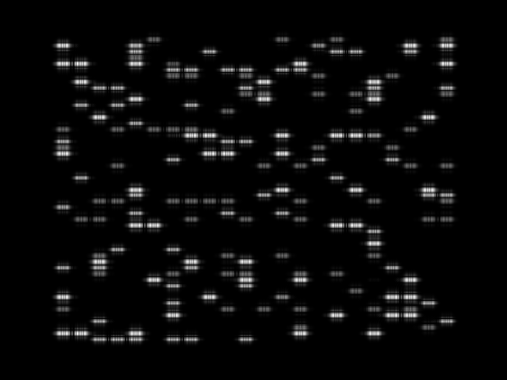} &
			
			\includegraphics[width=\psfsize\linewidth,valign=m]{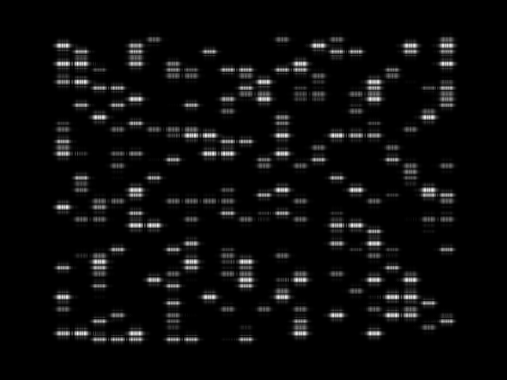}  &
			\includegraphics[width=\psfsize\linewidth,valign=m]{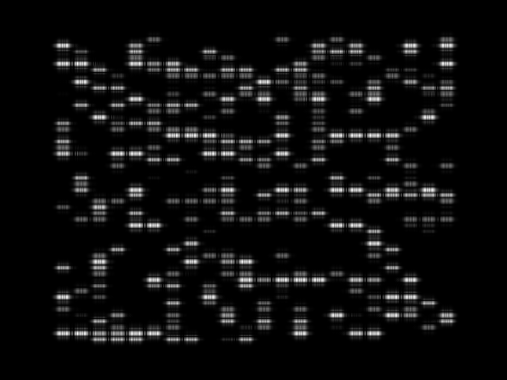}\\[\psfnewline]
			& Shift & Rescale & Rotate & Perspective\\[\psfnewline]
			
			\makecell{CelebA,\\Fully\\connected,\\\cref{sec:celeba}}  &
			\includegraphics[width=\psfsize\linewidth,valign=m]{figs/psfs/learned_mask_gender_768_psf.png}  &
			\includegraphics[width=\psfsize\linewidth,valign=m]{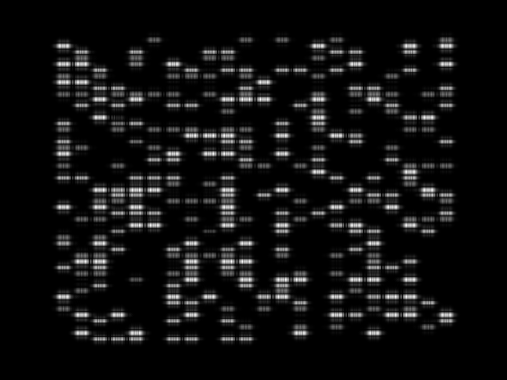} &
			
			\includegraphics[width=\psfsize\linewidth,valign=m]{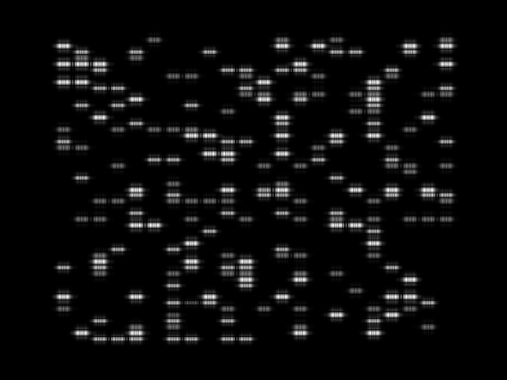}  &
			\includegraphics[width=\psfsize\linewidth,valign=m]{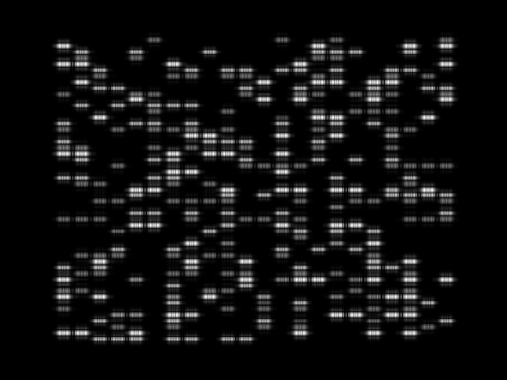}\\[\psfnewline]
			& Gender, 24$ \times $32 & Gender, 3$ \times $4 & Smiling, 24$ \times $32 & Smiling, 3$ \times $4\\[\psfnewline]
			
			\makecell{CIFAR10,\\VGG11,\\\cref{sec:cifar10}}  &
			\includegraphics[width=\psfsize\linewidth,valign=m]{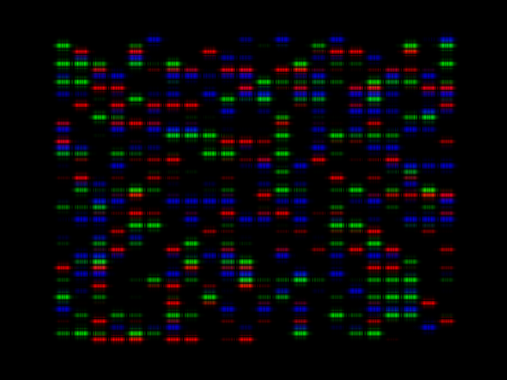}  &
			\includegraphics[width=\psfsize\linewidth,valign=m]{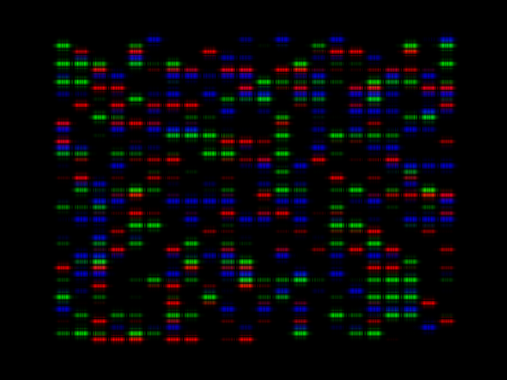} &
			
			\includegraphics[width=\psfsize\linewidth,valign=m]{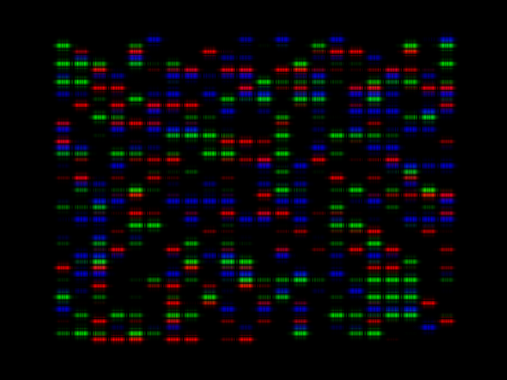}  &
			\includegraphics[width=\psfsize\linewidth,valign=m]{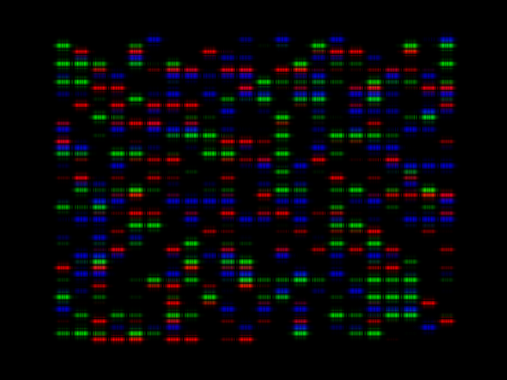}\\[\psfnewline]
			& 3$ \times $27$ \times $36 & 3$ \times $13$ \times $17 & 3$ \times $6$ \times $8 & 3$ \times $3$ \times $4
		\end{tabular}
	}
	\endgroup
	\caption{Point spread functions of learned masks for each experiments in \cref{sec:mnist,sec:celeba,sec:cifar10}.}
	\label{tab:learned_psfs}
\end{figure*}

\end{document}